\documentclass[journal,transmag]{IEEEtran}
\usepackage{amssymb}
\usepackage{amsmath}
\usepackage{commath}
\usepackage{mathtools}

\usepackage{import}
\usepackage{paralist}
\usepackage{svg}
\usepackage{textcomp}
\usepackage{gensymb}
\usepackage{dsfont}
\usepackage{subfig}
\usepackage{graphics} 
\usepackage{epsfig} 
\usepackage[normalem]{ulem}
\usepackage{siunitx}
\usepackage{bm}
\usepackage[shortlabels]{enumitem}
\usepackage{upgreek}
\usepackage{hyperref}

\DeclareMathOperator*{\minimize}{minimize}

\newcommand\Tstrut{\rule{0pt}{2.6ex}}         
\newcommand\Bstrut{\rule[-0.9ex]{0pt}{0pt}}   
\def\tablename{Table}

\newcommand\modtext[1]{#1}     
\newcommand\modtextBis[1]{#1}     
\newcommand\modtextDani[1]{#1}     

\ifCLASSINFOpdf
\else
\fi
\begin{document}
%
\title{Dynamic Complementarity Conditions and  Whole-Body Trajectory Optimization  for Humanoid Robot Locomotion}


\author{Stefano Dafarra\IEEEauthorrefmark{1,*},
Giulio Romualdi\IEEEauthorrefmark{1,3,*}, and
Daniele Pucci\IEEEauthorrefmark{1,2,*}
\thanks{\IEEEauthorrefmark{1}\modtext{Artificial and Mechanical Intelligence}, Italian Institute of Technology, Genoa, Italy, {\tt\small (e-mail: name.surname@iit.it)}.}
\thanks{\IEEEauthorrefmark{2}\modtext{Machine Learning and Optimisation, University of Manchester, Manchester, UK}.}
\thanks{\IEEEauthorrefmark{3}DIBRIS, University of Genoa, Genoa, Italy.}
\thanks{\IEEEauthorrefmark{*}~Member~IEEE.}
}


%




\maketitle


\begin{abstract}
The paper presents a planner to generate walking trajectories by using the centroidal dynamics and the full kinematics of a humanoid robot. The interaction between the robot and the walking surface is modeled explicitly via new conditions, the \emph{Dynamical Complementarity Constraints}. The approach does not require a predefined contact sequence and generates the footsteps automatically. 
We characterize the robot control objective via a set of tasks, and we address it by solving an optimal control problem. We show that it is possible to achieve walking motions automatically by specifying a minimal set of references, such as a constant desired center of mass velocity and a reference point on the ground. Furthermore, we analyze how the contact modelling choices affect the computational time. We validate the approach by generating and testing walking trajectories for the humanoid robot iCub.
\end{abstract}


%
\IEEEpeerreviewmaketitle

\section{Introduction}
\IEEEPARstart{T}{he} general problem of planning whole-body locomotion trajectories of humanoid robots remains an interesting and partially open challenge for the Robotics community~\cite{harada2010motion}.
Whole-body \emph{humanoid robot motion planning}, in fact, is often associated \modtextDani{with} optimization problems \modtextDani{having} high-dimensional search space, which further complexifies the process of finding either local or global solutions. \modtextDani{This process requires sophisticated models that characterise the robot
motions and interactions with its surrounding environment.
When 
one of the robot bodies makes contact with a \emph{rigid} environment, 
the models that may describe the robot motions and contact forces 
are known as \emph{complementarity conditions}, which add a degree of complexity when solving 
the associated whole-body trajectory optimization problem.}
A common approach to circumvent this barrier is to consider simplified, and often linear, robot models that seldom consider the contacts between the humanoid robot and its surrounding\modtextDani{s.}
This paper presents a non-linear trajectory optimization approach that employs novel definitions of the complementarity conditions tailored for \emph{whole-body humanoid robot motion planning}, here called \emph{dynamic complementarity conditions} (DCCs).

\modtextDani{After} the DARPA Robotics Challenge~\cite{DRC-what-happened}, it \modtextDani{became} popular to tackle  humanoid robot locomotion  using a hierarchical control architecture~\cite{feng2015optimization}.
Each loop of the architecture receives inputs from the robot and the environment, and provides references to the loop next. The inner the layer, the shorter the time horizon that is used to evaluate the outputs. Also, inner loops usually employ more complex robot models to evaluate their outputs, but the shorter time horizon often results in faster computations for obtaining these outputs.
More precisely, from outer to inner, the hierarchical control architecture is usually composed of the following loops: the \emph{trajectory optimization}; the \emph{simplified model control}; and the \emph{whole-body quadratic programming (QP) control} loop
\modtextDani{\cite{carpentier2017multi, romualdi2020benchmarking}}.

The \emph{trajectory optimization} loop is often in charge of generating foothold trajectories starting from high-level commands, such as those coming from \modtextDani{the user via a joypad}. The output of this layer is given to the \emph{simplified model control} loop --~also called \emph{receding horizon}~\cite{Mayne2000Stability}. Its aim is to generate desired and feasible \emph{centroidal} quantities associated with \emph{stable} walking instances~\modtextDani{\cite{orin08}}. 
The output of this loop is fed into the \emph{whole-body QP control} layer, which is in charge of stabilizing the planned trajectories exploiting the full robot model with a suitable \emph{Quadratic Programming} formulation. The \emph{trajectory optimization} and the \emph{simplified model control} loops aim at generating robot   trajectories, \modtextDani{so can be viewed as} 
the \emph{planning} layers of the architecture. The present paper proposes novel methods for combining these two layers\modtextDani{. W}hat follows is a short recap of the state of the art concerning the planning algorithms employed in the  \emph{trajectory optimization} and the \emph{simplified model control} loop\modtextDani{s}. 

The \emph{trajectory optimization} layer mainly deals with defining the contact locations of the \modtextDani{robot} locomotion pattern. Several strategies exist to address this problem, and they can be categorized depending on the amount of hand-crafted contact
 information an external user needs to specify to run the planning algorithms.
\emph{Fixed contact sequence, timing and location} planning assumes that contact positions and timings are either specified by an external user or by an outer control loop. Planning then focuses on the CoM state \modtextDani{\cite{romualdi2022online}}. In this case, the main complexity is
\modtextDani{given} 
by the non-linearity of the angular momentum dynamics. This problem can be faced by considering only the CoM linear acceleration~\cite{caron2016multi}, or by using a convex upper bound of the angular momentum to be minimized~\cite{daiplanning}. 
A convex optimization problem is obtained also by forcing the CoM trajectory to a polynomial with only one free variable~\cite{fernbach2019c}. 
Body postures can be planned together with the centroidal quantities, thus considering the robot kinematic structure 
\modtextDani{in the same formulation}~\cite{herzog2015trajectory,kudruss2015optimal, posa2016optimization, serra2016newton, fernbach2018croc}. These methods\modtextDani{, however, 
require} external contact planners. 
%
\modtextDani{Analogously, 
\emph{Differential Dynamic Programming} based methods 
can} generate whole-body motions~\cite{feng20133d, budhiraja2018differential, giraud2020motion, mastalli2020crocoddyl}\modtext{, and track them online \cite{dantec2021whole}.}
\emph{Predefined contact sequence} planning assumes to know in advance the contact sequence, \modtextDani{although the contact locations and timings are not fixed} \cite{carpentier2016versatile,caron2017make,winkler18}. 
Different sets of equations for each contact phase usually model the hybrid nature arising from the pattern of making and breaking contacts. 
\modtextDani{To solve the associated problem, one may use \emph{Mixed integer programming} that uses integer variables in the definition of the 
optimization framework 
to determine where and at which time a contact should be 
made \cite{deits2014footstep, mason2018mpc, mirjalili2018whole,ibanez2014emergence, aceituno2018simultaneous}}. 
\modtextDani{Exploiting integer variables, however, strongly affects the computational performances, especially 
when the robot may make several contacts with the environment.}
\emph{Complementarity-free} planning approaches model multi-body systems subject to contacts without directly enforcing the complementarity conditions \cite{drumwright2010modeling}. Equivalently accurate results are obtained by maximizing the rate of energy dissipation \cite{todorov2011convex}.
These methods consider contacts explicitly in the planner \modtextDani{formulation}, so contact location, timing and sequence are obtained as solutions to the optimization problem, \modtextDani{thus} generating complex robot motions \cite{mordatch2012discovery, tassa2012synthesis, erez2013integrated}. 

The other planning layer, the \emph{simplified model control} layer, is in charge of finding feasible trajectories for the robot center of mass (CoM) along the walking path. The computational burden to find feasibility regions, however, usually calls for simplified models to characterize the robot dynamics.
Indeed, when restricting the CoM on a plane at a fixed height and assuming constant angular momentum, it is possible to derive simple and effective control laws based on the Linear Inverted Pendulum (LIP) model \cite{Kajita2001, kajita20013d}, \modtext{the Capture Point \cite{Pratt2006}, and the \emph{Divergent Component of Motion} (DCM) \cite{takenaka2009real}}. 
These simplified linear models have allowed on-line model predictive control \cite{wieber2006trajectory,missura2014balanced, naveau2016reactive, griffin2016model, joe2018balance}, also providing  references for the footstep locations\modtext{.}

An emerging approach is that of combining the \emph{simplified model control}  into the \emph{trajectory optimization} layer so as to obtain both foothold trajectories and feasible centroidal quantities with a single optimization problem, which often employs complete or reduced robot models. When complete robot models are combined with impact dynamics, the control problem increases \modtextDani{its complexity considerably}. 

\modtext{
\subsection{Related works}
Neunert, et. al., \cite{neunert2018whole} presented one of the first real-time implementation of an MPC controller adopting the full robot model with contacts. This has been tested on a quadruped robot adopting a tailored soft contact model. Nevertheless, the use of soft contact models presents some limitations when applied to robots with finite size feet \cite{romualdi2021modeling}.
}

\modtext{
Instead of considering the entire robot model in a single optimal control problem, another possibility consists in \modtextDani{planning trajectories for 
the} centroidal and joint quantities separately, iterating until a consensus is reached between the two \modtextDani{sets of trajectories} \cite{herzog2015trajectory}. The footstep locations and timings can also be determined while planning for the robot momentum \cite{ponton2021efficient}.
}

\modtext{
Dai, et. al., \cite{dai2014whole} exploit the full robot kinematics and its centroidal dynamics in a single optimal control problem. They \modtextDani{follow the approach proposed by} Posa, et. al. \cite{posa2014direct} on the consecutive relaxations of the complementarity conditions, 
\modtextDani{so they avoid
to consider} soft contact models. Contact sequences can be
\modtextDani{predefined} or defined by the optimizer.
}
\modtext{
\subsection{Contribution}
}
This paper proposes new methods to \modtextDani{consider the \emph{simplified model control} objectives as part of
the} \emph{trajectory optimization} layer. So, feasible footsteps and whole-body robot trajectories are obtained as the solution to a single non-linear optimization problem. With our approach, neither contact sequences nor contact locations and timings need to be known beforehand, and no hand-crafted a-priori knowledge is injected in the system to generate walking motions. 

\modtext{
\modtextDani{The approach we propose 
follows similar arguments and methods presented 
in Dai, et. al., \cite{dai2014whole}. In particular, we take the same modeling approach that consists in} considering the full robot kinematics and  centroidal momentum. On the other hand, we introduce a \modtextDani{new and}  alternative formulation of the classical complementarity conditions 
tailored for trajectory optimization problems aimed at finding robot locomotion patterns. These are called \emph{dynamic complementarity conditions (DCCs)} and represent the first contribution of the paper.
}

\modtext{
As a second contribution, we show how the DCCs can be 
\modtextDani{embedded}
into a whole-body non-linear optimization framework to find feasible footsteps and whole-body robot trajectories with no a-priori knowledge about the robot locomotion. This represents another difference compared to the work \modtextDani{by} Dai, et. al., \cite{dai2014whole}, where a postural term in the cost function determines the robot motion\modtextDani{, and}
the contacts are predefined in the majority of the cases.
}

\modtext{
\modtextDani{We compute the robot trajectories
using} the \emph{Receding Horizon} principle. \modtextDani{We validate the whole approach 
using} the real iCub Humanoid robot. Finally, we perform a comparison between the DCCs and other classical complementarity formulations to show that DCCs \modtextDani{may lead to improved} 
computational time and contact condition accuracy. 
This represents the third contribution of the paper.
}

This paper extends \modtextDani{and encompasses} the authors previous work \cite{dafarra2020whole}.
\modtext{
The main differences \modtextDani{between the two papers} follow. $i)$~\modtextDani{The} previously introduced contact parametrization~\cite{dafarra2020whole} \modtextDani{is presented as one of the DCCs} and we compare its performances with other methods. \modtextBis{The comparison is performed in two contexts and with a crescendo of complexity. First, we analyze the performances of the complementarity methods when applied to a simple toy problem. Then, we use the whole-body non-linear optimization framework presented in the paper as a testing ground.} The comparisons run both on a personal laptop, and by using \texttt{Github Actions}, thus exploiting a reproducible setup. $ii)$~We validate the generated trajectories on the real iCub Humanoid robot. \modtextBis{$iii)$~We update the formulation of the whole-body non-linear optimization framework adding more tasks that lead to a better generation of walking trajectories.}
}

The remainder of the paper is as follows. Section \ref{sec:background} presents some basic humanoid robot modeling tools and introduces to the complementarity conditions of rigid contact models. Section \ref{sec:dcc} presents the \modtext{DCCs}. These are then adopted in a non-linear trajectory optimization framework, described in Sec. \ref{sec:dcc_nlmpc}. The output of the planner is validated and tested on the real humanoid robot iCub in Sec. \modtextBis{\ref{sec:validation}, while we draw comparisons between the DCCs and state of the art methods in Sec. \ref{sec:complementarity_comparison}.} Sec. \ref{sec:conclusions} concludes the paper.

\section{Background}
\label{sec:background}
\subsection{Notation}\label{sec:notation}
Throughout the paper we use the following notation.
\begin{itemize}
	\item Vector and matrices are expressed in bold symbols.
	\item The $i_{th}$ component of a vector $\bm{x}$ is denoted as $x_i$. 
	\item The transpose operator is denoted by $(\cdot)^{\top}$.
	\item The superscript $(\cdot)^*$ indicates desired values.
	\item Given a function of time $f(t)$, the dot notation denotes its time derivative, i.e.
		$\dot{f} := \frac{\dif f}{\dif t}$. Higher order derivatives are denoted with a corresponding amount of dots.
	\item $\mathcal{I}$ is a fixed inertial frame with respect to (w.r.t.) 
	which the robot's absolute pose is measured. Its $z$ axis is supposed to point against gravity, while the $x$ direction defines the forward direction.
	\item $\mathds{1}_n \in \mathbb{R}^{n \times n}$ denotes the identity matrix of dimension $n$.
	\item $\bm{0}_{n \times n} \in \mathbb{R}^{n \times n}$ denotes a zero matrix, while $\bm{0}_n = \bm{0}_{n \times 1}$ is a zero column vector of size $n$.
	\item $\bm{e}_i$ is the canonical base in $\mathbb{R}^n$. It corresponds to $\bm{e}_i = [0, 0, \dots, 1, 0, \dots, 0]^\top \in \mathbb{R}^n$, where the only unitary element is in the $i$-th position.
	\item The operator $\wedge$ defines the skew-symmetric operation associated with the cross product in $\mathbb{R}^3$. For example, $\left(a^\wedge\right)b$, with $a,b \in \mathbb{R}^3$, is equivalent to $a\times b$. Its inverse is the operator vee $\vee$, i.e. $\left(a^\wedge\right)^\vee = a$.
	\modtextBis{\item Given $\bm{x} \in \mathbb{R}^n$, and $\alpha \in \mathbb{R}$, $\bm{x} \succcurlyeq \alpha$ indicates that each component of $\bm{x}$ is greater or equal than $\alpha$, i.e. $x_i \geq \alpha ~ \forall i \in [1, n]$.}
	\item The weighted L2-norm of a vector $\bm{v} \in \mathbb{R}^n$ is denoted by $\|\bm{v}\|_{\bm{W}}$, $\bm{W} \in \mathbb{R}^{n\times n}$ \modtext{is} a positive definite weight matrix.
	\item $^{A}\bm{R}_{B} \in SO(3)$ and $^{A}\bm{H}_{B} \in SE(3)$ denote the rotation and transformation matrices which transform a vector expressed in the $B$ frame, $^B \bm{x}$, into a vector expressed in the $A$ frame, $^A \bm{x}$.
	\item ${}^D\bm{V}_{A,D} \in \mathbb{R}^6$ is the relative velocity between frame $A$ and $D$,  and its coordinates are expressed in the frame $D$. \modtext{The first three coordinates refer to the linear part of the velocity, while the bottom three to the angular velocity.}
	\item $\bm{x} \in  \mathbb{R}^3$ is the position of the center of mass \modtext{w.r.t.} $\mathcal{I}$.
	\item $\mathbf{n}(\cdot): \mathbb{R}^3 \rightarrow \mathbb{R}^3$ is a function returning the direction that is perpendicular to the walking plane given the $x$ and $y$ coordinates of the input point.
	\item $\mathbf{t}(\cdot): \mathbb{R}^3 \rightarrow \mathbb{R}^{3\times2}$ is a function returning two perpendicular directions normal to $\mathbf{n}(\cdot)$. The composition of $\mathbf{t}(\cdot)$ and $\mathbf{n}(\cdot)$, i.e. $\left[\mathbf{t}(\cdot)~\mathbf{n}(\cdot)\right]$, defines the rotation matrix ${}^\mathcal{I}\bm{R}_{plane}$.
	\item $h(\bm{p}): \mathbb{R}^3 \rightarrow \mathbb{R}$ defines the distance between $\bm{p}$ and the walking surface. 
	\item $\modtextDani{\text{diag}}(\cdot): \mathbb{R}^n \rightarrow \mathbb{R}^{n \times n}$ is a function casting the argument into the corresponding diagonal \modtext{matrix}.
	
\end{itemize}

\subsection{Humanoid Robot Modeling}
\label{sec:modelling}
The robot configuration space is characterized by the \emph{position} and the \emph{orientation} of the base frame $B$, and the joint configurations. Thus, it corresponds to the group $\mathbb{Q} = \mathbb{R}^3 \times SO{(3)} \times \mathbb{R}^n$ and an element $\bm{q} \in \mathbb{Q}$ can be defined as the following triplet: $\bm{q} = ({}^{\mathcal{I}}\bm{p}_B, {}^{\mathcal{I}}\bm{R}_{B}, \bm{s})$.

The \emph{velocity} of the multi-body system can be characterized by the \emph{algebra} $\mathbb{V}$ of $\mathbb{Q}$ defined by: $\mathbb{V} = \mathbb{R}^3 \times \mathbb{R}^3 \times \mathbb{R}^n$.
An element of $\mathbb{V}$ corresponds to $\bm{\nu}$.

We also assume that the robot is interacting with the environment exchanging $n_c$ distinct wrenches\footnote{As an abuse of notation, we define as \emph{wrench} a quantity that is not the dual of a 
	\emph{twist}, but a 6D force/moment vector.}. 
	By employing the Euler-Poincar\'e formalism \cite[Ch. 13.5]{Marsden2010}, we obtain:
\begin{equation}
	\label{eq:system_initial}
	\bm{M}(\bm{q})\dot{\bm{\nu}} + \bm{C}(\bm{q}, \bm{\nu})\bm{\nu} + \bm{G}(\bm{q}) =  \begin{bmatrix}
		\bm{0}_{6\times n} \\ \mathds{1}_n
	\end{bmatrix}\bm{\tau}_s + \sum_{k = 1}^{n_c} \bm{J}^\top_{\mathcal{C}_k} {}_k\textbf{f},
\end{equation}
where $\bm{M} \in \mathbb{R}^{\modtext{(n+6) \times (n+6)}}$ is the mass matrix, 
$\bm{C} \in \mathbb{R}^{(n+6) \times (n+6)}$ accounts for Coriolis and centrifugal effects, $\bm{G} \in \mathbb{R}^{n+6}$ is the gravity term. \modtextDani{The vector} $\bm{\tau}_s \in \mathbb{R}^{n}$ 
\modtextDani{represents the} 
actuation torques, while ${}_k\textbf{f} \in \mathbb{R}^{6}$ denotes the $k$-th external wrench applied by the environment on the robot\modtextDani{, namely}
\begin{equation}
	{}_k\textbf{f} = \begin{bmatrix}
		{}_k\bm{f} \\
		{}_k\bm{\tau}
	\end{bmatrix},
\end{equation} 
with ${}_k\bm{f},\, {}_k\bm{\tau} \in \mathbb{R}^3$ being the contact force and torque respectively. We assume the wrenches to be measured in a frame located on the contact link origin and oriented as $\mathcal{I}$.

As described in \cite[Sec. 5]{traversaro2017thesis}, it is possible to apply a coordinate transformation in the state space $(q,{\nu})$ that transforms the system dynamics~\eqref{eq:system_initial} into a new form where
the mass matrix is block diagonal, thus decoupling joint and base frame accelerations. Also, in this new set of coordinates,  the first six rows of Eq. \eqref{eq:system_initial} correspond to the \emph{centroidal dynamics}. In the specialized literature, this term is used to indicate the rate of change of the robot's momentum expressed at the center of mass, which then equals the summation of all external wrenches acting on the multi-body system \cite{orin2013centroidal, wensing2016improved}.
\subsection{Centroidal dynamics}\label{sec:intro_momentum}
By definition, the center of mass (CoM) $\bm{x} \in \mathbb{R}^3$ corresponds to the weighted average of all the links' CoM position:
\modtextDani{
\begin{equation}\label{eq:com_definition}
	\bm{x} = {}^\mathcal{I} \bm{H}_B\sum_i {}^{B} \bm{H}_i  ~ {}^i\textbf{p}_{\text{CoM}} \frac{m_i}{m},
\end{equation}
}
where ${}^i\textbf{p}_{\text{CoM}} \in \mathbb{R}^3$ is the (constant) CoM position of \modtextDani{the $i$th link expressed w.r.t. the $i$th coordinate systems. The scalars} $m, m_i \in \mathbb{R}^+$ are the robot total mass and the $i$-th link mass\modtextDani{, respectively.} 
In order to introduce the \emph{centroidal dynamics}, it is convenient to define a frame, called $\bar{G}$, whose origin is located on the CoM, while the orientation is parallel to the inertial frame $\mathcal{I}$.
We introduce ${}_{\bar{G}} \bm{h} \in \mathbb{R}^6$ \modtextDani{to denote} the robot total momentum expressed \modtextDani{w.r.t. $\bar{G}$, namely}:
\begin{equation}
	{}_{\bar{G}} \bm{h} = \begin{bmatrix}
		{}_{\bar{G}} \bm{h}^p \\
		{}_{\bar{G}} \bm{h}^\omega
	\end{bmatrix},
\end{equation}
where ${}_{\bar{G}} \bm{h}^p \in \mathbb{R}^3$ and ${}_{\bar{G}} \bm{h}^\omega \in \mathbb{R}^3$ are the linear and angular momentum\modtextDani{, respectively}. In addition, the following holds:
\begin{equation}\label{eq:com_from_momentum}
	\dot{\bm{x}}_\text{CoM} = \frac{1}{m}{}_{\bar{G}} \bm{h}^p.
\end{equation}

The robot total momentum corresponds to the summation of all the links momenta ${}_B\bm{h}^i$, measured in the base frame and projected on $\bar{G}$:
\begin{equation}
	{}_{\bar{G}}\bm{h} = \sum_i {}_{\bar{G}} \bm{X}^B {}_B\bm{h}^i.
\end{equation}
\modtextDani{The \emph{adjoint matrix} ${}_{\bar{G}} \bm{X}^B$} transforms a wrench expressed in $B$ into one expressed in $\bar{G}$. 
\modtextDani{Then, ${}_{\bar{G}}\bm{h}$ writes: }
\begin{equation}\label{eq:momentumExpanded}
	{}_{\bar{G}} \bm{h} = {}_{\bar{G}} \bm{X}^B \sum_i {}_{B}\bm{X}^i \bm{I}_i {}^i \bm{V}_{A,i},
\end{equation}
with $\bm{I}_i \in \mathbb{R}^{6\times 6}$ being the (constant) link inertia expressed in link frame. Hence, \modtextDani{one obtains:}
\begin{equation}\label{eq:cmm_intro}
	{}_{\bar{G}} \bm{h} = \bm{J}_\text{CMM}\bm{\nu},
\end{equation}
where $\bm{J}_\text{CMM} \in \mathbb{R}^{6\times n}$ is the \emph{Centroidal Momentum Matrix} (CMM) \cite{orin08}.
\modtextDani{The rate of change of the} centroidal momentum balances the external wrenches applied to the robot\modtextDani{, which yields:}
\begin{equation}\label{eq:centroidal_momentum_dynamics}
	\begin{split}
		{}_{\bar{G}} \dot{\bm{h}} &= \sum_{k = 1}^{n_c} {}_{\bar{G}}\bm{X}^k {}_k\textbf{f} + m \bar{\bm{g}}, \\
		&= \sum_{k = 1}^{n_c} \begin{bmatrix}
			{}^{\mathcal{I}}\bm{R}_k & \bm{0}_{3\times 3} \\
			({}^{\mathcal{I}}\bm{o}_k - \bm{x})^\wedge\,{}^{\mathcal{I}}\bm{R}_k & {}^{\mathcal{I}}\bm{R}_k
		\end{bmatrix} {}_k\textbf{f} + m \bar{\bm{g}}. 
	\end{split}
\end{equation}
The adjoint matrix ${}_{\bar{G}}\bm{X}^k \in \mathbb{R}^{6 \times 6}$ transforms the contact wrench from the application frame (located in ${}^{\mathcal{I}}\bm{o}_k$ with orientation ${}^{\mathcal{I}}\bm{R}_k$) to $\bar{G}$. Finally, $\bar{\bm{g}} = \left[\begin{smallmatrix} 0 & 0 & -g & 0 & 0 & 0\end{smallmatrix}\right]^\top$ is the 6D gravity acceleration vector.

Alternatively, the centroidal momentum dynamics can be obtained by differentiating Eq. \eqref{eq:cmm_intro}\modtextDani{, which writes:}
\begin{equation}\label{eq:momentum_derivative_cmm}
	{}_{\bar{G}} \dot{\bm{h}} = \bm{J}_\text{CMM}\dot{\bm{\nu}} +  \dot{\bm{J}}_\text{CMM}\bm{\nu},
\end{equation}
thus highlighting \modtextDani{the dependency of  ${}_{\bar{G}} \dot{\bm{h}}$ on $\dot{\bm{\nu}}$}.

\subsection{Recall on complementarity conditions}\label{sec:complementarity_intro}
Humanoid robots are 
\modtextDani{often equipped with flat feet, which simplify the walking and balancing task with respect to line and point foot.}
It is important, then, to model how the foot can land on the walking surface, also limiting the set of admissible contact wrenches. As an example, in case of line contacts, no torque can be exerted along the contact line. 

A common approach is to consider the foot as composed by a set of points, for example (but not limited to) four points located at the corners of the foot, \cite{wensing2013generation,dai2014whole, caron2017make}.
A pure force is supposed to be applied on each of the contact points. 

Define ${}_i\bm{p} \in \mathbb{R}^3$ as the $i-$th contact point location in an inertial frame $\mathcal{I}$, and ${}_i\bm{f} \in \mathbb{R}^3$ as the force exerted on that point. Such force is expressed on a frame located in ${}_i\bm{p}$ and with orientation parallel to $\mathcal{I}$.

In this paper, we exploit a \modtextDani{\emph{rigid contact model} to characterise the interactions between the robot feet and the environment floor. In the literature, other contact models exist, e.g., 
\emph{compliant models} that allow
a }degree of compenetration \cite{neunert2018whole, azad2013new, liu2015reactive}.
In a rigid setting\modtextDani{,} instead, the contact points are not supposed to penetrate the walking ground, i.e.:
$$
h({}_i \bm{p}) \geq 0. 
$$

The force ${}_i\bm{f}$ results from the interaction \modtextDani{between} the contact point \modtextDani{and} the ground, hence\modtextDani{, it is bounded}. Being a reaction force, its normal component with respect to the walking ground is supposed to be non-negative. In particular,
\begin{equation}\label{eq:forceNormality}
	\mathbf{n}({}_i\bm{p})^\top{}_i\bm{f} \geq 0.
\end{equation}
Additionally, we focus on the case where the contact force is not enough to overcome the static friction:
\begin{equation} \label{eq:friction}
	\|\mathbf{t}({}_i\bm{p})^\top {}_i\bm{f} \| \leq \mu_s ~\mathbf{n}({}_i\bm{p})^\top{}_i\bm{f},
\end{equation}
where \modtextDani{$\mu_s \in \mathbb{R}^+$} is the static friction coefficient. \modtext{It can be shown that, for what concern the validity of the contact wrench, it is sufficient to guarantee that the forces at the vertices satisfy  \eqref{eq:forceNormality}$-$\eqref{eq:friction} \cite{caron2015leveraging}.}


\modtextDani{Being a reaction force, 
the value of ${}_i\bm{f}$ 
is different from zero 
only if the associated contact point is in contact with the walking surface. This condition can be compactly written as:} 
\begin{equation}\label{eq:complementarity}
	h({}_i \bm{p}) ~\mathbf{n}({}_i\bm{p})^\top{}_i\bm{f} = 0.
\end{equation}
Such a constraint \modtextDani{may lead to a number of numerical issues when solving 
an optimization problem that makes use of it.} This is due to the fact that the feasible set is only \modtextDani{defined} by two lines, namely $h({}_i \bm{p}) = 0$ and $\mathbf{n}({}_i\bm{p})^\top{}_i\bm{f} = 0$, that intersect the origin. In particular, at this point, the \modtextDani{so-called} \emph{constraint Jacobian} is singular, thus violating the \emph{linear independence constraint qualification} (LICQ), on which most off-the-shelf solvers rely  \cite{betts2010practical}.


The conditions  \modtextDani{given by} \eqref{eq:complementarity} can be \emph{relaxed} considering that both $h({}_i \bm{p})$ and $\mathbf{n}({}_i\bm{p})^\top{}_i\bm{f}$ are positive quantities. Hence, instead of using an equality condition, it is possible to upper-bound their product with a small positive constant $\epsilon \in \mathbb{R}^+$:
\begin{equation}\label{eq:classical_complementarity}
	h({}_i \bm{p}) ~\mathbf{n}({}_i\bm{p})^\top{}_i\bm{f} \leq \epsilon.
\end{equation}
\modtextDani{The inequality \eqref{eq:classical_complementarity} is  referred to as \emph{Relaxed Complementarity}. Thanks to this relaxation, the feasibility region increases in its dimension. 
This common approach 
also corresponds to using
``bounded'' slack variables \cite{moreau1988unilateral, stewart2000implicit, posa2014direct,betts2010practical}.}

\modtext{\modtextDani{Once a contact is made, t}he Relaxed Complementarity condition cannot prevent the contact point to move on the walking plane\modtextDani{: even if the} friction constraints defined \modtextDani{by}  \eqref{eq:friction} are satisfied, the contact points are still free to move on the walking surface. \modtextDani{This problem is often} faced by imposing \modtextDani{that the product between the planar velocities and tangential forces is equal to zero \cite{dai2014whole, posa2014direct}, the relaxed version of which writes:}
\begin{equation}\label{eq:classicalPlanarComplementarity}
   -\bm{\epsilon}_p \leq \modtextDani{\text{diag}}\left(\mathbf{t}\left({}_i\bm{p}\right)^\top {}_i\dot{\bm{p}}\right)\mathbf{t}\left({}_i\bm{p}\right)^\top {}_i\bm{f} \leq \bm{\epsilon}_p,
\end{equation}
with $\bm{\epsilon}_p \in \mathbb{R}^2, \bm{\epsilon}_p \succcurlyeq 0$ the relaxation parameters.
\modtextDani{The relations \eqref{eq:classicalPlanarComplementarity} are referred to} as \emph{Relaxed Planar Complementarity}.
}

\section{Dynamic Complementarity Conditions} \label{sec:dcc}

The \emph{Relaxed Complementarity}  \eqref{eq:classical_complementarity} trades off the numerical tractability of the problem with the accuracy of the model. This is particularly important \modtextDani{when \eqref{eq:classical_complementarity} is integrated in optimal control problems for \emph{small} humanoid robots, whose feet little rise up from ground during the swing phases of walking motions.
The normal force allowed by the \emph{relaxed complementarity}  \eqref{eq:classical_complementarity} may have a non-negligible effect on the overall robot  \emph{stability} of these phases.}

This section proposes two novel approaches to describe the complementarity conditions of a rigid contact model, namely

\begin{enumerate}[label=\Alph*),font=\itshape]
	\item Dynamically enforced complementarity;
	\item Hyperbolic secant in velocity bounds.
\end{enumerate}
We call them \emph{Dynamic Complementarity Conditions} (DCCs).

\subsection{Dynamically enforced complementarity}\label{sec:dynamical_complementarity}
The complementarity constraints can be enforced using a Baumgarte stabilization method \cite{baumgarte1972stabilization}. \modtextDani{Thus, we enforce the conditions \eqref{eq:complementarity} so as to obtain convergence to zero. This is achieved by setting the following \emph{dynamical} constraint:}
\begin{equation}
	\frac{\dif}{\dif t} \left(h({}_i \bm{p}) ~\mathbf{n}({}_i\bm{p})^\top{}_i\bm{f}\right) = -K_\text{bs}\left(h({}_i \bm{p}) ~\mathbf{n}({}_i\bm{p})^\top{}_i\bm{f}\right),
\end{equation}
where $K_\text{bs} \in \mathbb{R}^+$ is a positive gain. Hence, the product $h({}_i \bm{p}) ~\mathbf{n}({}_i\bm{p})^\top{}_i\bm{f}$  exponentially decrease\modtextDani{s} to zero at a rate dependent on $K_\text{bs}$. \modtextDani{Direct calculations then lead to:}
\begin{equation}
\begin{split}
	\frac{\dif}{\dif t}\Big(\cdot\Big) =  &\frac{\dif}{\dif t}\left(h({}_i \bm{p})\right)\mathbf{n}({}_i\bm{p})^\top{}_i\bm{f} +\\ &+ h({}_i \bm{p}){}_i\bm{f}^\top \frac{\dif}{\dif t}\left(\mathbf{n}({}_i\bm{p})\right) + h({}_i \bm{p}) ~\mathbf{n}({}_i\bm{p})^\top{}_i\dot{\bm{f}},
\end{split}
\end{equation}
where we \modtextDani{use} the fact that \modtextDani{the left hand side of} \eqref{eq:complementarity} is scalar, \modtextDani{which ease the computations above.}
We can substitute the time derivative of the $h(\cdot)$ and $\mathbf{n}(\cdot)$ functions with the relations:
\begin{IEEEeqnarray}{RCL}
	\phantomsection \IEEEyesnumber
	\frac{\dif}{\dif t}\left(h({}_i \bm{p})\right) &=& \frac{\partial}{\partial {}_i \bm{p}}\left(h({}_i \bm{p})\right){}_i\bm{\dot{p}}, \IEEEyessubnumber \\
	\frac{\dif}{\dif t}\left(\mathbf{n}({}_i\bm{p})\right) &=& \frac{\partial}{\partial {}_i \bm{p}}\left(\mathbf{n}({}_i\bm{p})\right){}_i\bm{\dot{p}}. \IEEEyessubnumber
\end{IEEEeqnarray}

For simplicity, let us define $\zeta$ as follows:
\begin{equation*}
\begin{split}
	\zeta := &\frac{\partial}{\partial {}_i \bm{p}}\left(h({}_i \bm{p})\right){}_i\bm{\dot{p}}~\mathbf{n}({}_i\bm{p})^\top{}_i\bm{f} +\\&+ h({}_i \bm{p}){}_i\bm{f}^\top\frac{\partial}{\partial {}_i \bm{p}}\left(\mathbf{n}({}_i\bm{p})\right){}_i\bm{\dot{p}} + h({}_i \bm{p}) ~\mathbf{n}({}_i\bm{p})^\top{}_i\dot{\bm{f}}.
\end{split}
\end{equation*}
Finally, we obtain the condition
\begin{equation}\label{eq:dynamic_complementarity_equality}
	\zeta = -K_\text{bs}\left(h({}_i \bm{p}) ~\mathbf{n}({}_i\bm{p})^\top{}_i\bm{f}\right).
\end{equation}
In case of planar ground, we have the following relations:
\begin{IEEEeqnarray}{RCLRCL}
    \IEEEyesnumber
	h({}_i \bm{p}) &=& \bm{e}_3^\top {}_i \bm{p},\quad &\frac{\partial}{\partial {}_i \bm{p}}\left(h({}_i \bm{p})\right) &=& \bm{e}_3^\top, \IEEEyessubnumber\\
	\mathbf{n}({}_i\bm{p}) &=& \bm{e}_3, \quad &\frac{\partial}{\partial {}_i \bm{p}}\left(\mathbf{n}({}_i\bm{p})\right) &=& \bm{0}_{3\times 3}. \IEEEyessubnumber	
\end{IEEEeqnarray}
\modtextDani{In other words, the normal vector to the walking plane corresponds to $\bm{e}_3$, while the height $h({}_i \bm{p})$ of the point coincides with the point $z$-coordinate.
Hence, in the planar case, $\zeta$ writes:}
\begin{equation}
	\zeta_\text{planar} = {}_i\dot{\bm{p}}_z \cdot{}_i\bm{f}_z + {}_i\bm{p}_z \cdot {}_i\dot{\bm{f}}_z.
\end{equation}
We can relax Eq. \eqref{eq:dynamic_complementarity_equality} by exploiting again the fact that the product $h({}_i \bm{p}) ~\mathbf{n}({}_i\bm{p})^\top{}_i\bm{f}$ is positive by construction. Henceforth, if we impose \modtextDani{the following}:
\begin{equation}
	\zeta \leq -K_\text{bs}\left(h({}_i \bm{p}) ~\mathbf{n}({}_i\bm{p})^\top{}_i\bm{f}\right),
\end{equation}
we still have exponential convergence, at a rate which is higher or equal to the one specified by $K_\text{bs}$. Finally, similarly to \eqref{eq:classical_complementarity}, we add a further relaxation through a positive number $\varepsilon \in \mathbb{R}^+$ to increase the feasibility region,
\begin{equation}
	\zeta \leq -K_\text{bs}\left(h({}_i \bm{p}) ~\mathbf{n}({}_i\bm{p})^\top{}_i\bm{f}\right) + \varepsilon,
\end{equation}
obtaining the final version of the complementarity condition.

\subsection{Hyperbolic secant in velocity bounds}\label{sec:hyperbolic_secant}
We can impose Eq. \eqref{eq:complementarity} dynamically by enforcing the following set of constraints on the force derivative:
\begin{IEEEeqnarray}{RCLR}
	\phantomsection \IEEEyesnumber \label{eq:force_control_cases}
	-\bm{M}_f \leq &{}_i\dot{\bm{f}} & \leq \bm{M}_f     & \quad \text{if } h({}_i \bm{p}) = 0 \IEEEyessubnumber \label{eq:force_control_bounds}\\
			  &{}_i\dot{\bm{f}} & = -\bm{K}_f {}_i\bm{f} & \quad \text{if } h({}_i \bm{p}) \neq 0 \IEEEyessubnumber \label{eq:force_control_dissipation}
\end{IEEEeqnarray}
meaning that when the point is in contact, ${}_i\dot{\bm{f}}$ is free to take any value in $\left[-\bm{M}_f, \bm{M}_f\right]$ with $\bm{M}_f \in \mathbb{R}^3$ a (non-negative) control bound. On the other hand, if the contact point is not on the walking surface, the control input makes the contact force decreasing exponentially (Eq. \eqref{eq:force_control_dissipation}) at a rate depending on the positive definite control gain $\bm{K}_f \in \mathbb{R}^{3\times 3}$. Defining $\delta^*({}_i\bm{p})$ as a binary function such that
\begin{equation}
	\delta^*({}_i\bm{p}) = 
	\begin{cases}
	1 & \quad \text{if } h({}_i \bm{p}) = 0, \\
	0 & \quad h({}_i \bm{p}) \neq 0,
	\end{cases} 
\end{equation}
it is possible to write  \eqref{eq:force_control_cases} as a set of two inequalities:
\begin{IEEEeqnarray}{RCL}
	\IEEEyesnumber \phantomsection \label{eq:force_control_full}
- \bm{K}_f \left(1 - \delta^*({}_i\bm{p})\right) {}_i\bm{f} - \delta^*({}_i\bm{p})\bm{M}_f &\leq& {}_i\dot{\bm{f}}, \IEEEyessubnumber \label{eq:force_control_lb}\\
- \bm{K}_f \left(1 - \delta^*({}_i\bm{p})\right) {}_i\bm{f} + \delta^*({}_i\bm{p})\bm{M}_f &\geq& {}_i\dot{\bm{f}}. \IEEEyessubnumber \label{eq:hyperbolic_complementarity_lb}
\end{IEEEeqnarray}
Even if $\delta^*({}_i\bm{p})$ would require the adoption of integer variables, it is possible to use a continuous approximation, $\delta({}_i\bm{p})$, namely the hyperbolic secant:
\begin{equation}\label{eq:sech}
    \delta({}_i\bm{p}) = \text{sech}\left(k_h~h({}_i \bm{p})\right),
\end{equation}
where $k_h$ is a user-defined scaling factor. Notice that, when $\delta^*({}_i\bm{p}) = 0$, the bounds coincide and are equal to $-\bm{K}_f {}_i\bm{f}$.

As discussed in Sec. \ref{sec:dynamical_complementarity}, we can simplify the lower bound defined \modtextDani{by}  \eqref{eq:force_control_lb} allowing the force to decrease at a higher rate than \modtextDani{that} given by \eqref{eq:hyperbolic_complementarity_lb}. Hence, we can rewrite \eqref{eq:force_control_full} as:
\begin{equation}\label{eq:force_control_final}
-\bm{M}_f \leq {}_i\dot{\bm{f}} \leq - \bm{K}_f \left(1 - \delta({}_i\bm{p})\right) {}_i\bm{f} + \delta({}_i\bm{p})\bm{M}_f.
\end{equation}
Given Eq. \eqref{eq:friction}, it is enough to apply any of these equations only to the force component normal to the ground: if it decreases to zero, also planar components have to vanish to satisfy friction constraints. Hence, we can \modtextDani{reduce} \eqref{eq:force_control_final} \modtextDani{to}:
\begin{IEEEeqnarray}{RCL}
	\IEEEyesnumber \phantomsection
	-\bm{e}_3^\top\bm{M}_f &\leq& \bm{e}_3^\top{}_i\dot{\bm{f}} \IEEEyessubnumber \\
	- K_{f,z} \left(1 - \delta({}_i\bm{p})\right) \mathbf{n}({}_i\bm{p})^\top {}_i\bm{f} + \delta({}_i\bm{p})\bm{e}_3^\top\bm{M}_f &\geq& \bm{e}_3^\top{}_i\dot{\bm{f}},\IEEEeqnarraynumspace\IEEEyessubnumber \label{someUsefulStaffFdot}
\end{IEEEeqnarray}
with $K_{f,z}$ the corresponding element of $\bm{K}_f$.

\subsection{Summary on complementarity conditions} \label{sec:complementarity_list}
We \modtextDani{have presented} different methods for \modtextDani{enforcing} the complementarity constraints \modtextDani{given by} Eq. \eqref{eq:complementarity}\modtextDani{, namely}:
\begin{itemize}
	\item $h({}_i \bm{p}) ~\mathbf{n}({}_i\bm{p})^\top{}_i\bm{f} \leq \epsilon$;
	\item $\zeta \leq -K_\text{bs}\left(h({}_i \bm{p}) ~\mathbf{n}({}_i\bm{p})^\top{}_i\bm{f}\right) + \varepsilon$;
	\item $-\bm{e}_3^\top\bm{M}_f \leq \bm{e}_3^\top{}_i\dot{\bm{f}} \leq - K_{f,z} \left(1 - \delta({}_i\bm{p})\right) \mathbf{n}({}_i\bm{p})^\top {}_i\bm{f} + \delta({}_i\bm{p})\bm{e}_3^\top\bm{M}_f$.
\end{itemize}

It is also \modtextDani{pragmatic} to assume the force derivative to be bounded\modtextDani{\footnote{\modtextDani{Note that this is not necessary if one uses \eqref{eq:hyperbolic_complementarity_lb} instead of \eqref{someUsefulStaffFdot} since the bounds are already included in the constraint.}}, i.e. $-\bm{M}_f \leq {}_i\dot{\bm{f}} \leq \bm{M}_f$.
Observe that all the conditions presented in the above bullet points do not depend on the type of ground, which is here considered rigid.} The parameters involved do not have a direct physical meaning (like in compliant contact models), but rather determine the ``accuracy'' of the simulated behavior.

\section{DCCs-Based Non-Linear Trajectory Planning}\label{sec:dcc_nlmpc}

\modtextDani{We detail below} how the \emph{Dynamic Complementarity Conditions} presented in Sec. \ref{sec:dcc} can be \modtextDani{used} in a non-linear trajectory planning framework aimed at generating walking trajectories. 

\subsection{Kinematic Control}

First of all, we assume to have full control over the derivative of the contact point's positions and forces:
\begin{IEEEeqnarray}{RCL}
	\phantomsection \IEEEyesnumber \label{eq:force_velocity_control}
	{}_i\dot{\bm{p}} &=& \bm{u}_{{}_ip}, \IEEEyessubnumber \label{eq:velocity_control}\\
	{}_i\dot{\bm{f}} &=& \bm{u}_{{}_if}, \IEEEyessubnumber
\end{IEEEeqnarray}
where $\bm{u}_{{}_ip}, ~ \bm{u}_{{}_if} \in \mathbb{R}^3$ are the control inputs for the $i$-th contact point. 
While each contact point is supposed to be independent from the control point of view, they all need to remain on the same surface and maintain a constant relative distance\modtextDani{: the contact points} belong to the same rigid body. At the same time, we want them to be within the workspace reachable by the robot legs. We can achieve both objectives with the following algebraic condition acting on each contact point:
\begin{equation}\label{eq:dp_point_consistency}
	{}_i\bm{p} = {}^\mathcal{I} \bm{H}_\text{foot} {}^\text{foot}{}_i \bm{p},
\end{equation} 
where ${}^\text{foot}{}_i \bm{p}$ is the (fixed) position of the contact point within the foot surface, expressed in foot coordinates. 
Here, the transformation matrix ${}^\mathcal{I} \bm{H}_\text{foot}$ would depend on $\bm{q}$, including the joint configuration $\bm{s}$. As a consequence, the full kinematics of the robot is taken into \modtextDani{account}.

\modtextDani{Concerning the robot controls, we assume that the joint values can be assigned at will. In  the language of Automatic Control and thanks to the so-called backstepping hypothesis \cite[p. 589]{khalil2002}, full joint controllability is essentially equivalent to assuming  the joint velocities as a control input, namely:}
\begin{equation}
	\dot{\bm{s}} = \bm{u}_s.
\end{equation}
This assumption 
\modtextDani{requires   
an additional control loop: joint velocities, together with contact vertex positions and forces, are considered as references to an inner whole-body controller.}

Finally, the base rotation included in $\bm{q}$ is vectorized using the quaternion parametrization. The corresponding unitary quaternion is called ${}^\mathcal{I}\bm{\rho}_B \in \mathbb{H}$. The base position is indicated with the symbol ${}^\mathcal{I}\bm{p}_B \in \mathbb{R}^3$. The equations governing the dynamical evolution of the base are as follows:
\begin{IEEEeqnarray}{RCL}
	\IEEEyesnumber \phantomsection
	{}^\mathcal{I}\dot{\bm{p}}_B &=& {}^\mathcal{I}\bm{R}_B {}^B\bm{v}_{\mathcal{I},B} ,\IEEEyessubnumber \label{eq:quaternionLeftDer}\\
	{}^\mathcal{I}\dot{\bm{\rho}}_B &=& \bm{u}_\rho. \quad  \IEEEyessubnumber \label{eq:quaternionRotationDerivative}
\end{IEEEeqnarray}
${}^B\bm{v}_{\mathcal{I},B} \in \mathbb{R}^3$, $\bm{u}_\rho \in \mathbb{R}^4$ and $\bm{u}_s \in \mathbb{R}^n$ are control inputs, defining the base linear velocity, the quaternion derivative and the joints velocity, respectively. More specifically, ${}^B\bm{v}_{\mathcal{I},B}$ is the linear part of ${}^B\bm{V}_{\mathcal{I},B} \in \mathbb{R}^6$, the \emph{left-trivialized} (i.e. measured in body coordinates) base velocity. \modtextDani{The control input $\bm{u}_\rho$ is such that the base quaternion remains unitary over time.  }

\modtext{\subsection{Planar DCC}}
\label{Planar-DCC-subsection}
\modtext{As \modtextDani{mentioned} in Sec. \ref{sec:complementarity_intro}, we need to prevent the contact points to move on the walking plane when \modtextDani{they are} in contact \modtextDani{with the ground}. Instead of applying the \emph{Relaxed Planar Complementarity} \modtextDani{\eqref{eq:classicalPlanarComplementarity}}, we achieve the same result} by limiting the effect of the control input $\bm{u}_{{}_ip}$ along the planar components:
\begin{equation}\label{eq:planarControl}
	\mathbf{t}({}_i\bm{p})^\top {}_i \dot{\bm{p}} = \tanh\left(k_t~h({}_i \bm{p})\right) \left[\bm{e}_1 ~ \bm{e}_2\right]^\top \bm{u}_{{}_ip},
\end{equation}
where $k_t \in \mathbb{R}$ is a user-defined scaling factor. Eq. \eqref{eq:planarControl} \modtextDani{projects the control input $\bm{u}_{{}_ip}$} along the planar directions, \modtextDani{and forces these projections} to zero when $h({}_i\bm{p})$ is null\modtextDani{. A}t the same time, Eq. \eqref{eq:planarControl} reduces the velocity when the contact point is approaching the ground. \modtextDani{So,} \modtext{Eq. \eqref{eq:planarControl} can then be used to substitute Eq. \eqref{eq:velocity_control} as follows:
\begin{equation}
\label{eq:proposedPlanarComplementarity}
{}_i\dot{\bm{p}} =  \bm{\tau}({}_i\bm{p}) \bm{u}_{{}_ip},
\end{equation} where $\bm{\tau}({}_i\bm{p})$ is defined as}
\begin{equation}
	\bm{\tau}({}_i\bm{p}) ={}^\mathcal{I}\bm{R}_{plane} ~ diag\left(\begin{bmatrix}
		\tanh\left(k_t ~ h({}_i \bm{p})\right) \\
		\tanh\left(k_t ~ h({}_i \bm{p})\right) \\
		1
	\end{bmatrix}\right).
\end{equation}
Note that, from now on, $\bm{u}_{{}_ip}$ is assumed to be \modtextDani{expressed in the} $plane$ coordinates. Thus, the normal component of the velocity is directly \modtextDani{regulated} by $\bm{e}_3^\top \bm{u}_{{}_ip}$. Also, it is necessary to bound this control input, $\bm{u}_{{}_ip} \in \left[-\bm{M}_V, \bm{M}_V\right], \bm{M}_V \in \mathbb{R}^3$, to properly exploit the effect of the hyperbolic tangent.

\modtext{
Compared to the \emph{Relaxed Planar Complementarity} \modtextDani{\eqref{eq:classicalPlanarComplementarity}, the advantages provided by \eqref{eq:proposedPlanarComplementarity} are twofold. $i)$ The proposed DCC Planar Complementarity~\eqref{eq:proposedPlanarComplementarity}} naturally limits the point velocity to zero when approaching the contact surface. $ii)$~Through the parameter $k_t$, it is possible to control the height of the contact points when swinging. The smaller the $k_t$, the higher is the resulting swing height.
}
\subsection{Momentum control}
In Sec. \ref{sec:complementarity_intro}, we \modtextDani{assumed that the contact points can exert any force on the environment. Let us now} describe the effect of these forces on the robot motion through the centroidal dynamics introduced in Sec. \ref{sec:intro_momentum}, \modtextDani{which writes:} 
\begin{equation}
    \label{eq:centroidal_momentum_dynamics_proposed}
	{}_{\bar{G}} \dot{\bm{h}} = m\bar{\bm{g}} + \sum_i
	\begin{bmatrix}
		\bm{\mathds{1}}_3 \\
		({}_i\bm{p} - \bm{x})^\wedge
	\end{bmatrix} {}_i\bm{f}.
\end{equation}

Compared to Eq. \eqref{eq:centroidal_momentum_dynamics}, \modtextDani{Eq. \eqref{eq:centroidal_momentum_dynamics_proposed} reduces} the matrix ${}_{\bar{G}}\bm{X}^i$ since no torque is applied at the contact points.
We also need to make sure that the CoM position obtained by integrating Eq.~\eqref{eq:com_from_momentum} corresponds to the one obtained via the joints variables. This is done through the following algebraic equation:
\begin{equation}\label{eq:comConsistency}
	\bm{x} = \text{CoM}({}^\mathcal{I}\bm{p}_B, {}\mathcal{I}\bm{\rho}_B, s),
\end{equation} 
where $\text{CoM}({}^\mathcal{I}\bm{p}_B, {}\mathcal{I}\bm{\rho}_B, s)$ is the function mapping base pose and joints position to the CoM position, i.e. the right hand side of Eq. \eqref{eq:com_definition}. \modtextDani{Eq. \eqref{eq:comConsistency} links the linear momentum and the joint variables, so we also need to link the angular momentum and
the joints evolution. To do so, we use the Centroidal Momentum Matrix $\bm{J}_\text{CMM}$,
thus} obtaining:
\begin{equation}\label{constr:CMM}
	{}_{\bar{G}} \bm{h}^\omega = \left[\bm{0}_{3 \times 3} ~ \bm{\mathds{1}}_3\right]\bm{J}_\text{CMM} \bm{\nu} \in \mathbb{R}^3.
\end{equation}

The system velocity $\bm{\nu}$ contains the base angular velocity ${}^B \bm{\omega}_{\mathcal{I},B}$. It can be substituted with the quaternion derivative through the map $\bm{\mathcal{G}}$. It is defined in \cite[Section 1.5.4]{graf2008quaternions} as 
\begin{equation} \label{eq:g_map_quaternion} 
	\bm{\mathcal{G}}({}\mathcal{I}\bm{\rho}_B) = \begin{bmatrix} {-{}^\mathcal{I}\rho_{B,1}} & {{}^\mathcal{I}\rho_{B,0}} & {{}^\mathcal{I}\rho_{B,3}} & {-{}^\mathcal{I}\rho_{B,2}} \\ {-{}^\mathcal{I}\rho_{B,2}} & {-{}^\mathcal{I}\rho_{B,3}} & {{}^\mathcal{I}\rho_{B,0}} & {{}^\mathcal{I}\rho_{B,1}} \\ {-{}^\mathcal{I}\rho_{B,3}} & {{}^\mathcal{I}\rho_{B,2}} & {-{}^\mathcal{I}\rho_{B,1}} & {{}^\mathcal{I}\rho_{B, 0}} \end{bmatrix}
\end{equation}
such that 
\begin{equation} \label{eq:base_angular_velocity}
	{}^B \bm{\omega}_{\mathcal{I},B} = 2 \bm{\mathcal{G}}({}^\mathcal{I}\rho_B)\bm{u}_\rho.
\end{equation}
Hence, it depends on the same control input of Eq. \eqref{eq:quaternionRotationDerivative}.

\subsection{The complete differential-algebraic system of equations} \label{sec:complete_dae}
By summarizing all the ODEs and algebraic conditions, we obtain the following inequality constrained DAE. \\
$\bullet$ Dynamical Constraints 
\begin{IEEEeqnarray}{RCLL}
	\IEEEyesnumber \phantomsection \label{constr:system_dynamics}
	{}_i\dot{\bm{f}} &=& \bm{u}_{{}_if}, & \forall \text{ contact point}, \IEEEyessubnumber \label{constr:force_derivative}\\
	{}_i\dot{\bm{p}} &=& \bm{\tau}({}_i\bm{p}) \bm{u}_{{}_ip}, & \forall \text{ contact point},  \IEEEyessubnumber \label{constr:position_derivative}\\
	{}_{\bar{G}} \dot{\bm{h}} &=& \IEEEeqnarraymulticol{2}{L}{m\bar{\bm{g}} + \sum_i
		\begin{bmatrix}
			\bm{\mathds{1}}_3 \\
			({}_i\bm{p} - \bm{x})^\wedge
		\end{bmatrix} {{}_i\bm{f}}}, \IEEEyessubnumber\label{constr:momentum_derivative}\\	
	\dot{\bm{x}} &=& \IEEEeqnarraymulticol{2}{L}{\frac{1}{m}  \left({}_{\bar{G}} \bm{h}^p\right),}  \IEEEyessubnumber \label{constr:com_derivative}\\
	{}^\mathcal{I}\dot{\bm{p}}_B &=& {}^\mathcal{I}\bm{R}_B {}^B\bm{v}_{\mathcal{I},B}, \IEEEyessubnumber \label{constr:basePosDerivative}\\
	{}^\mathcal{I}\dot{\bm{\rho}}_B &=& \bm{u}_\rho, \IEEEyessubnumber \label{constr:quatDerivative}\\
	\dot{\bm{s}} &=& \bm{u}_s. \IEEEyessubnumber
\end{IEEEeqnarray}
$\bullet$ Equality Constraints 
\begin{IEEEeqnarray}{RCL}
	\IEEEyesnumber \phantomsection \label{constr:equalities}
	{}_i\bm{p} &=& {}^A \bm{H}_\text{foot} {}^\text{foot}{}_i \bm{p}, \quad \forall \text{ contact point}, \IEEEyessubnumber \label{constr:pointPositions}\\
	\bm{x} &=& \text{CoM}({}^\mathcal{I}\bm{p}_B, {}^\mathcal{I}\bm{\rho}_B, \bm{s}), \IEEEyessubnumber \\
	{}_{\bar{G}} \bm{h}^\omega &=& \left[\bm{0}_{3 \times 3} ~ \mathds{1}_3\right]\bm{J}_\text{CMM} \begin{bmatrix}
		{}^B\bm{v}_{\mathcal{I},B} \\
		2 \mathcal{\bm{G}}({}^\mathcal{I}\bm{\rho}_B)\bm{u}_\rho\\
		\bm{u}_s
	\end{bmatrix}, \IEEEyessubnumber  \\
	\|{}^\mathcal{I}\bm{\rho}_B\|^2 &=& 1. \IEEEyessubnumber \label{eq:quatNorm}
\end{IEEEeqnarray}
$\bullet$ Inequality Constraints, applied for each contact point
\begin{IEEEeqnarray}{RCL}
	\IEEEyesnumber \phantomsection \label{constr:inequalities}
	\mathbf{n}({}_i\bm{p})^\top{}_i\bm{f} &\geq& 0, \IEEEyessubnumber\\
	\|\mathbf{t}({}_i\bm{p})^\top {}_i\bm{f} \| &\leq& \mu_s ~\mathbf{n}({}_i\bm{p})^\top{}_i\bm{f}, \IEEEyessubnumber\\
	-\bm{M}_V &\leq& \bm{u}_{{}_ip} \leq \bm{M}_V, \IEEEyessubnumber\\
	-\bm{M}_f &\leq& \bm{u}_{{}_if} \leq \bm{M}_f ,\IEEEyessubnumber\\
	h({}_i \bm{p}) &\geq& 0, \IEEEyessubnumber\\
	\IEEEeqnarraymulticol{3}{C}{\text{Complementarity, see Sec. \ref{sec:complementarity_list}}}.\IEEEyessubnumber
\end{IEEEeqnarray}


\subsection{Walking specific constraints} \label{sec:additional_constraints}
While \modtextDani{taking a step}, the robot legs \modtextDani{do not have to} collide with each other. Self collisions constraints are usually hard to consider and may slow down consistently the determination of a solution. A simpler solution to avoid self-collisions between legs consists of avoiding cross-steps. We assume the frame attached to the right foot to have the positive $y-$direction pointing toward left. In this case, it would be sufficient to impose the $y-$component of the ${}^{r}\bm{x}_{l}$ (i.e. the relative position of the left foot expressed in the right foot frame) to be greater than a given quantity, i.e.:
\begin{equation}
\bm{e}_2^\top {}^{r}\bm{x}_{l} \geq d_\text{min}.
\end{equation}
Too wide motions of the swing leg may cause other self-collision, especially between the \modtextDani{robot} arms and thigh\modtextDani{s}. Hence, we set an upper-bound on the difference between the height of the two feet. To simplify the definition of the constraint, we consider the mean position of every contact point:
\begin{equation}
-M_{hf} \leq e_3^\top\left({}_{\#}\bm{p}_l - {}_{\#}\bm{p}_r\right) \leq M_{hf},
\end{equation}
where ${}_{\#}\bm{p}_l$ and ${}_{\#}\bm{p}_r$ are the mean positions of all the contact points of the left and right feet respectively, i.e. ${}_\# \bm{p}_\circ =\frac{1}{n_p} \sum^{n_p}_i {}_i\bm{p}_\circ$. The number of contact points in a single foot is $n_p$, and $M_{hf} \in \mathbb{R}^+$ is the constraint upper-bound. 

Some additional constraints can be considered:
\begin{IEEEeqnarray}{RCL}
    \label{contraintsFalling}
	\IEEEyesnumber
	x_{z \text{ min}} &\leq& h(\bm{x}),  \IEEEyessubnumber \label{constr:com_height_limit}\\
	-\bm{M}_{h_\omega} &\leq& {}_{\bar{G}} \bm{h}^\omega \leq \bm{M}_{h_\omega}. \IEEEyessubnumber \label{constr:angular_momentum_bounds}
\end{IEEEeqnarray}
Eq. \eqref{constr:com_height_limit} avoids \modtextDani{the emergence of} solutions that set the CoM position too close or even below the ground. Eq. \eqref{constr:angular_momentum_bounds} set\modtextDani{s} a bound $\bm{M}_{h_\omega} \in \mathbb{R}^3$ on the angular momentum. \modtextDani{When considered together, Eq. \eqref{contraintsFalling} tend to avoid the emergence of trajectories that  cause excessive motions or let the robot fall.}

\subsection{Tasks in Cartesian space} \label{sec:cartesian_tasks}
In order to make the robot move toward a desired position, it is necessary to specify tasks in Cartesian space. 
\subsubsection{Contact point centroid position task}\label{sec:centroid_task}
We define as a task the L2 norm of the error between a point attached to the robot and its desired position in an absolute frame. Suppose we choose the CoM position as a target point. By moving its desired value forward in space, the robot could simply lean toward the goal without moving the feet. This undesired behavior may lead the robot to fall. It is possible to avoid the robot leaning forward by locating the target point on the feet instead of the CoM. In particular, we select the centroid of the \modtextDani{feet} contact points as target, thus avoiding specifying a desired placement for each foot:
\begin{equation}
\Gamma_{{}_\# p} = \frac{1}{2} \|{}_\# \bm{p} - {}_\# \bm{p}^*\|^2_{\bm{W}_\#},
\end{equation}
where ${}_\# \bm{p} =\frac{1}{2} ({}_{\#}\bm{p}_l + {}_{\#}\bm{p}_r)$ and ${}_\# \bm{p}^* \in \mathbb{R}^2$ is its desired value. 

\subsubsection{CoM linear velocity task}\label{sec:com_velocity_cost}
While walking, we want the robot to keep \modtextDani{(an approximately)} constant forward motion. In fact, since foot positions are not scripted, it may be possible to plan two consecutive steps with the same foot. Requiring a constant forward velocity help\modtextDani{s} avoiding such phenomena. This task can be defined as:
\begin{equation}\label{eq:momentum_cost}
\Gamma_{{}_{\bar{G}} h^p} = \frac{1}{2} \|{}_{\bar{G}} \bm{h}^p - m  \dot{\bm{x}}^*\|^2_{\bm{W}_v}
\end{equation}
with $\dot{\bm{x}}^*$ a desired CoM velocity. The weights $\bm{W}_v$ allows selecting and weighting the different directions separately.

\subsubsection{Foot yaw task}
The task on the centroid of the contact points defined in Sec.~\ref{sec:centroid_task} \modtextDani{allows us to specify} the direction the robot has to step \modtextDani{towards. The \emph{foot yaw task} specifies} at which angle the foot should be oriented with respect to the $z$-axis, i.e. the foot yaw angle. Define $\gamma^*_\circ$ as the desired yaw angle for either the left or the right foot ($\circ$ is a placeholder). We construct a unitary vector $\bm{\ell}^*_{\circ} \in \mathbb{R}^2$ belonging to the $xy$-plane (of $\mathcal{I}$), oriented such that the angle with the $x$-axis of $\mathcal{I}$ corresponds to  $\gamma^*_\circ$. Its components are: 
\begin{equation}
\bm{\ell}^*_\circ = \begin{bmatrix}
\cos(\gamma^*_\circ) \\
\sin(\gamma^*_\circ)
\end{bmatrix}.
\end{equation}
Similarly, the vector $\bm{\ell}_{\circ} \in \mathbb{R}^2$ is fixed to the foot and parallel to the foot $x$-axis, but expressed in the $\mathcal{I}$ frame. This vector can be easily obtained as a linear combination of the contact points position. The goal of \modtextDani{the \emph{foot yaw task}} is to align $\bm{\ell}_{\circ}$ to $\bm{\ell}^*_{\circ}$\modtextDani{, which can be (locally) achieved by minimizing their cross-product, thus leading to}:
\begin{equation}\label{eq:yaw_task_partial}
\Gamma^{\prime}_\text{yaw} = \sum_{l, r} \frac{1}{2} \left\| \begin{bmatrix}
-\sin(\gamma^*_\circ) & \cos(\gamma^*_\circ)
\end{bmatrix}\bm{\ell}_{\circ} \right\|^2.
\end{equation}
Notice that Eq. \eqref{eq:yaw_task_partial} has a minimum also when $\bm{\ell}_{\circ}$ is null. In other words, $\Gamma^{\prime}_\text{yaw}$ can be minimized by shrinking the projection on the $xy$-plane of the vector attached to the robot foot. This is undesired because it would set the foot to be perpendicular to the ground. Hence, we consider also a second vector attached to the foot and perpendicular to $\bm{\ell}_{\circ}$, called $\bm{\ell}^\bot_{\circ}$. This vector is parallel, and has the same direction of the foot $y$-axis. Hence, the final task has the following form:
\begin{equation} \label{eq:yaw_task}
\begin{split}
\Gamma_\text{yaw} = &\sum_{l, r} \frac{1}{2} \left\| \begin{bmatrix}
-\sin(\gamma^*_\circ) & \cos(\gamma^*_\circ)
\end{bmatrix}\bm{\ell}_{\circ} \right\|^2 \\
+ &\sum_{l, r} \frac{1}{2} \left\| \begin{bmatrix}
\cos(\gamma^*_\circ) & \sin(\gamma^*_\circ)
\end{bmatrix}\bm{\ell}^\bot_{\circ} \right\|^2.
\end{split}
\end{equation}
Eq. \eqref{eq:yaw_task} does not prevent the foot to have roll and pitch motions during the swing phase.

\subsection{Regularization tasks}\label{sec:regularization_tasks}
The dynamical system \modtextDani{in Sec.} \eqref{sec:complete_dae} depends on a high number of variables. Despite the additional constraints \modtextDani{defined in} Sec. \ref{sec:additional_constraints} and the Cartesian tasks \modtextDani{presented in} Sec. \ref{sec:cartesian_tasks}, a consistent part of the \modtextDani{system} dynamics is not taken into \modtextDani{account} nor constrained. For this reason, it is necessary to introduce regularization tasks that contribute in generating walking trajectories.

\subsubsection{Frame orientation task}\label{sec:orientation_task}
While moving, we want a robot frame to \modtextDani{take a desired} orientation ${}^\mathcal{I}\bm{R}^*_\text{frame}$. We weight the distance of the rotation matrix ${}^\mathcal{I}\tilde{\bm{R}}_\text{frame} = {}^\mathcal{I}\bm{R}^{*\top}_\text{frame}{}^\mathcal{I}\bm{R}_\text{frame}$ from the identity. To this end, we convert ${}^\mathcal{I}\tilde{\bm{R}}_\text{frame}$ into a quaternion (through a function \texttt{quat}, that implements the Rodrigues formula\modtext{, while keeping the real part always non-negative}), and weight its difference from the identity quaternion $\bm{I}_q$, thus exploiting the Euclidean distance of quaternions as a metric for rotation error \cite{ravani1983motion}:
\begin{equation}\label{cost:frameOrientation}
\Gamma_\text{frame} = \frac{1}{2}\left\|\texttt{quat}\left({}^A\tilde{\bm{R}}_\text{frame}\right) - \bm{I}_q\right\|^2.
\end{equation}
It can be applied on multiple bodies, like the torso and waist.

\modtext{
\subsubsection{Base quaternion derivative regularization task}\label{sec:baseQuatVel_task}
The base quaternion derivative task allows tracking a desired angular velocity for the base. During walking, this tasks helps in regularizing the robot motion. It is defined as follows:
\begin{equation}
    \Gamma_{\text{reg} u_\rho} = \frac{1}{2}\left\|\bm{u}_\rho - \bm{u}_\rho^*\right\|^2
\end{equation}
with $\bm{u}_\rho^* \in \mathbb{R}^4$ a desired quaternion rate of change.
}

\subsubsection{Force regularization task} \label{sec:forceRegularization}
While considering \modtextDani{foot} contact forces independent \modtextDani{from each other}, they still belong to a single \modtextDani{rigid} body. Thus, we prescribe the contact forces in a foot to be as similar as possible, refraining from using partial contacts if not strictly necessary. This can be obtained through the following:
\begin{equation}\label{cost:forceRegularization}
\Gamma_{\text{reg} f} = \sum_{l, r}\sum^{n_p}_i \frac{1}{2} \left\|{}_i\bm{f} - \modtextDani{\text{diag}}({}_i\bm{\alpha}^*)\sum^{n_p}_j {}_j\bm{f}\right\|^2_{\bm{W}_f}.
\end{equation}
Here ${}_i\bm{\alpha}^* \in \mathbb{R}^3$ determines the desired ratio for force $i$ with respect to the total force. For example, if we want all the forces in a foot to be equal, it is sufficient to select all the components of ${}_i\bm{\alpha}^* \in \mathbb{R}^3$ equal to $\frac{1}{n_p}$. In this case, the corresponding CoP is the centroid ${}_\# \bm{p}$. In other cases, it may be helpful to move the CoP somewhere else in the foot. In this case\modtextDani{, it is} sufficient to compute the corresponding ${}_i\bm{\alpha}^*$.

\subsubsection{Joint regularization task} \label{sec:regularization}
In order to avoid solutions with huge joint variations, we can introduce a regularization task for the joint variables:
\begin{equation}\label{cost:jointsRegularization}
\Gamma_{\text{reg} s} = \frac{1}{2}\left\|\dot{\bm{s}} + \bm{K}_s(\bm{s} - \bm{s}^*)\right\|^2_{\bm{W}_j},
\end{equation}
with $\bm{s}^*$ a desired joints configuration. The minimum for this cost is achieved when $\dot{\bm{s}}= -\bm{K}_s(\bm{s} - \bm{s}^*)$, with $\bm{K}_s \in \mathbb{R}^{n\times n}$ \modtextDani{a positive semi-definite matrix}. When this equality holds, joint values converge exponentially to their desired values $\bm{s}^*$. In this way, both joint velocities and joint positions are regularized.

\subsubsection{Swing height task}
When performing a step, the swing foot \modtextDani{height} usually ensures some level of robustness with respect to ground asperity.
In order to specify a desired swing height, we can \modtextDani{define} the following task:
\begin{equation}
    \label{heightTask}
	\Gamma_{\text{swing}} = \sum_{l, r}\sum^{n_p}_i\frac{1}{2}\left(\bm{e}_3^\top {}_i\bm{p} - {}_sh^*\right)^2\left\|\left[\bm{e}_1 ~ \bm{e}_2\right]^\top \bm{u}_{{}_ip}\right\|^2.
\end{equation}
\modtextDani{Eq. \eqref{heightTask}} penalizes the distance between the $z$-component of each contact point position from a desired height ${}_sh^* \in \mathbb{R}$ when the corresponding planar velocity is \modtextDani{different from zero}. Trivially, this cost has two minima: when the planar velocity is zero (thus the point is not moving) or when the height of the points is equal to the desired one.

\modtext{
\subsubsection{Contact Control Regularization Tasks}
Even if we assume the contact point velocities and force derivatives to be control inputs, we want to avoid the planner to use them always at the limit. Hence, we add some basic regularization tasks:
\begin{IEEEeqnarray}{RCL}
\Gamma_{\text{reg} u_p} &=& \sum_{l, r}\sum^{n_p}_i\frac{1}{2}\left\|\bm{u}_{{}_ip} - \bm{u}_{{}_ip}^* \right\|^2_{\bm{W}_{u_p}}, \\
\Gamma_{\text{reg} u_f} &=& \sum_{l, r}\sum^{n_p}_i\frac{1}{2}\left\|\bm{u}_{{}_if} - \bm{u}_{{}_if}^* \right\|^2_{\bm{W}_{u_f}},
\end{IEEEeqnarray}
with $\bm{u}_{{}_ip}^*, \bm{u}_{{}_if}^* \in \mathbb{R}^3$ the corresponding desired quantities. 
}

\subsection{The complete optimal control problem} \label{sec:oc}
Given the set of equations listed in Section \ref{sec:complete_dae} and the tasks described in Section \ref{sec:cartesian_tasks} and \ref{sec:regularization_tasks}\modtextDani{,} it is possible to \modtextDani{define} an optimal control problem whose complete formulation is presented below. Here the vector $\mathbf{w}$ contains the set of weights defining the relative ``importance'' of each task.

\begin{IEEEeqnarray}{CRCLL}
	\IEEEyesnumber \phantomsection
	\minimize_{\bm{\chi},\, \bm{\mathcal{U}}} & \IEEEeqnarraymulticol{4}{C}{\mathbf{w}^\top 
	\modtext{
	\begin{bmatrix*}[l]
		\Gamma_{{}_\# p} \\
		\Gamma_{{}_{\bar{G}} h^p}\\
		\Gamma_\text{frame} \\
		\Gamma_{\text{reg} u_\rho} \\
		\Gamma_{\text{reg} f} \\
		\Gamma_{\text{reg} s} \\
		\Gamma_{\text{swing}} \\
		\Gamma_\text{yaw} \\
		\Gamma_{\text{reg} u_p} \\
		\Gamma_{\text{reg} u_f}
	\end{bmatrix*}, }\label{costFunction}} \IEEEyessubnumber\\
	\text{subject to:}&  \nonumber\\
	\IEEEeqnarraymulticol{2}{R}{\dot{\bm{\chi}}} &=& \bm{f}(\bm{\chi}, \bm{\mathcal{U}}), & \text{ see Eq. \eqref{constr:system_dynamics}}, \IEEEyessubnumber\\
	\IEEEeqnarraymulticol{2}{R}{\mathbf{l}} &\leq& \bm{g}\left(\bm{\chi}, \bm{\mathcal{U}}\right) \leq \mathbf{u},\,\, &\text{see Eq.s \eqref{constr:equalities}${-}$\eqref{constr:inequalities}},\IEEEeqnarraynumspace \IEEEyessubnumber\\
	\IEEEeqnarraymulticol{2}{R}{\bm{e}_2^\top{{}^{r}\bm{x}_{l}}} &\geq& d_\text{min},~ \IEEEyessubnumber \label{constr:minDistance}\\
	\IEEEeqnarraymulticol{2}{R}{x_{z \text{ min}}} &\leq& \bm{e}_3^\top{\bm{x}},  \IEEEyessubnumber\\
	\IEEEeqnarraymulticol{2}{R}{-\bm{M}_{h_\omega}} &\leq& {}_{\bar{G}} \bm{h}^\omega \leq \bm{M}_{h_\omega}, \IEEEyessubnumber\\
	\IEEEeqnarraymulticol{2}{R}{-M_{hf}} &\leq& \IEEEeqnarraymulticol{2}{L}{e_3^\top\left({\#}\bm{p}_l - {\#}\bm{p}_r\right) \leq M_{hf}.} \IEEEyessubnumber
\end{IEEEeqnarray}
Here, the state variables $\bm{\mathcal{X}}$ are those derived in time, while $\bm{\mathcal{U}}$ contains all the control inputs. Thus:
\begin{equation} \label{eq:dp_state_control}
\bm{\chi} = 
\begin{bmatrix}
{}_i\bm{f} \\
{}_i\bm{p} \\
\vdots	\\
{}_{\bar{G}} \bm{h} \\
\bm{x} \\
{}^\mathcal{I} \bm{p}_B \\
{}^\mathcal{I}\bm{\rho}_B \\
\bm{s}
\end{bmatrix}, \quad
\bm{\mathcal{U}} = 
\begin{bmatrix}
\bm{u}_{{}_if} \\
\bm{u}_{{}_ip} \\
\vdots \\
{}^B\bm{v}_{\mathcal{I},B} \\
\bm{u}_\rho \\
\bm{u}_s
\end{bmatrix},
\end{equation}
where the symbol $\vdots$ represents the repetition of the corresponding variables for each contact point. 

\subsection{Considerations} \label{sec:considerations}
The optimal control problem \modtextDani{presented} in Sec. \ref{sec:oc} is designed such that (almost) no constraint is task specific. As a consequence, it is particularly important to define the cost function carefully since the solution \modtext{is} a trade-off between all the various tasks. On the other hand, the detailed model of the system allows \modtextDani{us to achieve} walking motions without specifying a desired CoM trajectory or by fixing the angular momentum to zero.
Nevertheless, due to the limited time horizon, it is better to prevent the solver from finding solutions that would lead to unfeasible states in future planner \modtextBis{runs}. For this reason, Eq. \eqref{constr:com_height_limit} and Eq. \eqref{constr:angular_momentum_bounds} are added.

\modtext{In order to plan for trajectories longer than the prediction horizon, we apply the Receding Horizon Principle \cite{Mayne90MPC}. Hence the planner is called iteratively, initialized with the results from the previous \modtextBis{run}. A}
possible effect resulting from the application of the Receding Horizon principle is the emergence of ``procrastination'' phenomena. Due to the moving horizon, the solver may continuously delay in actuating motions, since the task keeps being shifted in time. A simple fix to this phenomena is to increase the task weights with time, such that it is more convenient to reach a goal position earlier.

\modtext{
In addition, since the optimal control problem described in Sec. \ref{sec:oc} uses a quaternion as parametrization for the base rotation, it is necessary to preserve its unit norm. Since this problem is solved using iterative non-linear optimization solvers, the enforcement of Eq. \eqref{eq:quatNorm} is not enough, as the solver might perform intermediate unfeasible iterations. Hence, we normalize the base quaternion every time we evaluate the costs and constraints. 
}

Finally, given that the problem under consideration is non-convex, the optimizer will find a local minimum. This may result in a sub-optimal solution for the given tasks, but this fact does not limit the applicability of the results to the robot. \modtext{In this context\modtextDani{, a proper problem initialization plays a pivotal role.} For the first \modtextBis{run} only, the planner is initialized by setting the force on the left foot to be null for the entire horizon. The position of the left foot is such that the centroid of the contact points is on the desired position. In this way, we hint the optimizer to use the left foot for the first step. In successive \modtextBis{runs}, the solver is warm-started with the solution computed in the previous run. Even if it affects only the first \modtextBis{run}, the initial guess can strongly influence the generated motion.}

\modtextBis{\section{Validation} \label{sec:validation}}

\begin{figure}[tpb]
    \centering
    \subfloat[$t=0.5s$] {\includegraphics[width=.22\columnwidth]{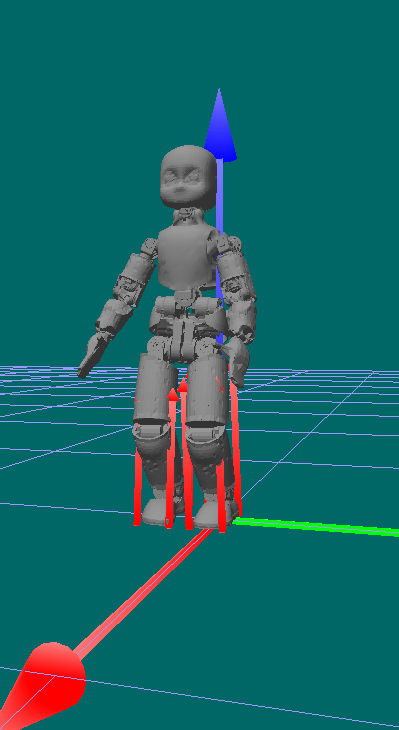}}
    \hspace{.001\columnwidth}
    \subfloat[$t=1.5s$] {\includegraphics[width=.22\columnwidth]{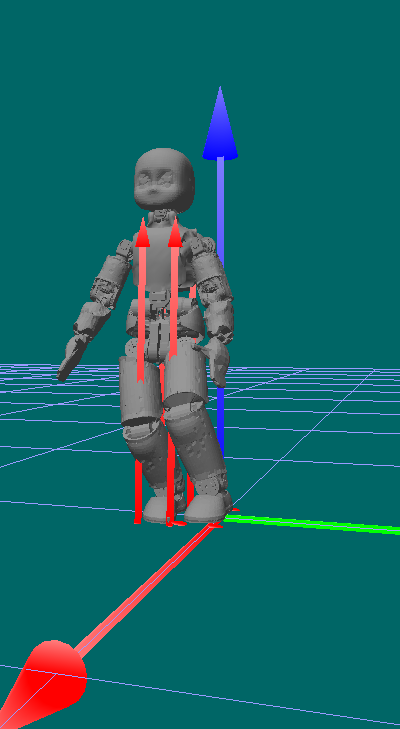}}
    \hspace{.001\columnwidth}
    \subfloat[$t=2.5s$] {\includegraphics[width=.22\columnwidth]{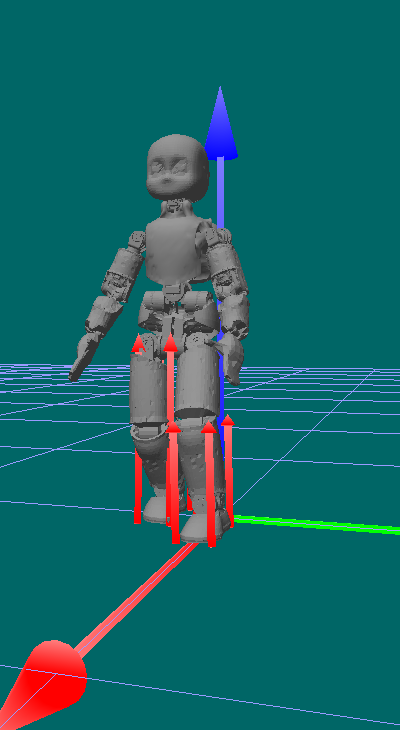}}
    \hspace{.001\columnwidth}
    \subfloat[$t=3.5s$] {\includegraphics[width=.22\columnwidth]{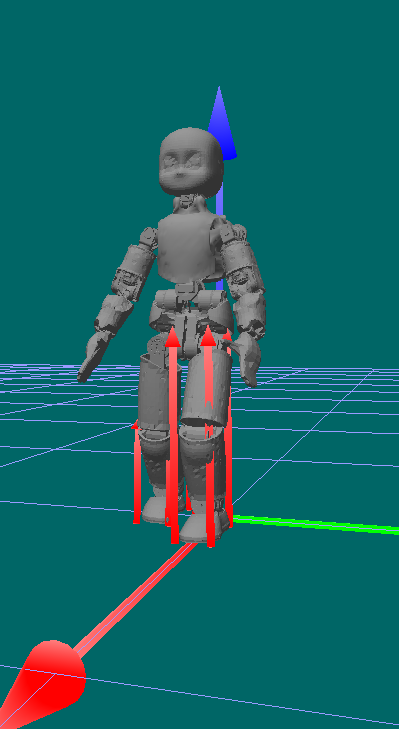}}
    \caption{Snapshots of the generated walking motion. The red arrows indicate the force required at each contact point scaled by a factor of 0.01. These images have been obtained using the complementarity constraints of Sec. \ref{sec:dynamical_complementarity}.}
    \label{fig:slow_straight}
\end{figure}
\begin{figure}[tpb]
	\centering
	\subfloat[\modtext{Relaxed complementarity}] {\includegraphics[width=.75\columnwidth]{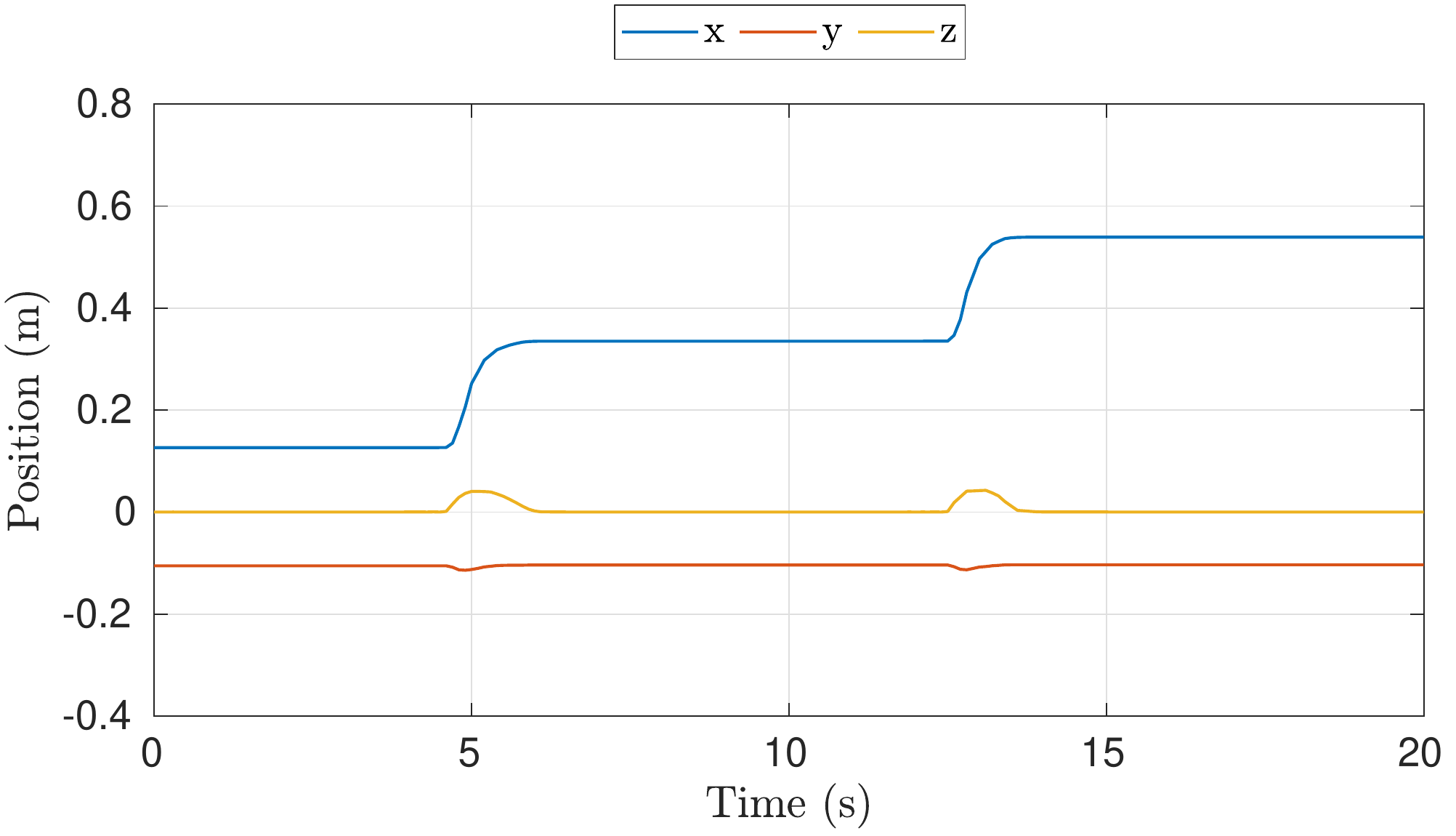}}

	\subfloat[\modtext{Dynamically enforced complementarity}] {\includegraphics[width=.75\columnwidth]{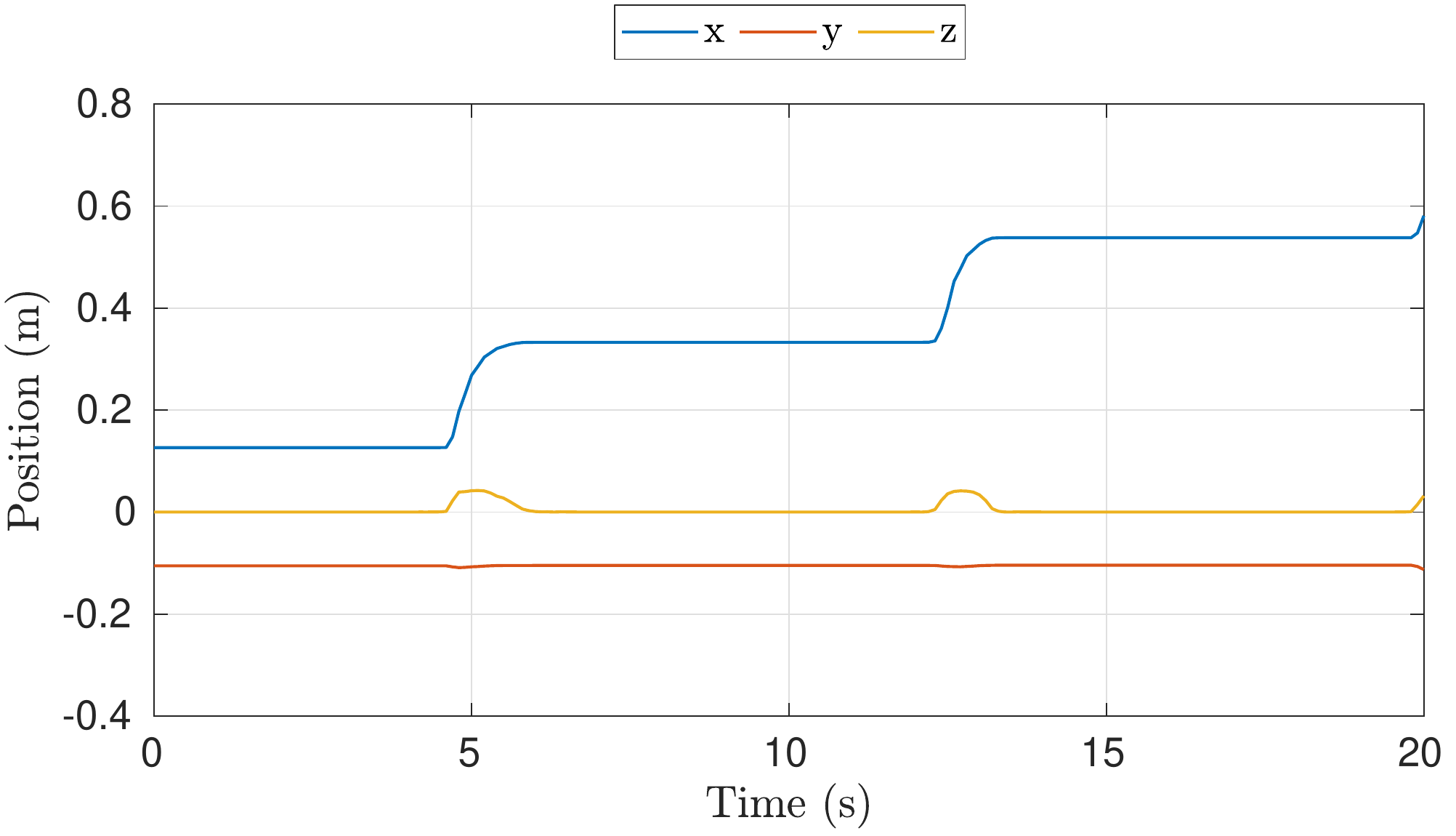}}
	
	\subfloat[\modtext{Hyperbolic secant in control bounds}] {\includegraphics[width=.75\columnwidth]{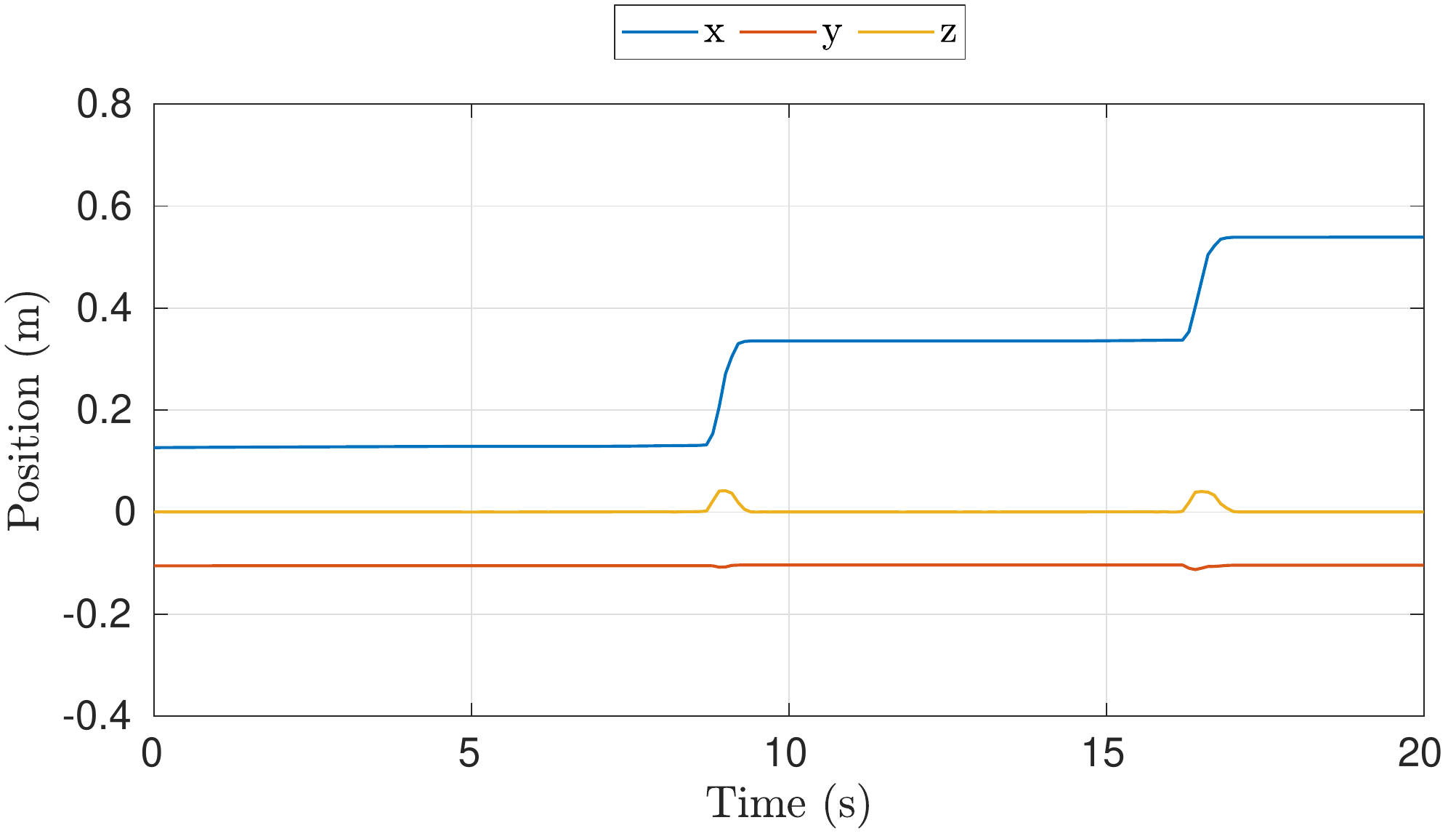}}
	\caption{Planned position of one of the right foot contact points using different complementarity constraints. The walking phases are recognizable, but they are not defined beforehand. The controller does not specify directly when a phase begins and ends.}
	\label{fig:point_position}
\end{figure}

\begin{figure}[tpb]
	\centering
	\subfloat[\modtext{Relaxed complementarity}] {\includegraphics[width=.75\columnwidth]{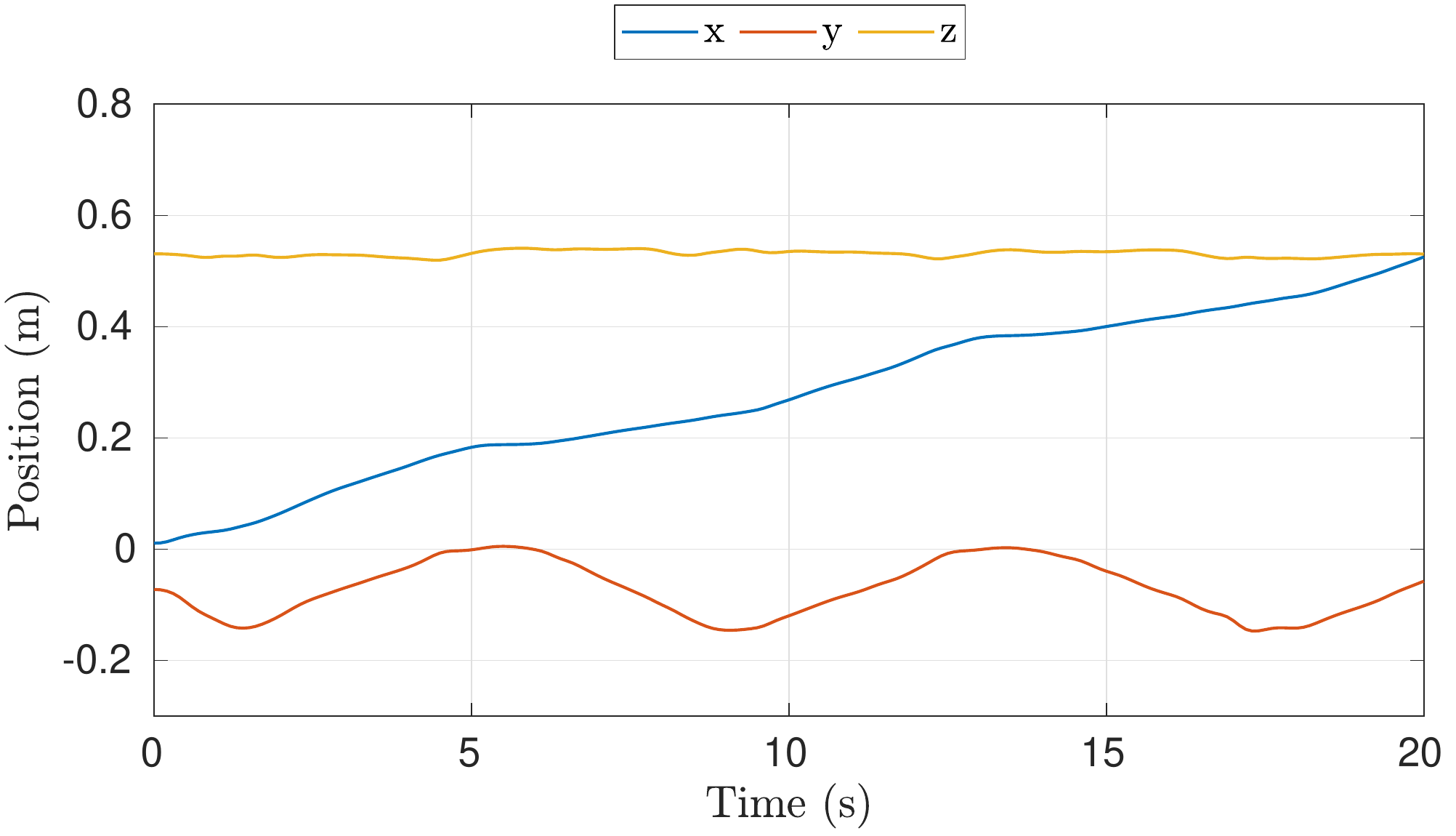}}

	\subfloat[\modtext{Dynamically enforced complementarity}] {\includegraphics[width=.75\columnwidth]{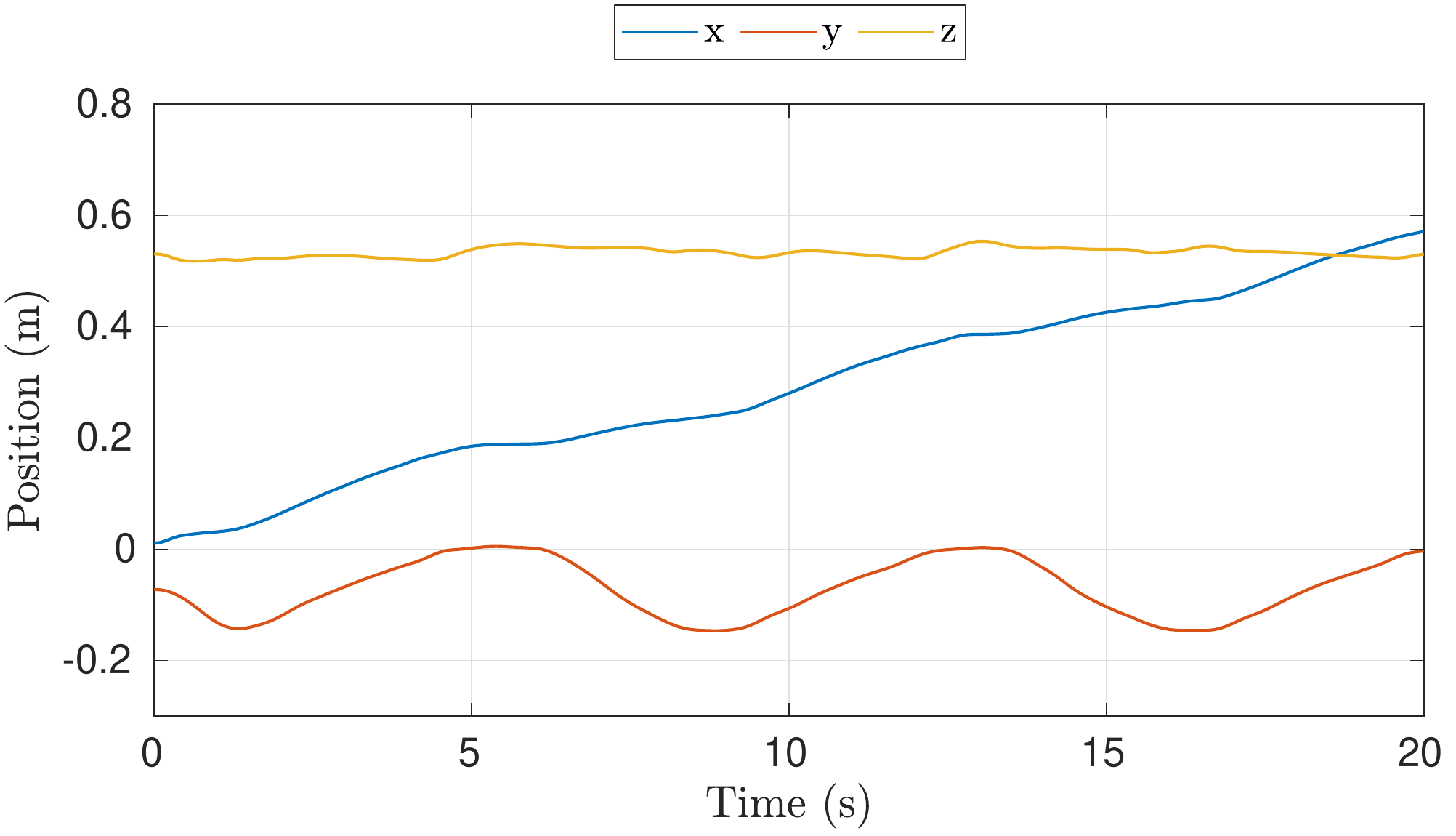}}

	\subfloat[\modtext{Hyperbolic secant in control bounds}] {\includegraphics[width=.75\columnwidth]{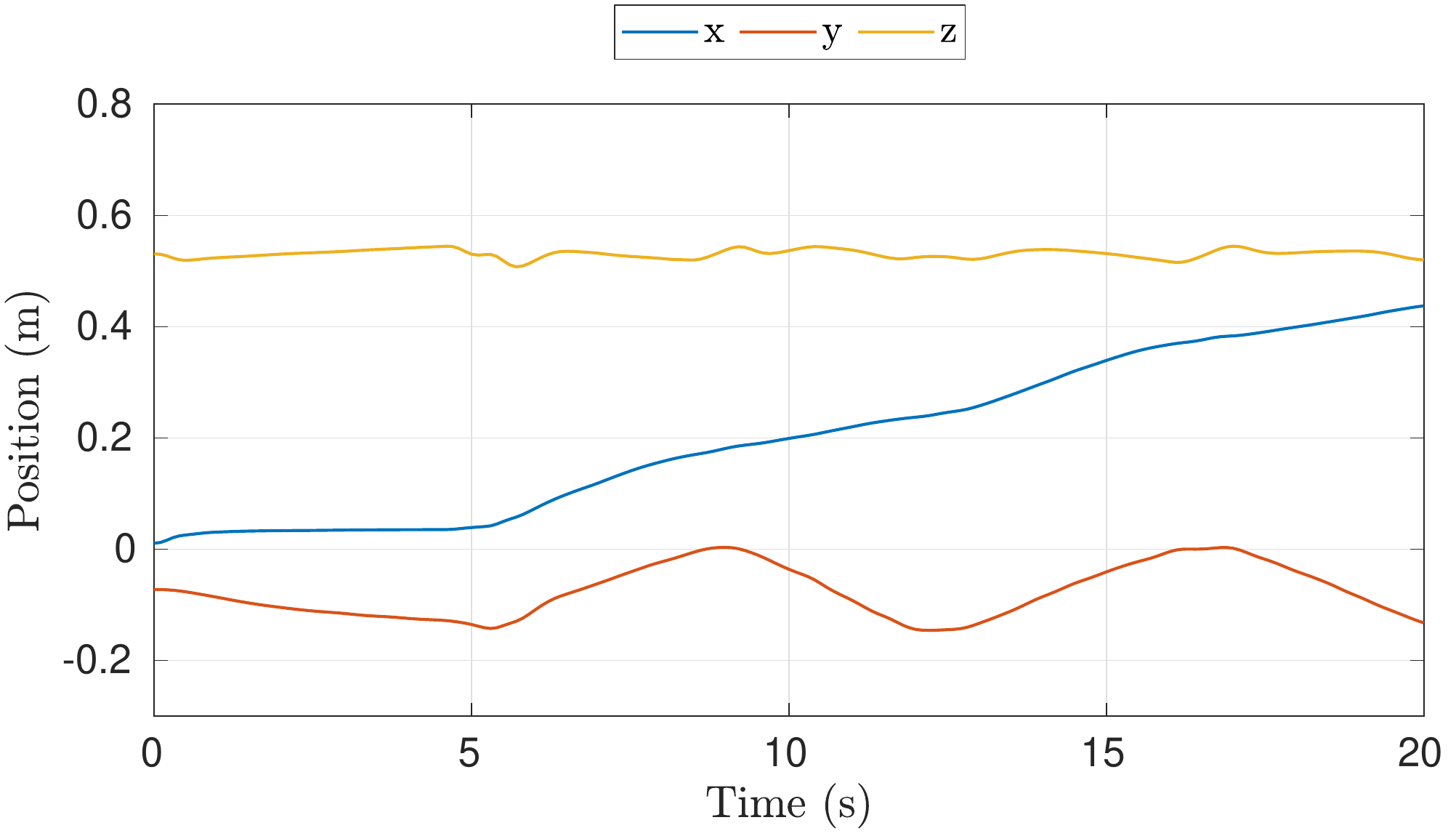}}
	\caption{Planned CoM position using different complementarity constraints. It is possible to notice a continuous velocity on the $x$ direction. The plots appear a little irregular. This may be a consequence of the chosen time step.}
	\label{fig:dp_com_position}
\end{figure}

\begin{figure}[tpb]
	\centering
	\subfloat[\modtext{Relaxed complementarity}] {\includegraphics[width=.75\columnwidth]{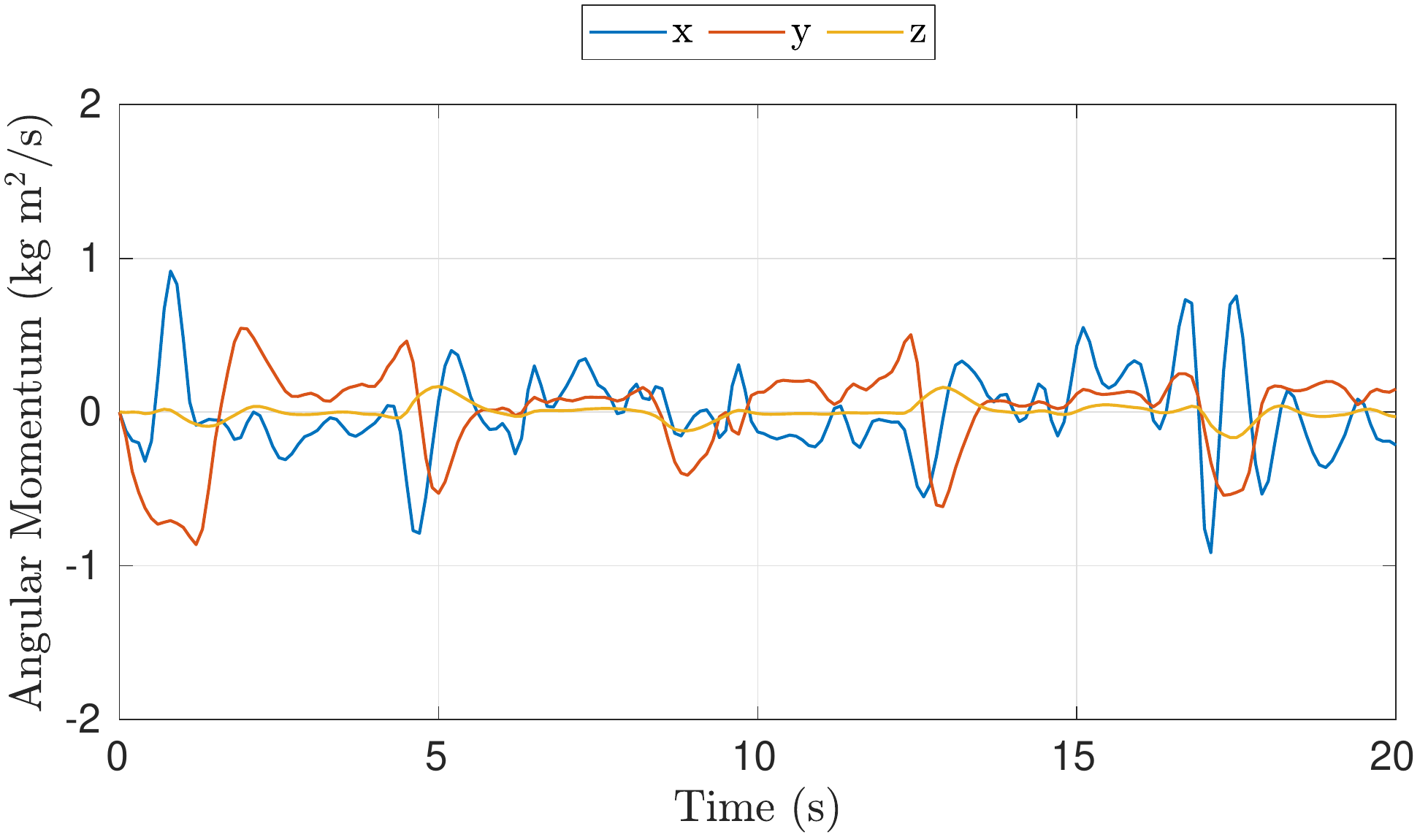}}

	\subfloat[\modtext{Dynamically enforced complementarity}] {\includegraphics[width=.75\columnwidth]{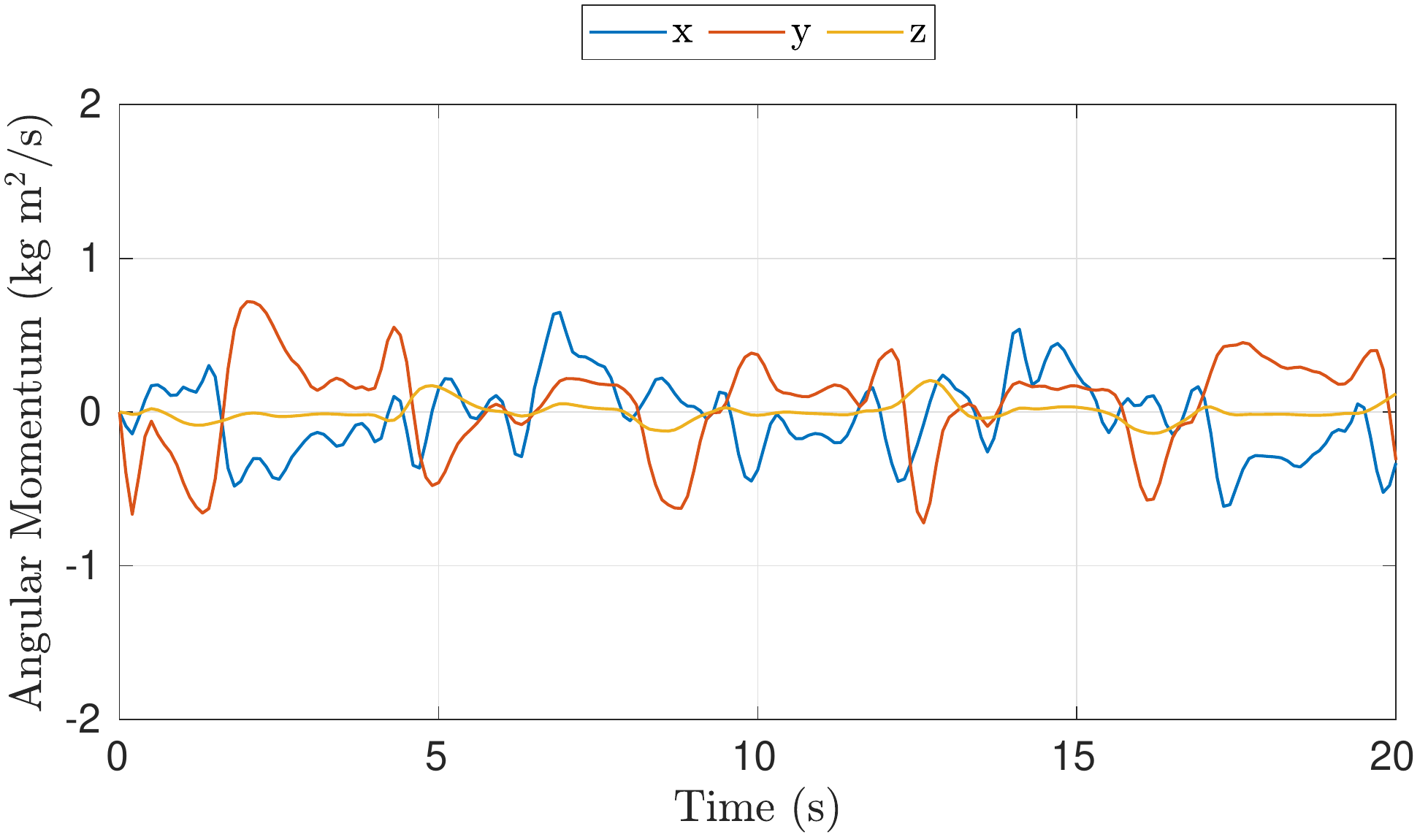}}
	
	\subfloat[\modtext{Hyperbolic secant in control bounds}] {\includegraphics[width=.75\columnwidth]{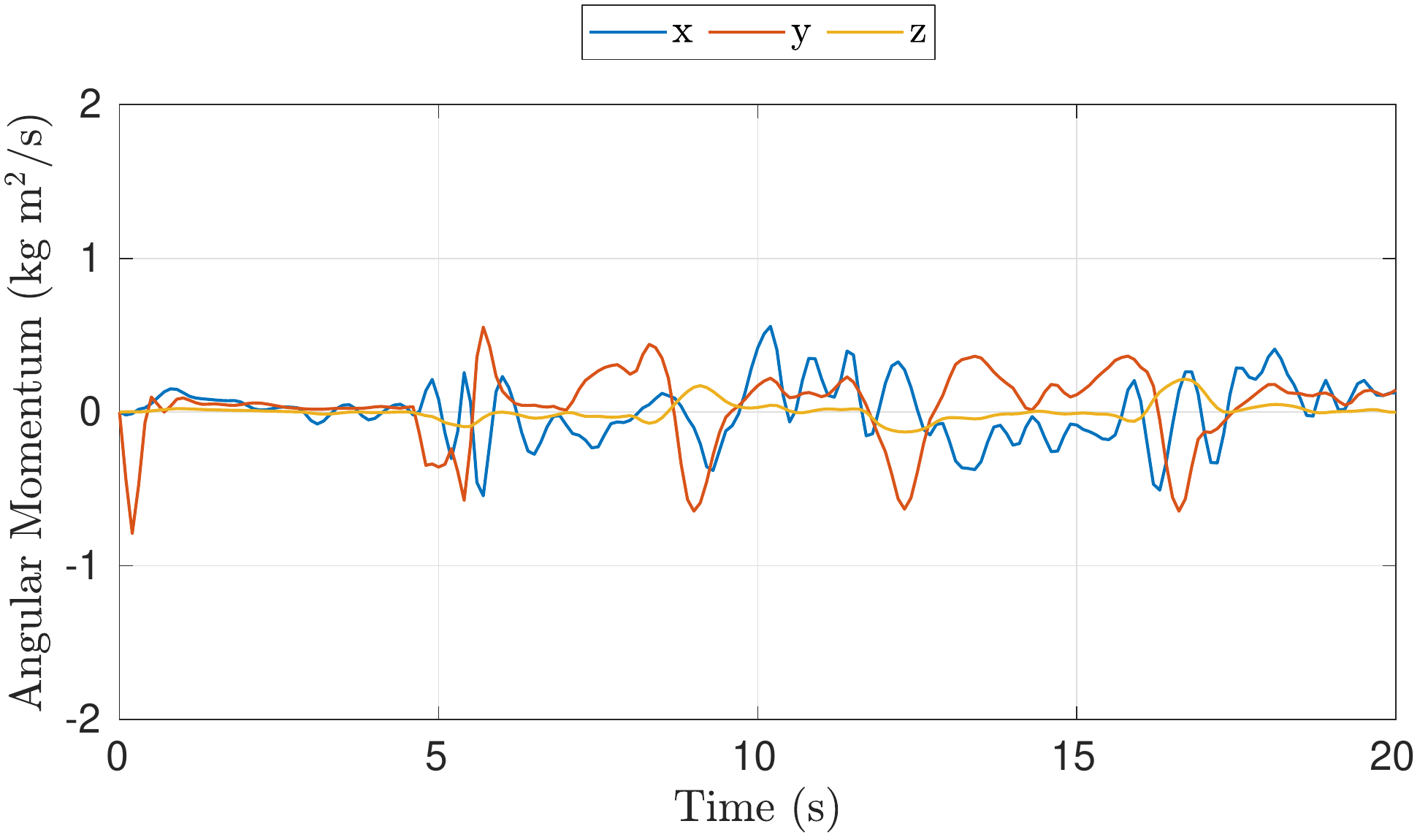}}
	\caption{Planned angular momentum obtained adopting different complementarity constraints. }
	\label{fig:dp_angular_momentum}
\end{figure}

\begin{table*}[tpb]
	\caption{Cost function weights used in the comparison of the complementarity methods. \modtext{The terms $\tilde{w}_{{}_\# p}(t)$ and $\tilde{w}_{\text{frame}-\text{torso}}(t)$ are defined in Eq. \eqref{eq:centroid_task_weight} and \eqref{eq:torso_task_weight}, respectively.}}
	\centering
	\begin{tabular}{|c c c c c c c c c c c|} 
		\hline
		$w_{{}_\# p}$ & $w_{{}_{\bar{G}} h^p}$ & $w_{\text{frame}-\text{torso}}$ & $w_{\text{frame}-\text{pelvis}}$ & \modtext{$w_{\text{reg} u_\rho}$} & $w_{\text{reg} f}$ & $w_{\text{reg} s}$ & $w_{\text{swing}}$ & $w_{\text{yaw}}$ & \modtext{$w_{\text{reg} u_p}$} &\modtext{$w_{\text{reg} u_f}$} \Tstrut\Bstrut\\[0.5ex]
		\hline\hline
		$100.0 \cdot \tilde{w}_{{}_\# p}(t)$ & $1.0$ & \modtext{$90.0\cdot \tilde{w}_{\text{frame}-\text{torso}}(t)$} & $50.0$ & $0.001$ & $0.1$ & $0.1$ & $1000$ & $1000$ & \modtext{$5.0$} & \modtext{$0.0001$}\Tstrut\Bstrut\\[0.5ex]
		\hline
	\end{tabular}
	\label{tab:complementarity_weights}
\end{table*}

We present the results obtained when solving the optimal control problem \modtextDani{presented} in Section \ref{sec:oc}. In particular, we test its capabilities to generate whole-body walking trajectories for a flat ground using the model of the iCub humanoid robot~\cite{Nataleeaaq1026}. All the tests presented in this section have been carried on a 7$^{th}$ generation Intel\textsuperscript{\textregistered} Core i7@2.8GHz laptop.

The optimal control problem described in Sec. \ref{sec:oc} is solved using a Direct Multiple Shooting method \cite{betts2010practical}. The system dynamics, defined in Eq. \eqref{constr:system_dynamics}, is discretized adopting an implicit trapezoidal method with a fixed integration step. The corresponding optimization problem is solved thanks to \texttt{Ipopt} \cite{IPOpt2006}, using the \texttt{MA57} linear solver \cite{hsl2007collection}. 
\modtextBis{The solver requires at least the first derivative of the cost and constraints with respect to the optimization variables. These are computed explicitly using the derivation presented in \cite{dafarra2020phd}}. 
The pipeline from the problem definition to its solution is implemented using the \texttt{iDynTree::optimalcontrol}\footnote{\url{https://github.com/robotology/idyntree/tree/devel/src/optimalcontrol}} library, allowing for easy testing of other integrators or solvers. The code is implemented entirely in C++ and open-source\footnote{\url{https://github.com/dic-iit/dynamical-planner}}.

\modtext{The walking trajectories are generated with a fixed prediction window of 2 seconds. The horizon is large enough to predict at least one full step.}



The planner is set up using an integration step of $100 \mathrm{ms}$ , while the time horizon is $2\mathrm{s}$. The choice of a large integration step serves for two reasons. First, it reduces the number of variables used by the optimization problem (fixing the time horizon). Secondly, it allows inserting another control loop at higher frequency. After each \modtextBis{run} of the planner, \modtext{the first and the second state are added to the final trajectory. The latter is also used as a feedback state for the new planner \modtextBis{run} in a \emph{receding horizon} fashion.}

When planning, we control 23 of the robot joints. We consider four contact points for each foot, located at the vertices of the rectangle enclosing the robot foot. Concerning the references, the desired position for the centroid of the contact points is moved $10 \mathrm{cm}$ along the walking direction every time the robot performs a step. A simple state machine where the reference is moved as soon as a step is completed, is enough to generate a continuous walking pattern. The speed is modulated by prescribing a desired CoM forward velocity equal to $5 \mathrm{cm/s}$. 
\modtext{The desired joint positions are fixed, and equal to the initial configuration of the robot. The terms $\bm{u}_\rho^*, \bm{u}_{{}_ip}^*$ and $\bm{u}_{{}_if}^*$ are set to zero.} The value of $k_t$ (see Sec. \ref{Planar-DCC-subsection}) is $10 \mathrm{m}^{-1}$. 

Fig. \ref{fig:slow_straight} shows some snapshots of the first generated step, while Fig. \ref{fig:point_position} shows the position of one of the right contact points. It is possible to recognize the different walking phases, though they are not planned a priori. Nevertheless, the controller does not specify explicitly when a phase begins and ends. \modtext{It is also possible to notice that the trajectories obtained using the \emph{Hyperbolic secant} method, Sec. \ref{sec:hyperbolic_secant}, have been affected by some initial ``procrastination'', with a longer initial double support phase. Those produced with the other two methods are instead very similar.}

Figure \ref{fig:dp_com_position} presents the planned CoM position. Here, it is possible to notice that the position along the $x$ direction grows at a constant rate. This is a direct consequence of the task on the CoM velocity presented in Sec. \ref{sec:com_velocity_cost}. 
Notice that the bound on the CoM height, $x_{z \text{ min}}$, is set to half of the initial robot height, but such constraint is never activated.
These are the results of consecutive runs of the optimal control problem described in Sec. \ref{sec:oc}. From one \modtextBis{run} to another, the solver may find slightly different solutions because of the shift in the prediction horizon, causing the irregularities shown in the figures. In addition, there is no regularization on the CoM position, whose trajectory is fully determined by the solver.

Figure \ref{fig:dp_angular_momentum} shows the planned angular momentum, that is not fixed to zero, but limited to $10~\mathrm{kg}~{\mathrm{m}^2}/{\mathrm{s}}$. 
\modtext{This limit is never reached. Thanks to the regularization terms on the base and joint velocities, the angular momentum is kept close to zero.}

It is worth stressing that none of the tasks described above define how and when to raise the foot. By prescribing a reference for the centroid of the contact points and by preventing the motion on the contact surface, swing motions are planned automatically. Nevertheless, this advantage comes with a cost. It is difficult to define a desired swing time and, more importantly, the relative importance of each task, i.e. the values of $\mathbf{w}$, must be chosen carefully. As anticipated above, in order to avoid ``procrastination'' phenomena, we adopted a time-varying weight for the centroid position of the contact points, described in Sec. \ref{sec:centroid_task}. In particular, we define
\begin{equation}\label{eq:centroid_task_weight}
	\tilde{w}_{{}_\# p}(t) = \alpha_{ts}(t) w_{ts} + 1,
\end{equation}
where ${w}_{ts} = 30$, while $\alpha_{ts}$ determines the current percentage of the step. It is computed as follows:
\begin{equation}
	\alpha_{ts}(t) = \frac{t - t_{s, \text{init}} }{t_s^*},
\end{equation}
where $t_s^*$ is the desired step duration, while $t_{s, \text{init}}$ is the first moment in which a specific position reference is active.

\modtext{At the same time, since only the initial states of a given solution are considered in the final trajectory, it is necessary to increase the importance of some tasks at the beginning of the horizon. \modtextDani{One specific example} is the orientation of the torso. In order to avoid undesired motions for \modtextDani{the robot upper-body}, we set a weight for \modtextDani{the torso orientation} task that decreases exponentially with time. In particular, we have
\begin{equation}\label{eq:torso_task_weight}
	\tilde{w}_{\text{frame}-\text{torso}}(t) = 100 e^{-10t} + 1,
\end{equation}
where we assume, without loss of generality, that $t \in [0, 2]$.
}

The weights we adopted in the cost function are listed in Table \ref{tab:complementarity_weights}. During experiments, we adopted an incremental approach to determine them. We added the tasks one by one, starting from $\Gamma_{{}_\# p}$ and then we gradually refine the walking motion by tuning one cost at a time. 

\modtext{
\subsection{Real Robot Validation} \label{sec:dp_robot_experiments}
}
\begin{figure*}[tpb]
	\centering
	\subfloat[$t=t_0$] {\includegraphics[width=.2\textwidth]{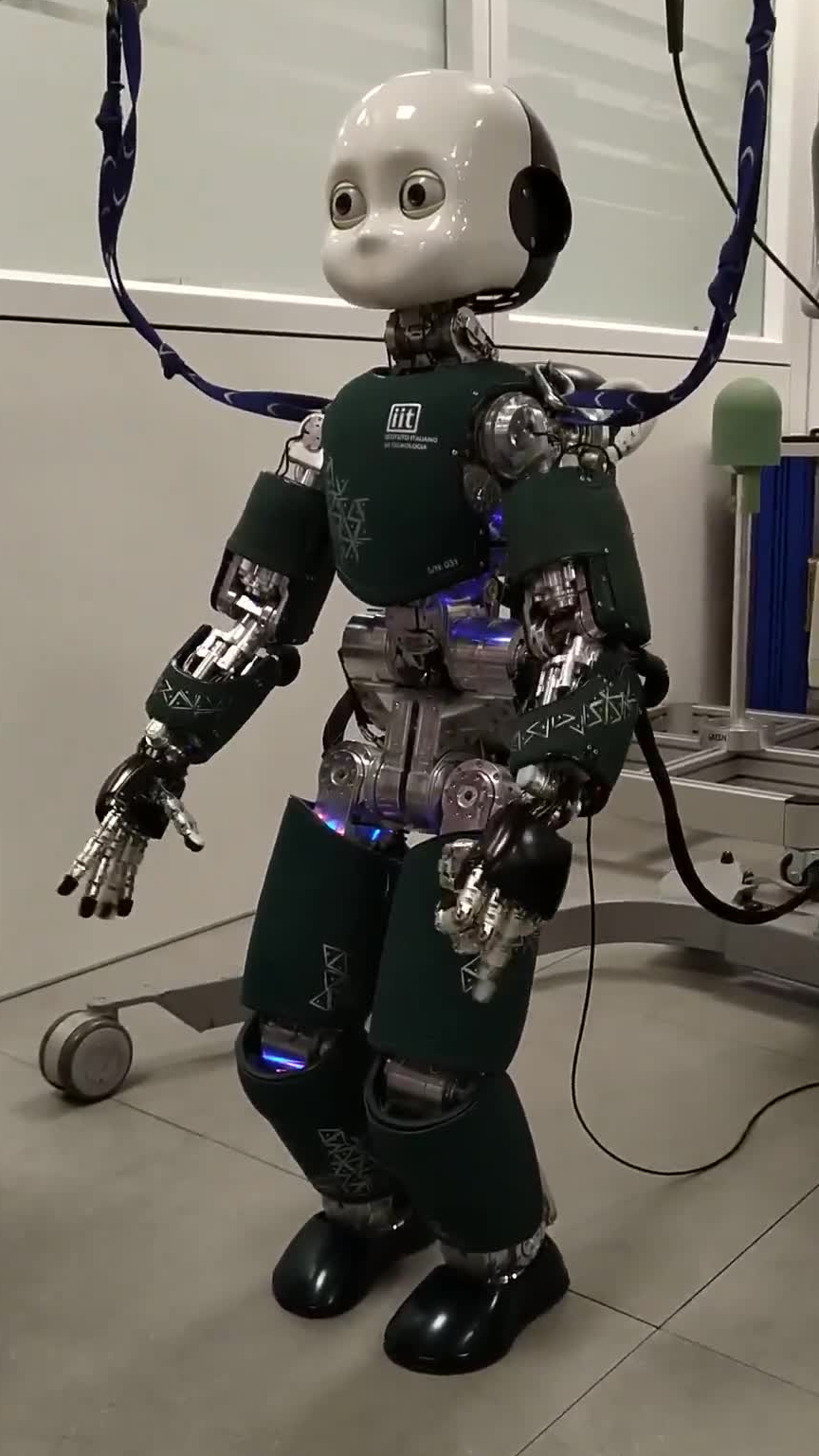}}
	\hspace{.01\textwidth}
	\subfloat[$t=t_0 + 1s$] {\includegraphics[width=.2\textwidth]{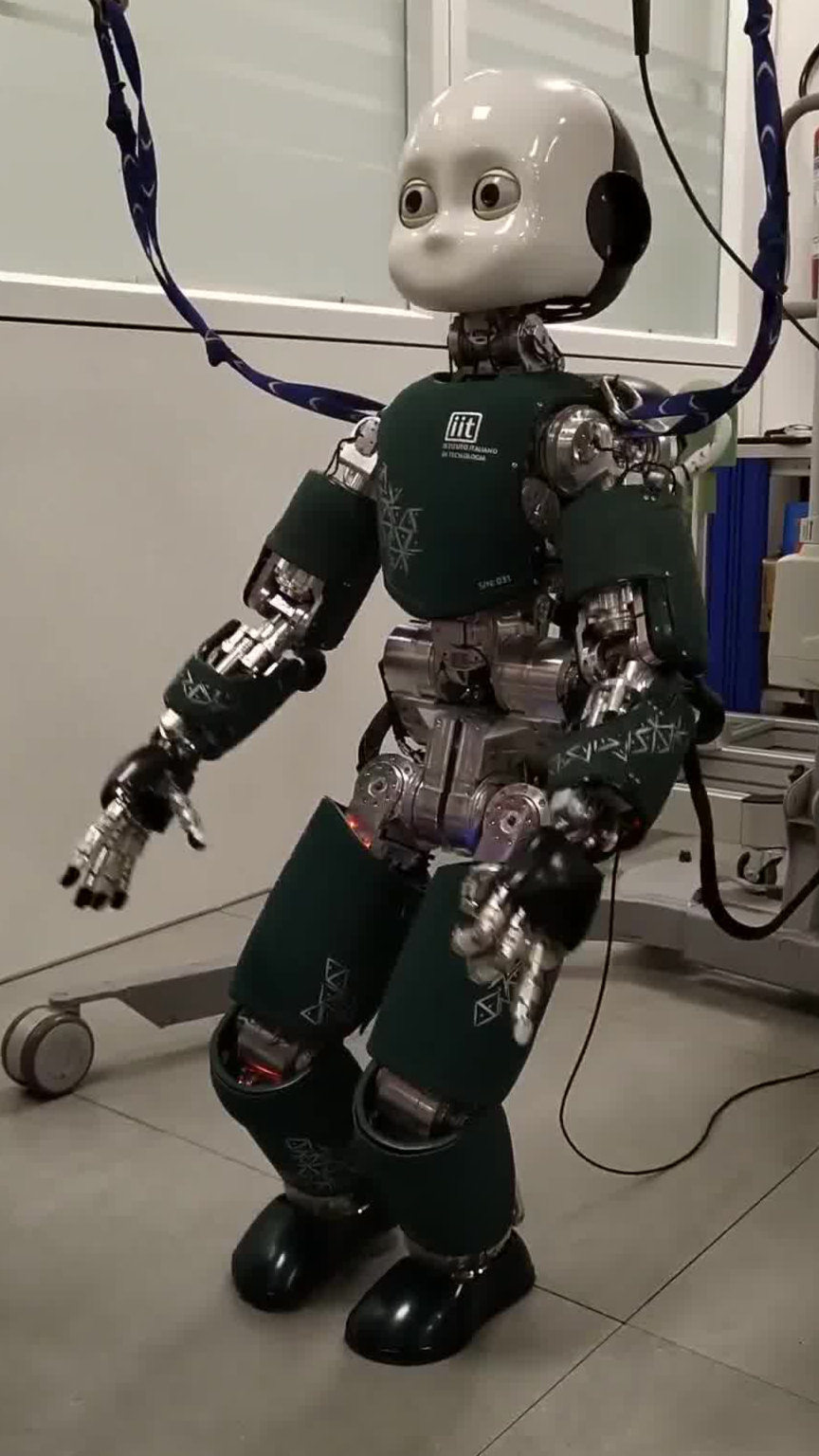}}
	\hspace{.01\textwidth}
	\subfloat[$t=t_0 + 2s$] {\includegraphics[width=.2\textwidth]{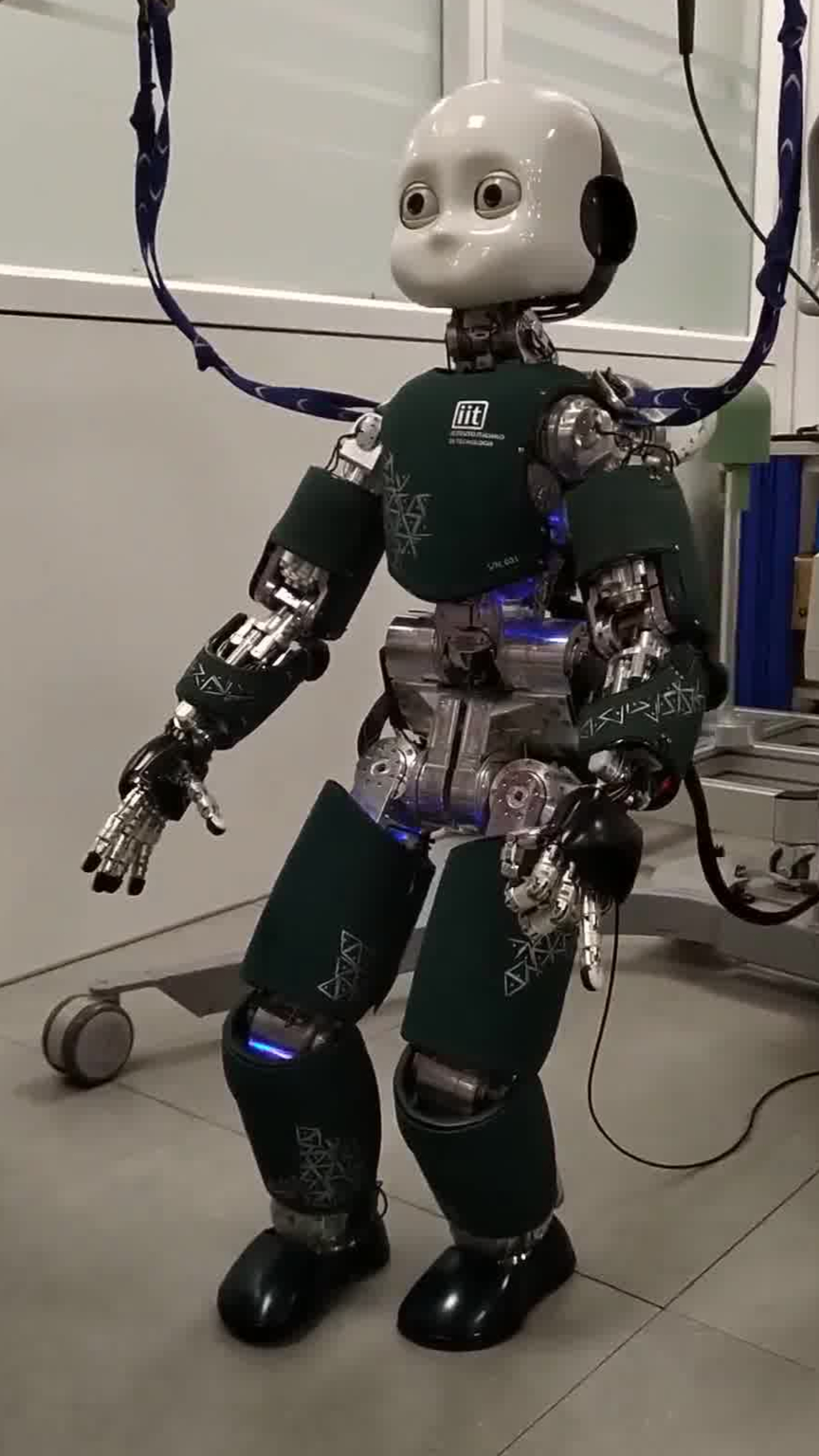}}
	\hspace{.01\textwidth}
	\subfloat[$t=t_0 + 3s$] {\includegraphics[width=.2\textwidth]{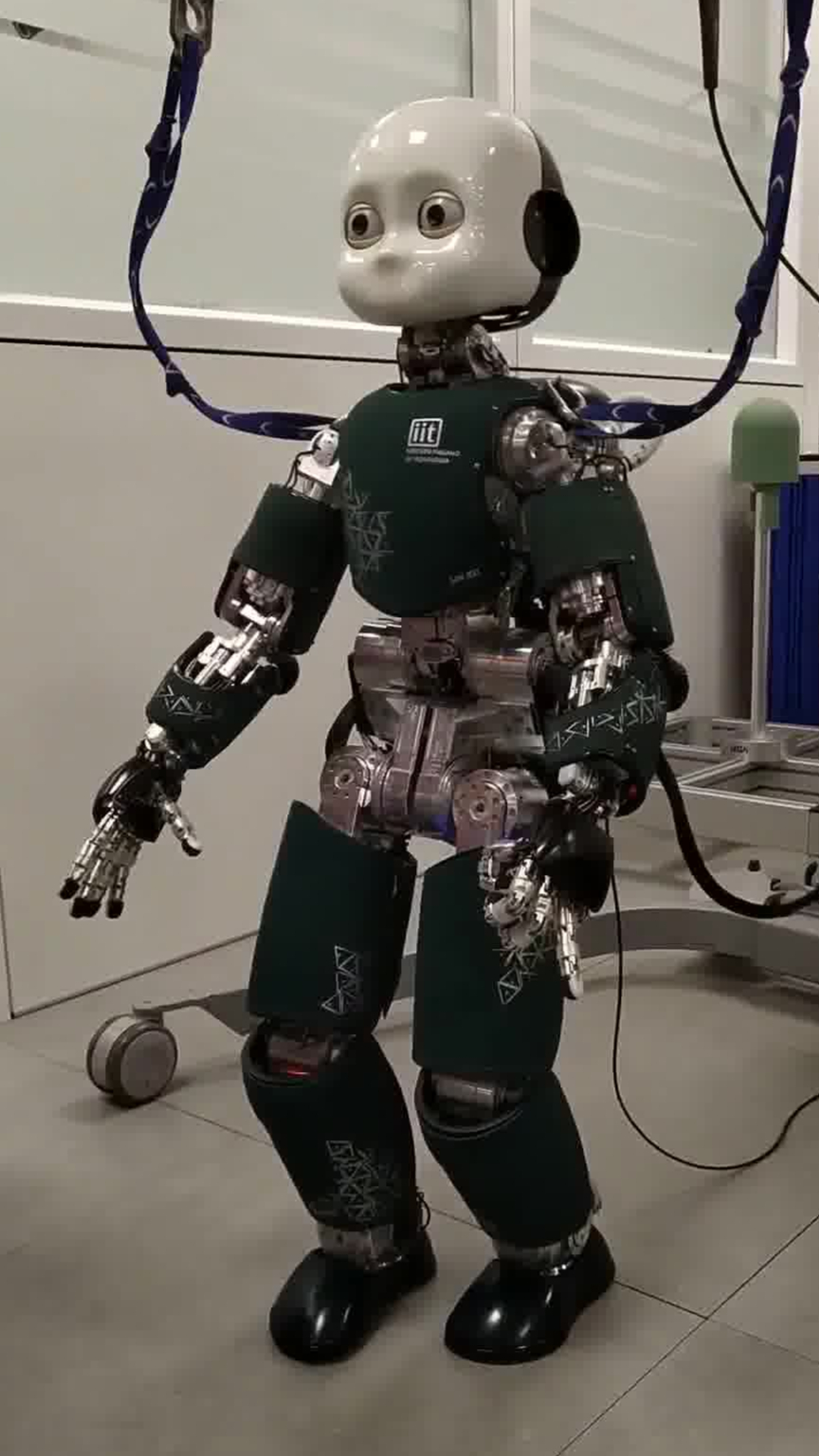}}
	
	\subfloat[$t=t_0 + 4s$] {\includegraphics[width=.2\textwidth]{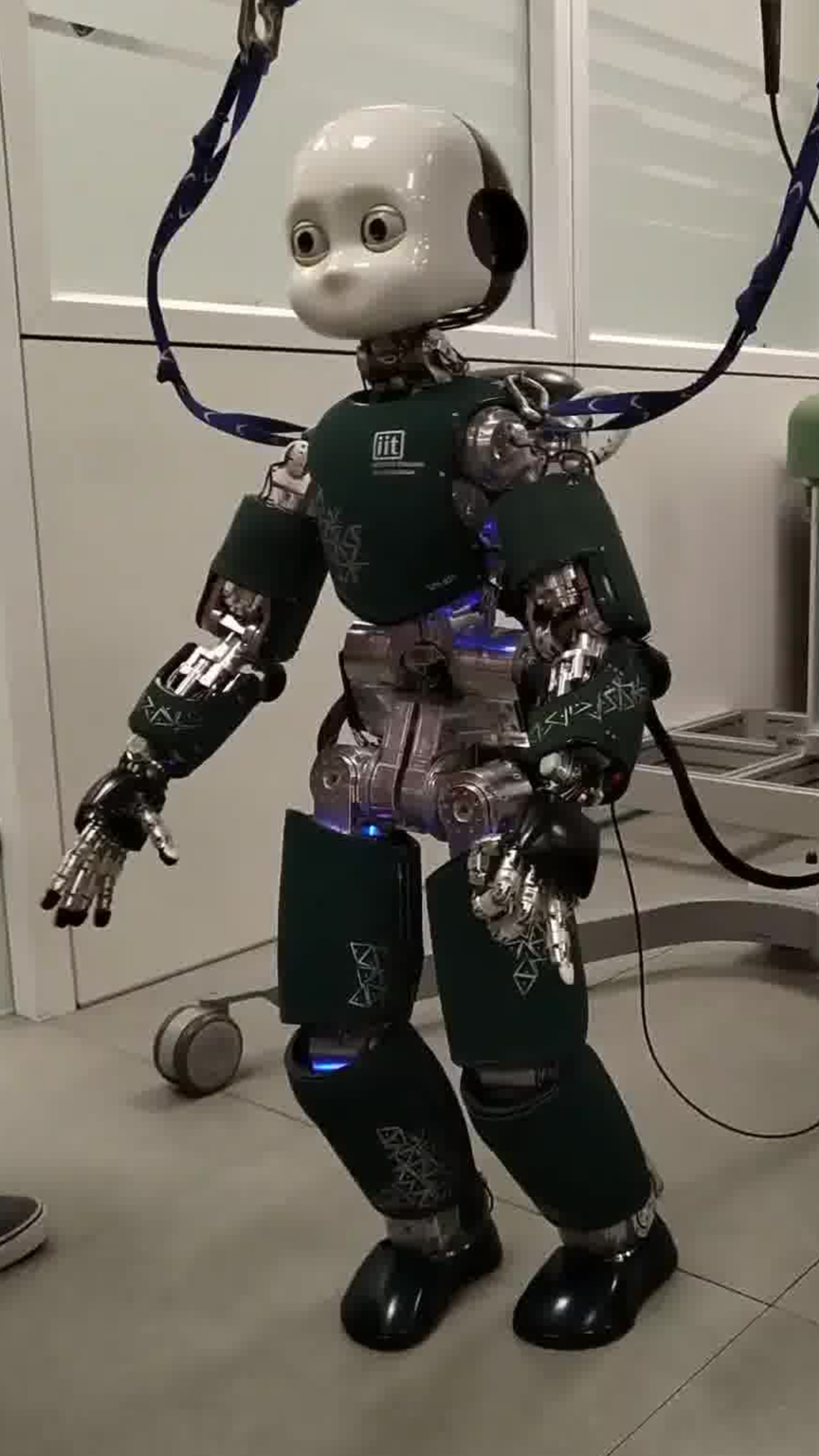}}
	\hspace{.01\textwidth}
	\subfloat[$t=t_0 + 5s$] {\includegraphics[width=.2\textwidth]{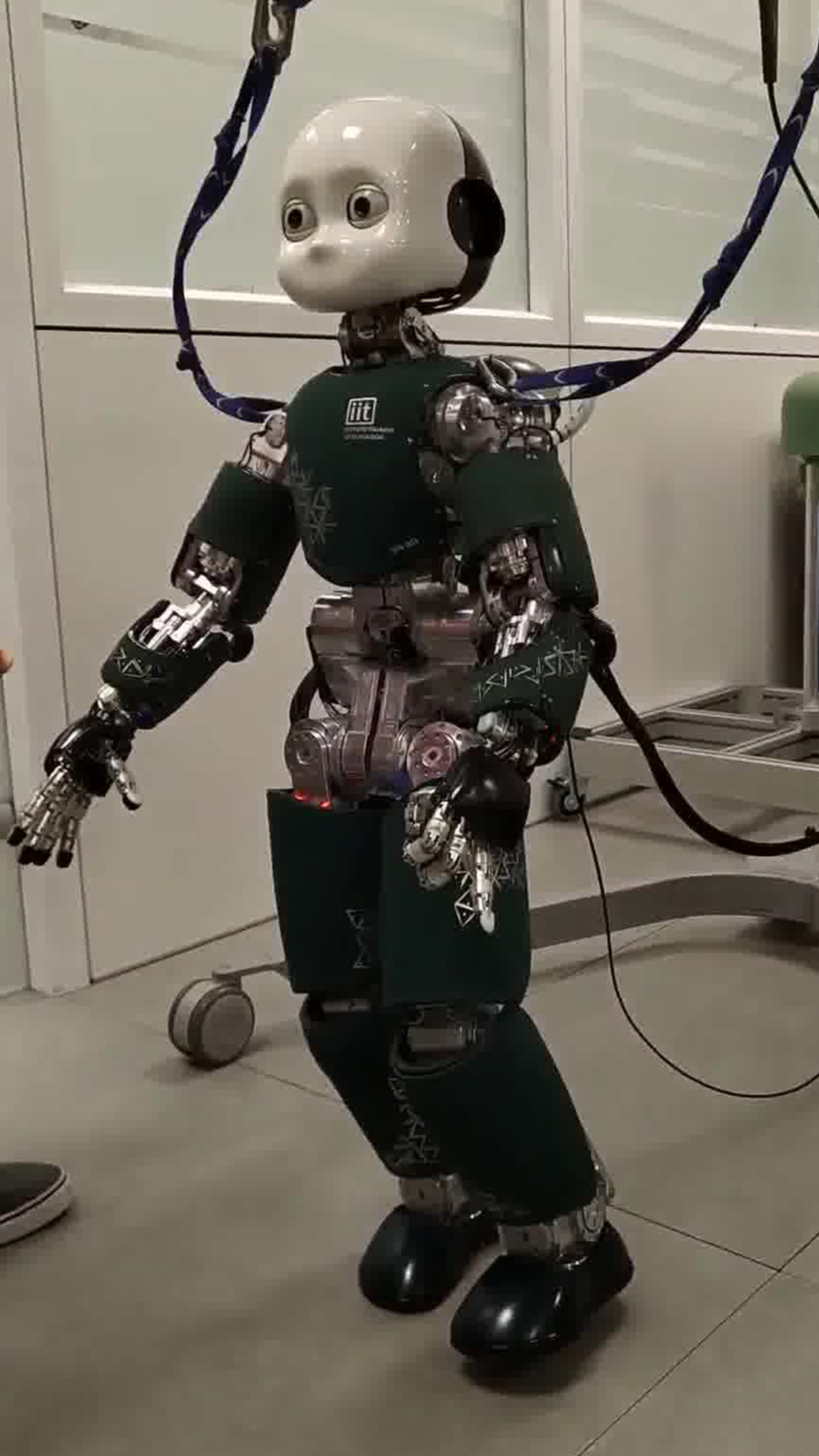}}
	\hspace{.01\textwidth}
	\subfloat[$t=t_0 + 6s$] {\includegraphics[width=.2\textwidth]{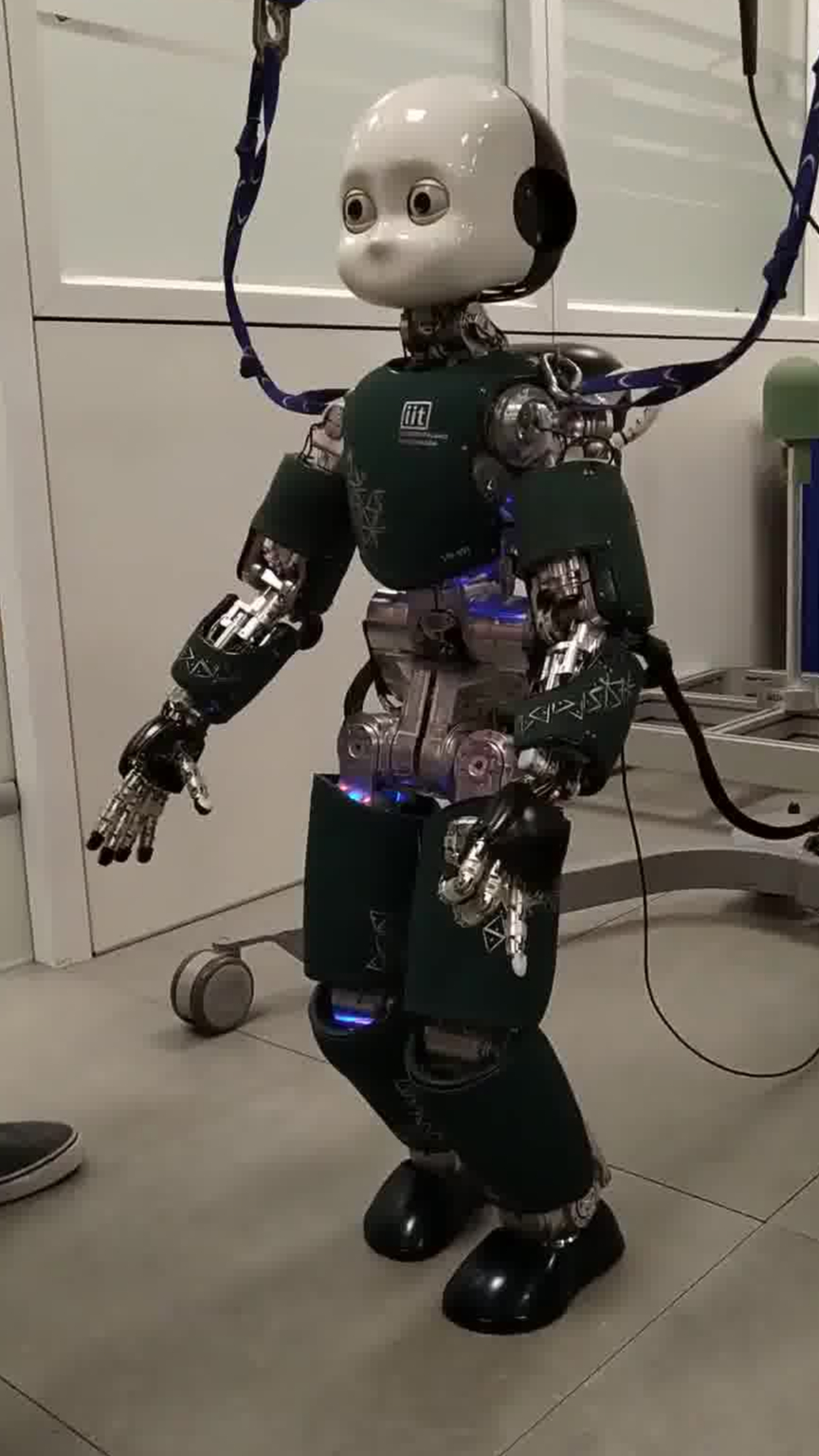}}
	\hspace{.01\textwidth}
	\subfloat[$t=t_0 + 7s$] {\includegraphics[width=.2\textwidth]{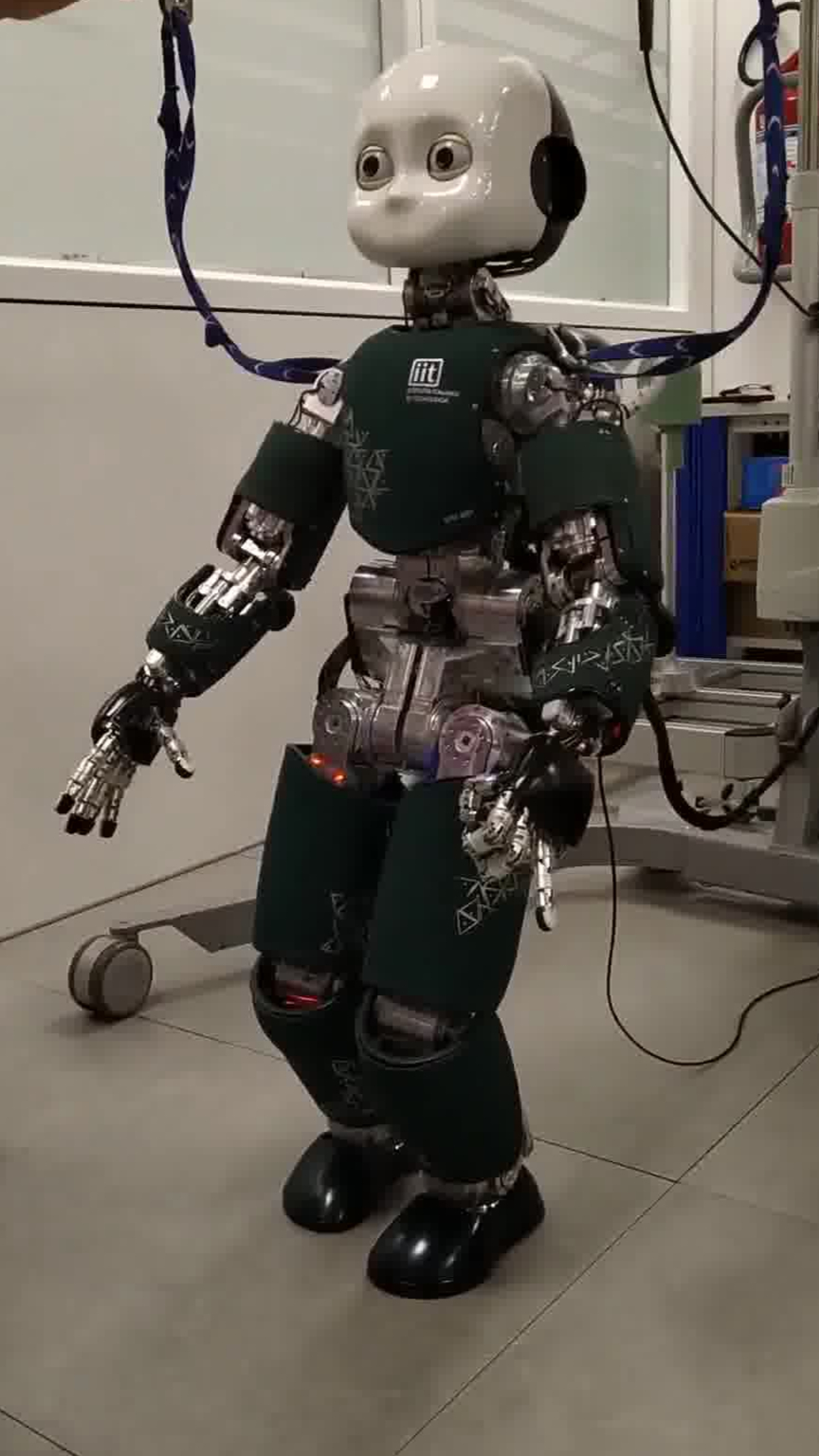}}
	\caption{Snapshots of the robot walking using the planned trajectories. The planner generates joint references which are interpolated and stabilized through a joint position controller.}
	\label{fig_dp:snapshots}
\end{figure*}

\begin{table*}[tpb]
	\caption{Cost function weights used in the real robot validation. The term $\tilde{w}_{{}_\# p}(t)$ is defined in Eq. \ref{eq:centroid_task_weight}.}
	\centering
	\begin{tabular}{|c c c c c c c c c c c|} 
		\hline
		$w_{{}_\# p}$ & $w_{{}_{\bar{G}} h^p}$ & $w_{\text{frame}-\text{torso}}$ & $w_{\text{frame}-\text{pelvis}}$ & \modtext{$w_{\text{reg} u_\rho}$} & $w_{\text{reg} f}$ & $w_{\text{reg} s}$ & $w_{\text{swing}}$ & $w_{\text{yaw}}$ & \modtext{$w_{\text{reg} u_p}$} &\modtext{$w_{\text{reg} u_f}$} \Tstrut\Bstrut\\[0.5ex]
		\hline\hline
		$100.0 \cdot \tilde{w}_{{}_\# p}(t)$ & $1.0$ & $90.0$ & $20.0$ & $\modtext{0.0}$ & $0.1$ & $0.1$ & $1000$ & $1000$ & \modtext{$5.0$} & \modtext{$0.0$}\Tstrut\Bstrut\\[0.5ex]
		\hline
	\end{tabular}
	\label{tab:robot_weights}
\end{table*}
	
To further validate the output of the planner presented in this part, we tested the generated trajectories on the iCub humanoid robot. In particular, they are used as a reference for a joint position controller. Since their frequency is at 10$\mathrm{Hz}$, we interpolate them to have a reference point every 10$\mathrm{ms}$. Hence, the trajectories are replayed on the robot closing the loop only at joint level. The robot performed several steps, Fig.~\ref{fig_dp:snapshots}, demonstrating the feasibility of the generated trajectories.

\modtext{Compared} to the results shown in Sec. \ref{sec:validation}, we reduce the forward speed to $3\mathrm{cm}/\mathrm{s}$, advancing the mean point reference of $6\mathrm{cm}$ at every step. In this case, it has been also useful to move the desired CoP position, as anticipated in Sec. \ref{sec:forceRegularization}. In particular, by moving it toward the inner foot edge, the robot walks more robustly. The cost function weights are listed in Table \ref{tab:robot_weights}, while $w_{ts} = 60$.

The trajectories are generated off-line for a time span of twenty seconds. They are tested in open-loop, thus limiting the maximum velocity achievable by the robot. \modtext{This specific test is aimed at validating the formulation of the optimal control problem described in Sec. \ref{sec:oc}, independently from the choice of the complementarity method (in this case we selected the dynamically-enforced complementarity method). In the next section, we analyze the difference between each method.}

\modtextBis{\section{DCCs comparisons}\label{sec:complementarity_comparison}}

In this section, we analyze the differences amongst the complementarity methods described in Sec. \ref{sec:complementarity_list}. \modtextBis{In order to asses the relative performances of the different methods, we adopt the DCCs in two contexts with a crescendo of complexity. First, in Sec. \ref{sec:toy_problem_comparison}, we apply the DCCs in a toy problem involving a single contact. Then, in Sec. \ref{sec:comparison_nlp}, we exploit the output of the non-linear trajectory planning framework presented in Sec. \ref{sec:dcc_nlmpc}.}

As a measure of performance, we adopt the product between the normal force and the height of the contact point from the ground, i.e. ${}_ip_z\cdot{}_if_z$. In other words, we test the accuracy with which Eq. \eqref{eq:complementarity} is satisfied, simplifying the formulation thanks to the planar ground assumption. 
\modtext{As another performance measure, we use the average computational time needed to \modtextBis{solve the corresponding optimization problem.}}

\modtextBis{In order to have the results of Sec. \ref{sec:toy_problem_comparison} and Sec. \ref{sec:comparison_nlp} comparable, we use the same integration method, time horizon, and integration step described in Sec. \ref{sec:validation} in both cases. Similarly, the \texttt{Ipopt} configuration is identical.}

Moreover, for the sake of reproducibility, the comparisons involving parameter variations run also on \texttt{Github Actions}\footnote{See \url{https://github.com/ami-iit/paper_dafarra_2022_tro_dcc-planner}.}. Thus,  we provide a comparison using a standard machine. In this latter case, \texttt{Ipopt} uses the openly available \texttt{MUMPS} \cite{amestoy2000mumps} linear solver.

\modtext{The various methods have been tuned to obtain the best complementarity accuracy, while remaining able to generate a walking pattern}. Indeed, parameters that are too ``restrictive'' (e.g. an $\epsilon$ too small) may prevent the solver from finding a walking strategy because the points are not able to move. On the other hand, low accuracy may mean the solver requires a force of considerable magnitude when the point is still raised from the ground. \modtextBis{The same parameters have been adopted also for the toy problem.}
The parameters are chosen as follows:
\begin{itemize}
	\item Relaxed complementarity: $\epsilon = \modtext{0.004} [\mathrm{N\cdot m}]$.
	\item Dynamically enforced complementarity: $K_\text{bs} = 20\left[\frac{1}{\mathrm{s}}\right]$, $\varepsilon = \modtext{0.05}\left[\frac{\mathrm{N\cdot m}}{\mathrm{s}}\right]$.
	\item Hyperbolic secant in control bounds: $K_{f,z} = 250\left[\frac{1}{\mathrm{s}}\right]$, $k_h = \modtext{500}\left[\frac{1}{\mathrm{m}}\right]$.
\end{itemize}

$\bm{M}_f$ is set to 100$\mathrm{N}/\mathrm{s}$ for all the components and it is common for all the methods. 

\modtextBis{\subsection{Comparison using a toy problem} \label{sec:toy_problem_comparison}}
\modtextBis{
We consider a point mass falling vertically from a given height and approaching ground. We assume the ground to be infinitely rigid, and the ground level to be at zero. These settings render the analysed toy problem interesting also from a robotics perspective since it exemplifies the case of a humanoid robot foot approaching ground. 
Then, the mass position $x_m \in \mathbb{R}$ has to satisfy the following constraint:
\begin{equation}
    x_m \geq 0.
\end{equation}
When the mass hits the ground, a force is applied to it. This is modelled through the \emph{Dynamic Complementarity Conditions} presented in Sec. \ref{sec:dcc}. We assume the impact to be completely inelastic. On the other hand, the dynamics of the mass would not be continuous since its velocity should suddenly go to zero at the moment of the impact. In our toy problem, we circumvent this issue by assuming to be able to control the mass acceleration, as if a propeller is powering the mass.
}

\modtextBis{
The resulting system has the following dynamics:
\begin{IEEEeqnarray}{RCL}
	\phantomsection \IEEEyesnumber \label{eq:simpleMassDyn}
	\dot{x}_m &=& v_m, \IEEEyessubnumber\\
	\dot{v}_m &=& g + \frac{1}{m}\left(f_m + p_m\right) , \IEEEyessubnumber\\
	\dot{f}_m &=& u_f, \IEEEyessubnumber
\end{IEEEeqnarray}
where $v_m, g, f_m, p_m, u_f \in \mathbb{R}$ are respectively the mass velocity, the gravity acceleration, the contact force, the propeller thrust force, and the contact force derivative. The control inputs are given by $p_m$ and $u_f$, and the magnitude of $u_f$ is limited to a maximum value $M_{u_f}$. The objective is to minimise the use of $p_m$ as if the propeller had fuel constraints.
}

\modtextBis{
We insert the dynamical system of Eq. \eqref{eq:simpleMassDyn} and the DCCs in the following optimal control problem:
\begin{IEEEeqnarray}{CRCL}
	\IEEEyesnumber \phantomsection \label{eq:simpleMassOc}
	\minimize & \IEEEeqnarraymulticol{3}{C}{p_m^2} \IEEEyessubnumber\\
	\text{subject to:}&  \nonumber\\
	\IEEEeqnarraymulticol{2}{R}{\dot{x}_m} &=& v_m, \IEEEyessubnumber\\
	\IEEEeqnarraymulticol{2}{R}{\dot{v}_m} &=& g + \frac{1}{m}\left(f_m + p_m\right) , \IEEEyessubnumber\\
	\IEEEeqnarraymulticol{2}{R}{\dot{f}_m} &=& u_f, \IEEEyessubnumber \\
	\IEEEeqnarraymulticol{2}{R}{x_m} &\geq& 0, \IEEEyessubnumber \\
	\IEEEeqnarraymulticol{2}{R}{f_m} &\geq& 0, \IEEEyessubnumber \\
	\IEEEeqnarraymulticol{2}{R}{-M_{u_f}} &\leq& u_f \leq M_{u_f}, \IEEEyessubnumber \\
	\IEEEeqnarraymulticol{2}{R}{\text{C}}&\IEEEeqnarraymulticol{2}{L}{\text{omplementarity, see Sec. \ref{sec:complementarity_list}}}.\IEEEyessubnumber
\end{IEEEeqnarray}
}
\modtextBis{
\subsubsection{Toy problem solution}
}
\begin{figure}[tpb]
	\centering
	\subfloat[\modtextBis{Relaxed complementarity}] {\includegraphics[width=.75\columnwidth]{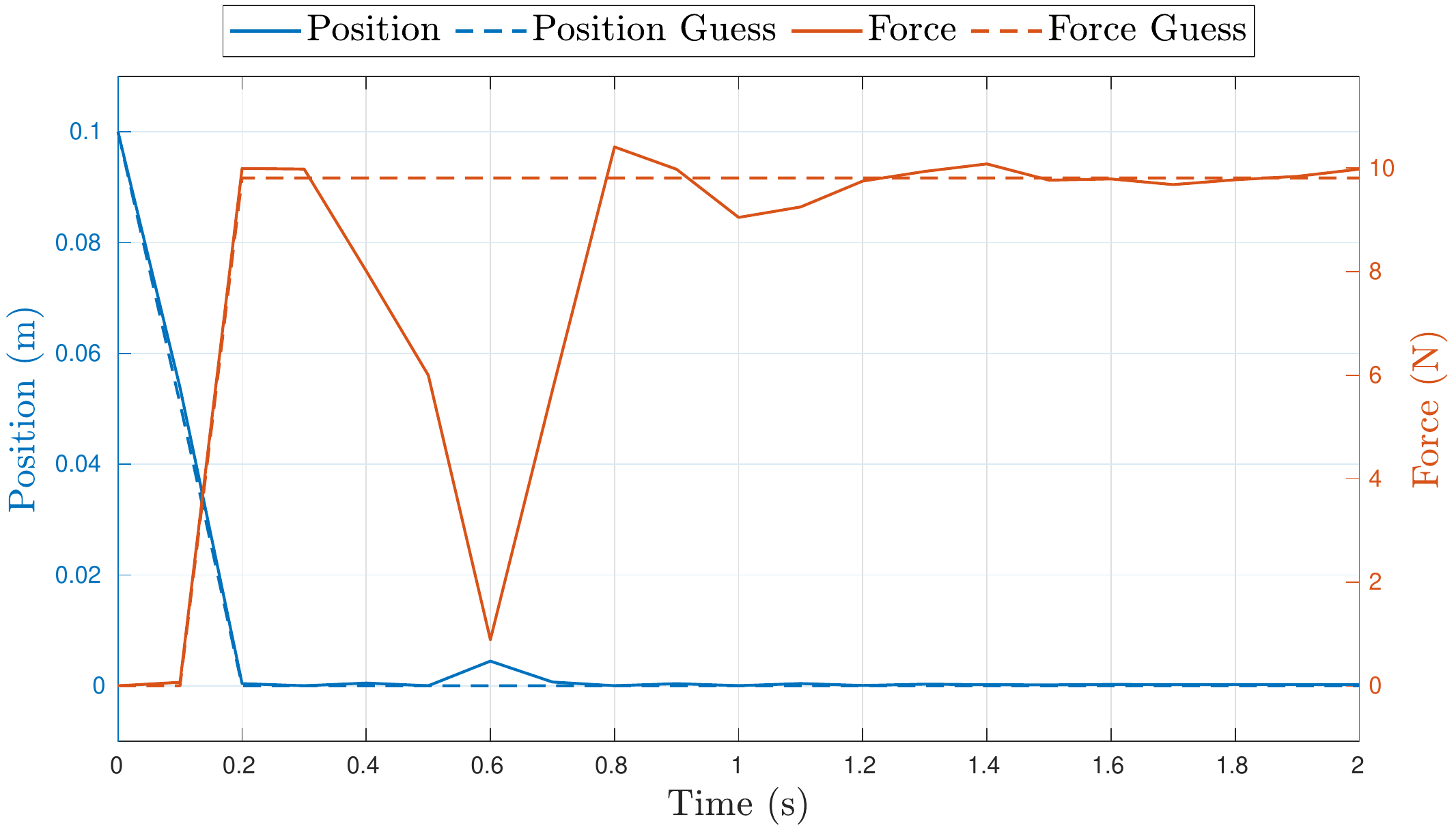}}

	\subfloat[\modtextBis{Dynamically enforced complementarity}] {\includegraphics[width=.75\columnwidth]{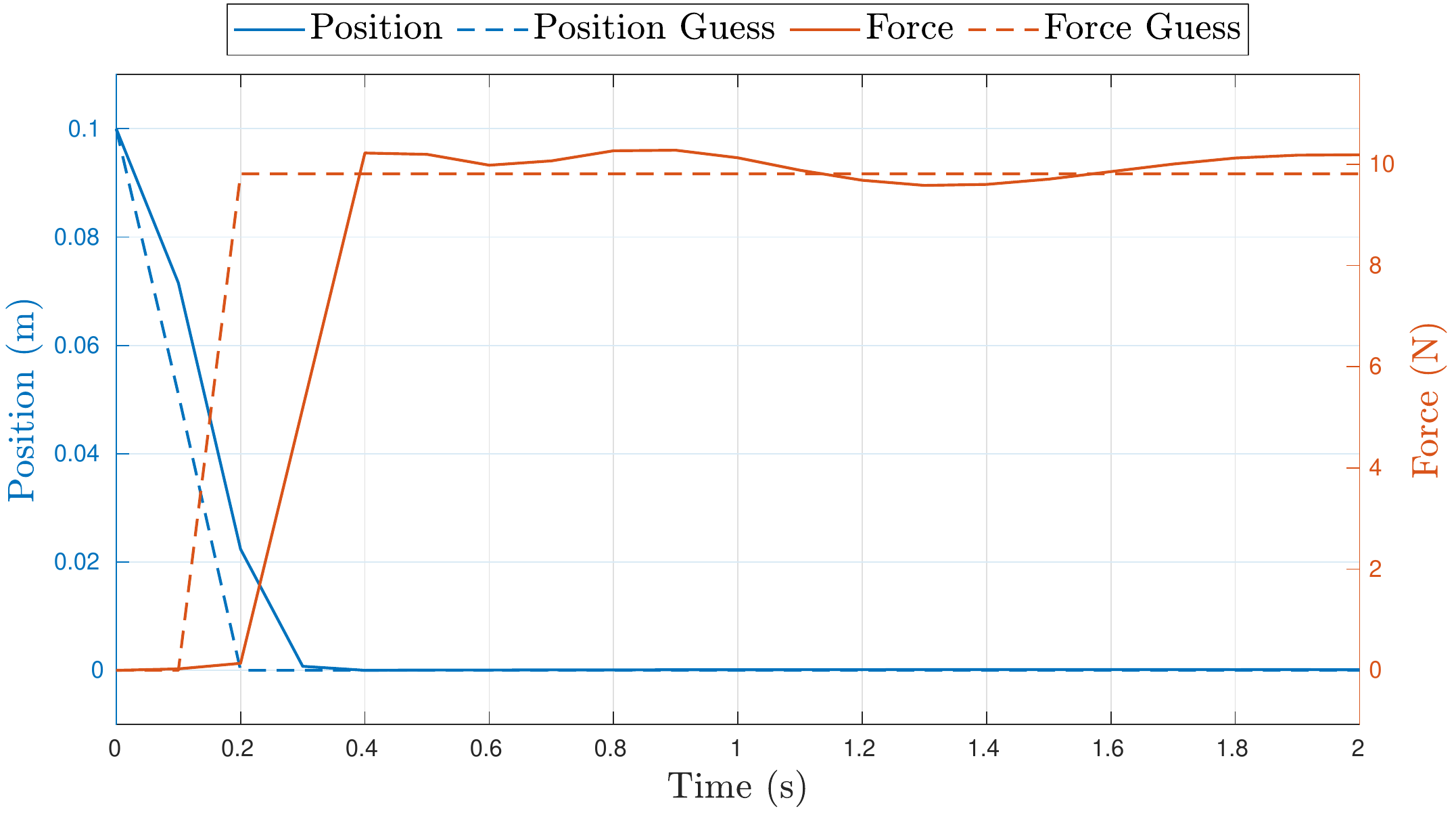}}
	
	\subfloat[\modtextBis{Hyperbolic secant in control bounds}] {\includegraphics[width=.75\columnwidth]{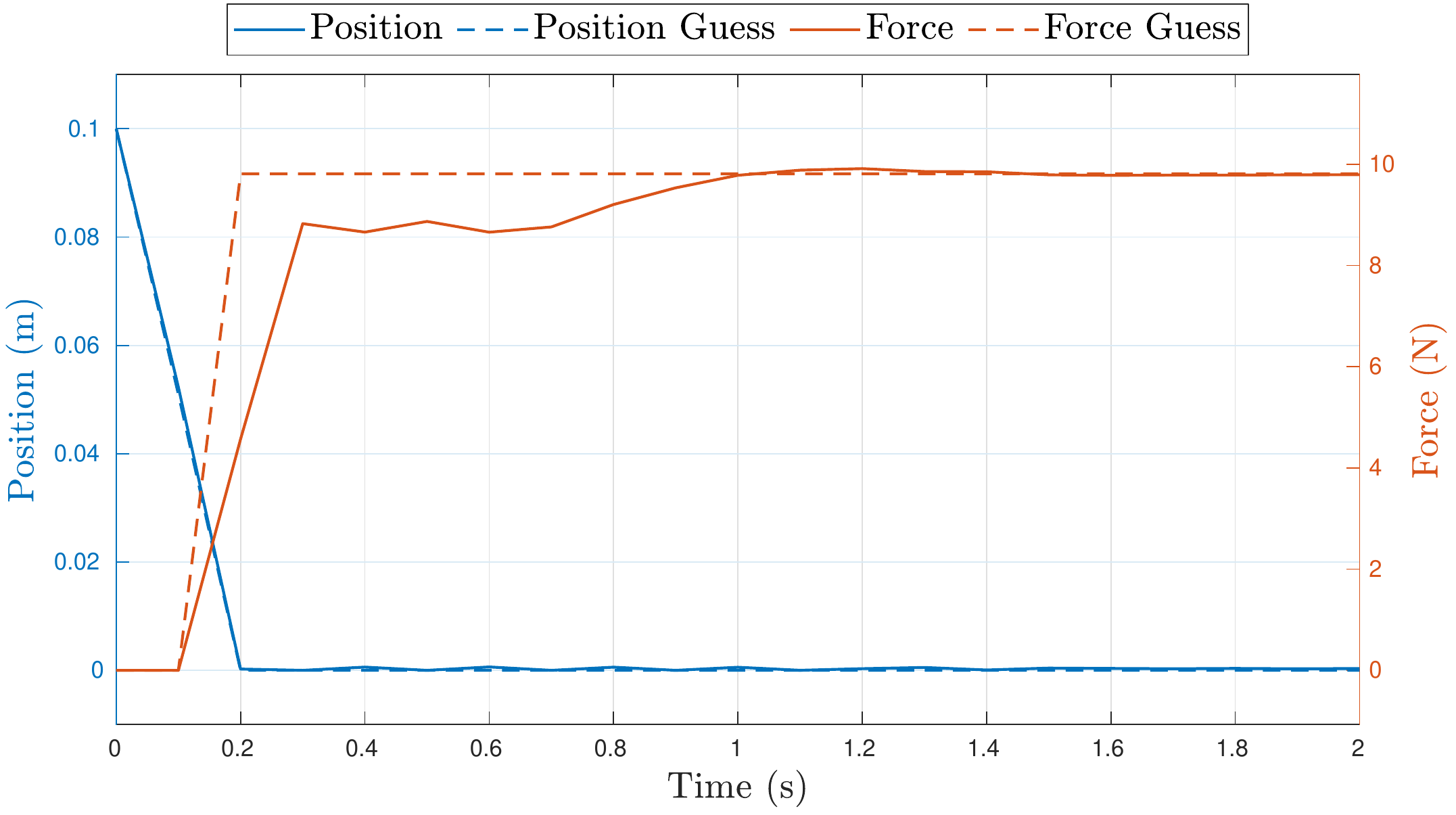}}
	\caption{\modtextBis{Planned mass position, and contact force of the toy problem adopting different complementarity constraints. The dashed lines correspond to the initialization provided to the optimal control problem.}}
	\label{fig:toy_pos_force}
\end{figure}

\begin{figure}[tpb]
	\centering
	\subfloat[\modtextBis{Relaxed complementarity}] {\includegraphics[width=.75\columnwidth]{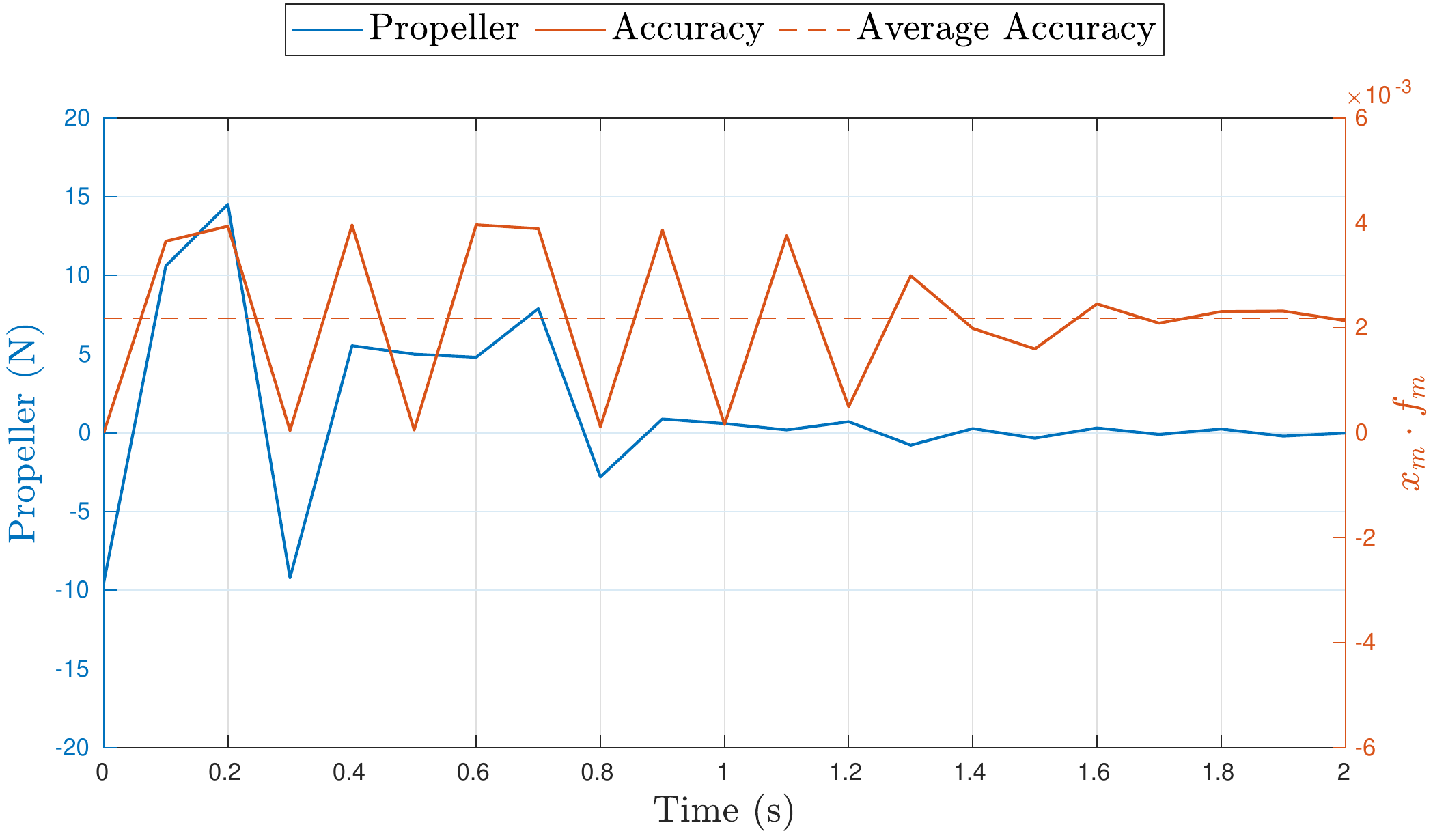}}

	\subfloat[\modtextBis{Dynamically enforced complementarity}] {\includegraphics[width=.75\columnwidth]{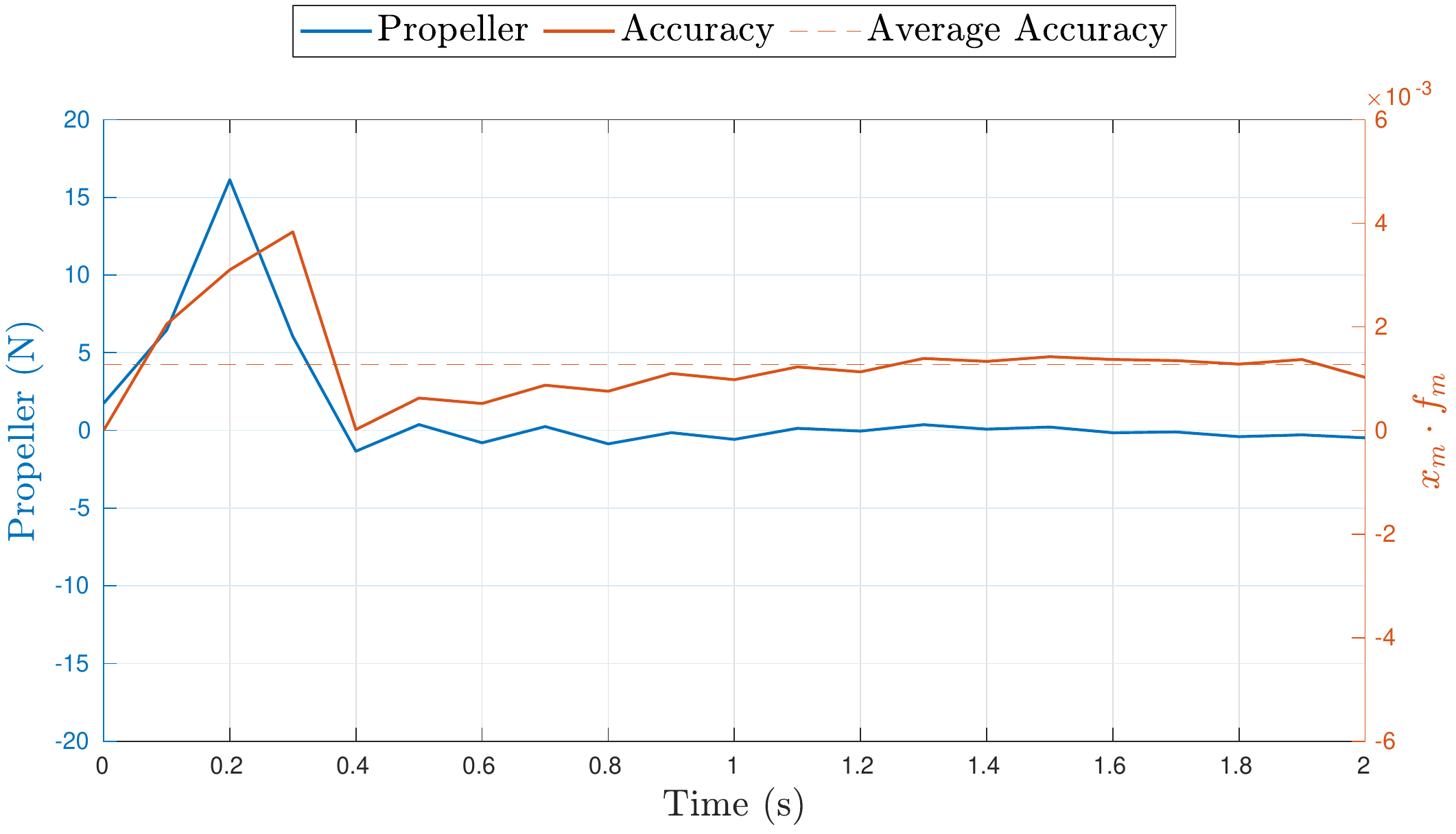}}
	
	\subfloat[\modtextBis{Hyperbolic secant in control bounds}] {\includegraphics[width=.75\columnwidth]{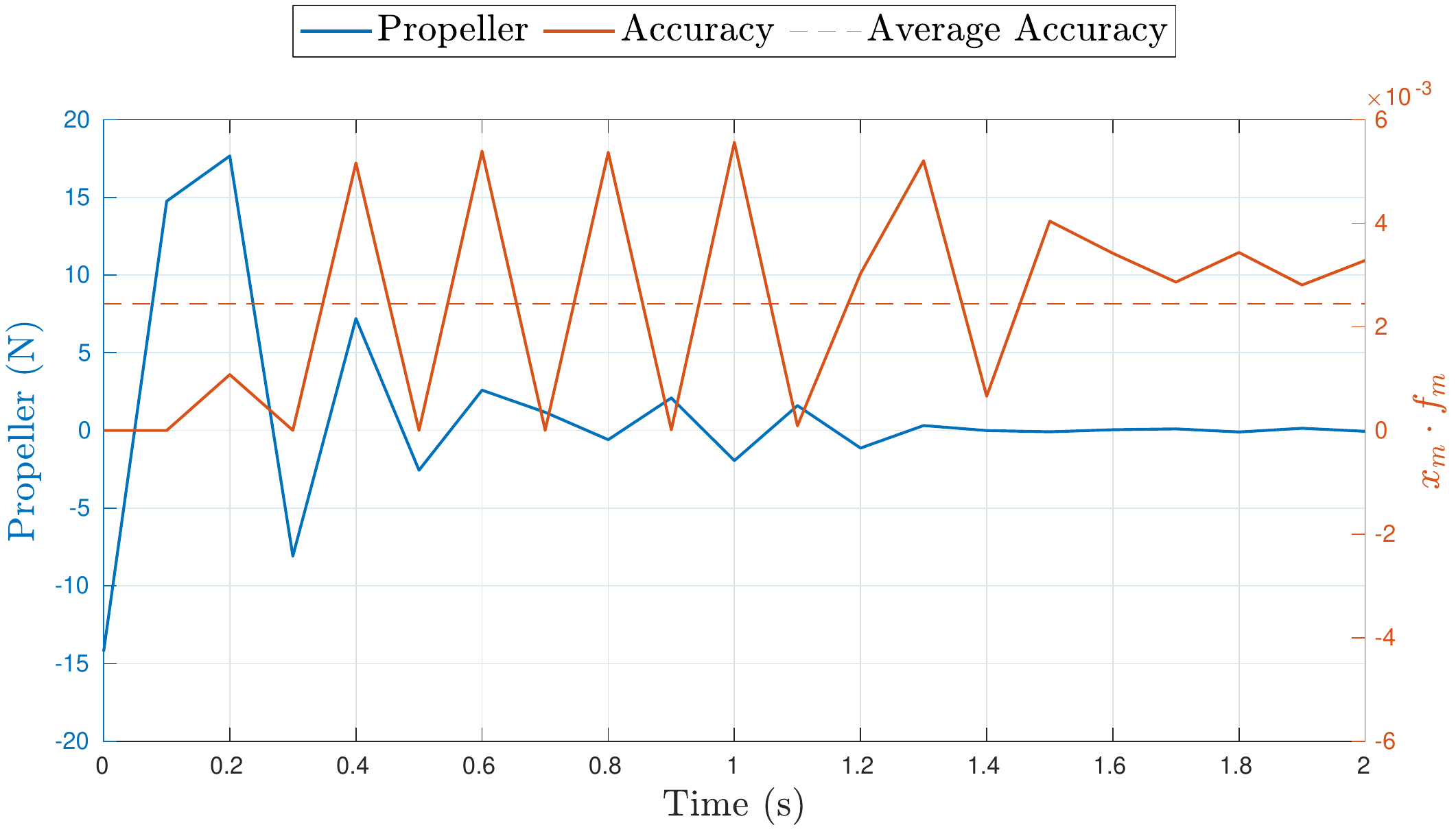}}
	\caption{\modtextBis{Propeller value, and complementarity accuracy, of the toy problem adopting different complementarity constraints. The dashed red line is the average accuracy.}}
	\label{fig:toy_accuracy}
\end{figure}

\modtextBis{
The optimal control problem described by Eq. \eqref{eq:simpleMassOc} is transcribed into an optimization problem using a Direct Multiple Shooting method and solved through \texttt{Ipopt}. We use a prediction window of 2$\mathrm{s}$, with an integration step of 100$\mathrm{ms}$ as in Sec. \ref{sec:validation}. The initial guess of the optimization problem is as follows. We compute the expected mass position and velocity assuming no propeller is used. The initial guess for the position and velocity is set to zero after the impact. Similarly, the guess for contact force is set equal to the weight after the impact.}

\modtextBis{Differently from Sec. \ref{sec:validation}, the implementation has been fast-prototyped using \texttt{MATLAB}, exploiting the \texttt{CasADi} framework \cite{andersson2019casadi}. This excludes possible implementation-specific biases.
}

\modtextBis{Fig. \ref{fig:toy_pos_force} presents the output of the trajectory optimization for a unitary mass falling from 10$\mathrm{cm}$. We present the results using the different complementarity methods. The dashed lines correspond to the initialization fed to the optimizer. Fig. \ref{fig:toy_accuracy} shows the accuracy achieved by each method and the use of the propeller. The red dashed line displays the average accuracy.}

\modtextBis{The major complexity of the problem is represented by the complementarity constraints, as all the other constraints are linear. Hence, any difference in the output can be safely ascribed to the chosen complementarity method. In the following we analyze the perfomance of each complementarity method in case of parameter variations.}

\modtextBis{\subsubsection{Comparison using different initial positions}}

\begin{table*}[tbp]
	\centering
	\begin{tabular}{ r || c c c | c c c | c c c}
		\multicolumn{10}{c}{Test running on a laptop using the \texttt{MA57} linear solver} \Tstrut\Bstrut\\
		\hline
		\multicolumn{1}{c||}{\modtextBis{Initial Height}}& \multicolumn{3}{c|}{\modtextBis{Relaxed}} &\multicolumn{3}{c|}{\modtextBis{Dynamically enforced}} & \multicolumn{3}{c}{\modtextBis{Hyperbolic secant}} \Tstrut\Bstrut\\
		\hline
		& Comp Time & Avg Accuracy & Cost & Comp Time & Avg Accuracy & Cost & Comp Time & Avg Accuracy & Cost \Tstrut\\
		&  [$\mathrm{s}$]  & [$\mathrm{N}\cdot \mathrm{m}$] & [$\mathrm{N}^2$] & [$\mathrm{s}$] & [$\mathrm{N}\cdot \mathrm{m}$] & [$\mathrm{N}^2$] & [$\mathrm{s}$] & [$\mathrm{N}\cdot \mathrm{m}$] &  [$\mathrm{N}^2$]\\
		\modtextBis{5$\mathrm{cm}$}  & 0.0779 & 0.0021 & \textbf{218.31} & 0.0848          & 0.0013          & 298.48          & \textbf{0.0509} & \textbf{0.0001} & 1493.0\Tstrut\\
		\modtextBis{7$\mathrm{cm}$}  & 0.0680 & 0.0021 & 366.36          & 0.0702          & \textbf{0.0019} & \textbf{305.35} & \textbf{0.0525} & 0.0026          & 458.95       \\
		\modtextBis{9$\mathrm{cm}$}  & 0.1040 & 0.0021 & 602.05          & \textbf{0.0808} & \textbf{0.0013} & \textbf{325.93} & 0.5353          & 0.0021          & 696.14       \\ 
		\modtextBis{11$\mathrm{cm}$} & 0.0758 & 0.0020 & \textbf{363.75} & \textbf{0.0476} & \textbf{0.0012} & 365.32          & 0.0609          & 0.0025          & 1060.9       \\
		\modtextBis{13$\mathrm{cm}$} & 0.0709 & 0.0021 & \textbf{390.14} & \textbf{0.0437} & \textbf{0.0013} & 429.06          & 0.0598          & 0.0025          & 1500.0       \\
		\modtextBis{15$\mathrm{cm}$} & 0.0630 & 0.0020 & \textbf{458.18} & \textbf{0.0547} & \textbf{0.0011} & 469.49          & 0.1627          & 0.0017          & 1110.2\Bstrut\\ 
		\hline
	\end{tabular}

    \vspace*{.1 cm}

	\begin{tabular}{ r || c c c | c c c | c c c}
		\multicolumn{10}{c}{Test running on \texttt{Github Actions} using the \texttt{MUMPS} linear solver} \Tstrut\Bstrut\\
		\hline
		\multicolumn{1}{c||}{\modtextBis{Initial Height}}& \multicolumn{3}{c|}{\modtextBis{Relaxed}} &\multicolumn{3}{c|}{\modtextBis{Dynamically enforced}} & \multicolumn{3}{c}{\modtextBis{Hyperbolic secant}} \Tstrut\Bstrut\\
		\hline
		& Comp Time & Avg Accuracy & Cost & Comp Time & Avg Accuracy & Cost & Comp Time & Avg Accuracy & Cost \Tstrut\\
		&  [$\mathrm{s}$]  & [$\mathrm{N}\cdot \mathrm{m}$] & [$\mathrm{N}^2$] & [$\mathrm{s}$] & [$\mathrm{N}\cdot \mathrm{m}$] & [$\mathrm{N}^2$] & [$\mathrm{s}$] & [$\mathrm{N}\cdot \mathrm{m}$] &  [$\mathrm{N}^2$]\\
		\modtextBis{5$\mathrm{cm}$}  & 0.0756          & 0.0021 & \textbf{219.43} & 0.0586          & 0.0013          & 288.55          & \textbf{0.0509} & \textbf{0.0001} & 1493.0\Tstrut\\
		\modtextBis{7$\mathrm{cm}$}  & 0.2036          & 0.0020 & 347.64          & \textbf{0.0656} & \textbf{0.0015} & \textbf{335.32} & 0.0770          & 0.0026          & 458.95       \\
		\modtextBis{9$\mathrm{cm}$}  & 0.1231          & 0.0021 & 600.02          & \textbf{0.0449} & \textbf{0.0013} & \textbf{350.52} & 0.1831          & 0.0025          & 701.91       \\ 
		\modtextBis{11$\mathrm{cm}$} & 0.0927          & 0.0021 & \textbf{323.58} & \textbf{0.0475} & \textbf{0.0012} & 365.32          & 0.0907          & 0.0026          & 1061.2       \\
		\modtextBis{13$\mathrm{cm}$} & 0.0712          & 0.0020 & \textbf{403.46} & \textbf{0.0426} & \textbf{0.0011} & 420.55          & 0.0923          & 0.0025          & 1500.3       \\
		\modtextBis{15$\mathrm{cm}$} & \textbf{0.0578} & 0.0020 & \textbf{458.10} & 0.0587          & \textbf{0.0012} & 473.44          & 0.1744          & 0.0020          & 1120.5\Bstrut\\ 
		\hline
	\end{tabular}
	\caption{\modtextBis{The computational time, the average accuracy, and the cost of each complementarity method when solving the toy problem at different initial heights. The results are the average of 100 repetitions. In bold the best resulting method by row.}}
	\label{tab:toy_varying_height}
\end{table*}

\modtextBis{
In this section, we analyze the performance of the different complementarity methods by varying the initial mass height. Since the toy problem can be seen as an approximation of a robot foot approaching the ground, we consider an initial height that is compatible with the step heights achievable by a small humanoid robot.
Table \ref{tab:toy_varying_height} presents the computational time, the average accuracy of the output trajectory, and the corresponding cost when using different complementarity methods, averaged over 100 repetitions. We highlight in bold the lowest computational time, the best accuracy, and the lowest cost for each initial height. Overall, the \emph{Dynamically enforced} method is the one providing best accuracy, except for the 5$\mathrm{cm}$ case, where the \emph{Hyperbolic secant} method performs better. The same result holds also when running the test on \texttt{Github Actions}.
For what concerns the computational time, the \emph{Hyperbolic secant} method performs better for heights lower than 9$\mathrm{cm}$, while the \emph{Dynamically enforced} is the fastest in the other cases, with the exception of the 15$\mathrm{cm}$ height. In fact, in the test running on \texttt{Github Actions}, the \emph{Relaxed} method performed better. This latter method, is also the one providing the smallest cost in the majority of the cases. We can notice a correlation between the accuracy and the cost value. In fact, a higher use of the propeller reduces the time in which the mass is in contact with the ground, resulting in a lower average accuracy. Hence, there is a trade-off between the use of the propeller and the accuracy. Such trade-off is visible in the 5$\mathrm{cm}$ case, where the \emph{Hyperbolic secant} method has very little accuracy, but with the highest cost. Another thing to notice is that the cost increases with the initial height. In fact, higher initial heights involve higher impact velocities that are counteracted by the use of the propeller.

Finally, we highlight that the \emph{Dynamically enforced} method is the one providing the least variations in terms of computational time across different heights.
}

\modtextBis{\subsubsection{Comparison after parameter variations} \label{sec:toy_parameters_variation}}
\begin{figure}[tpb]
    \centering
    \subfloat[Test running on a laptop using the \texttt{MA57} linear solver]{\includegraphics[width=0.9\columnwidth]{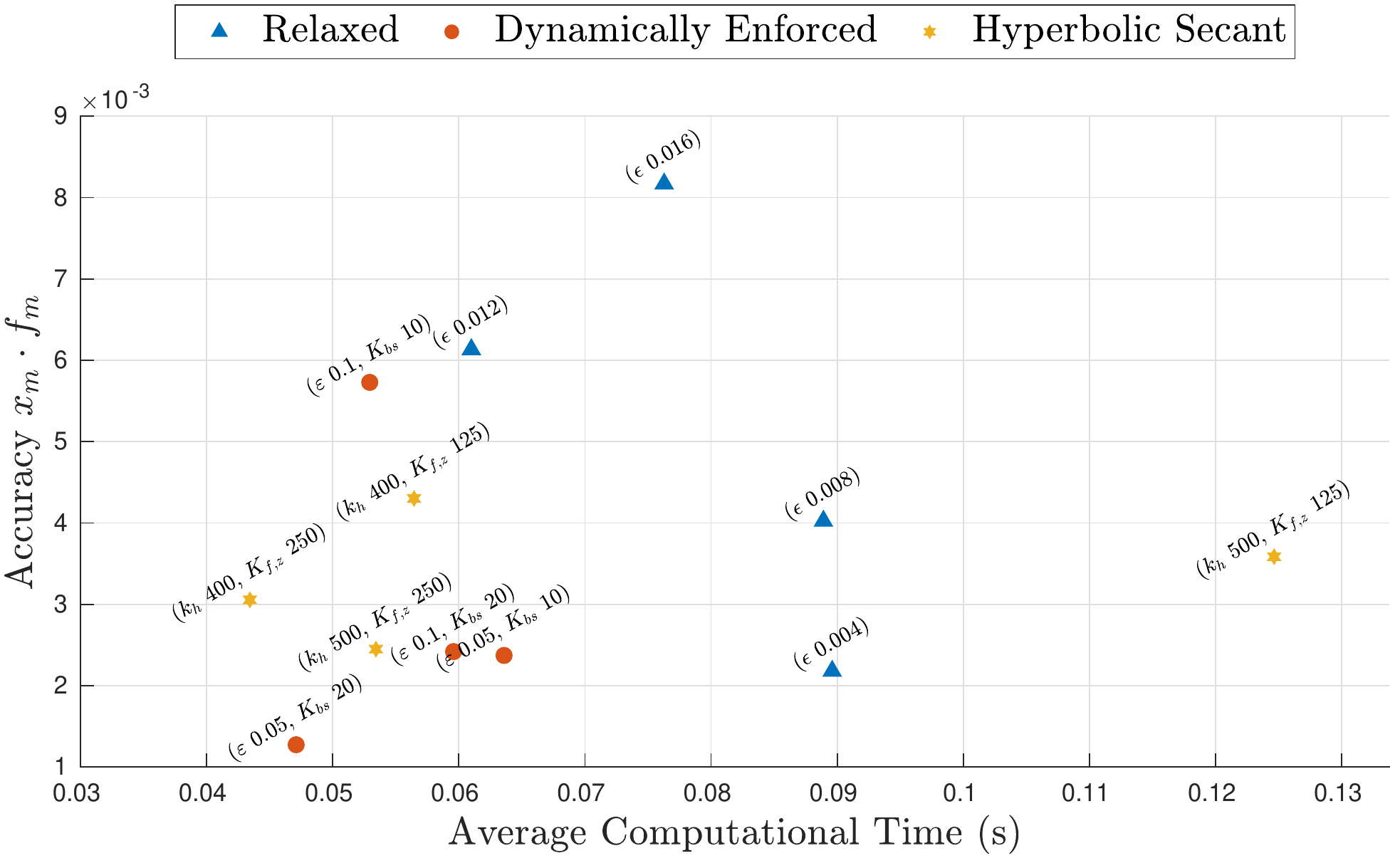}}
    
    \subfloat[Test running on \texttt{Github Actions} using the \texttt{MUMPS} linear solver]{\includegraphics[width=0.9\columnwidth]{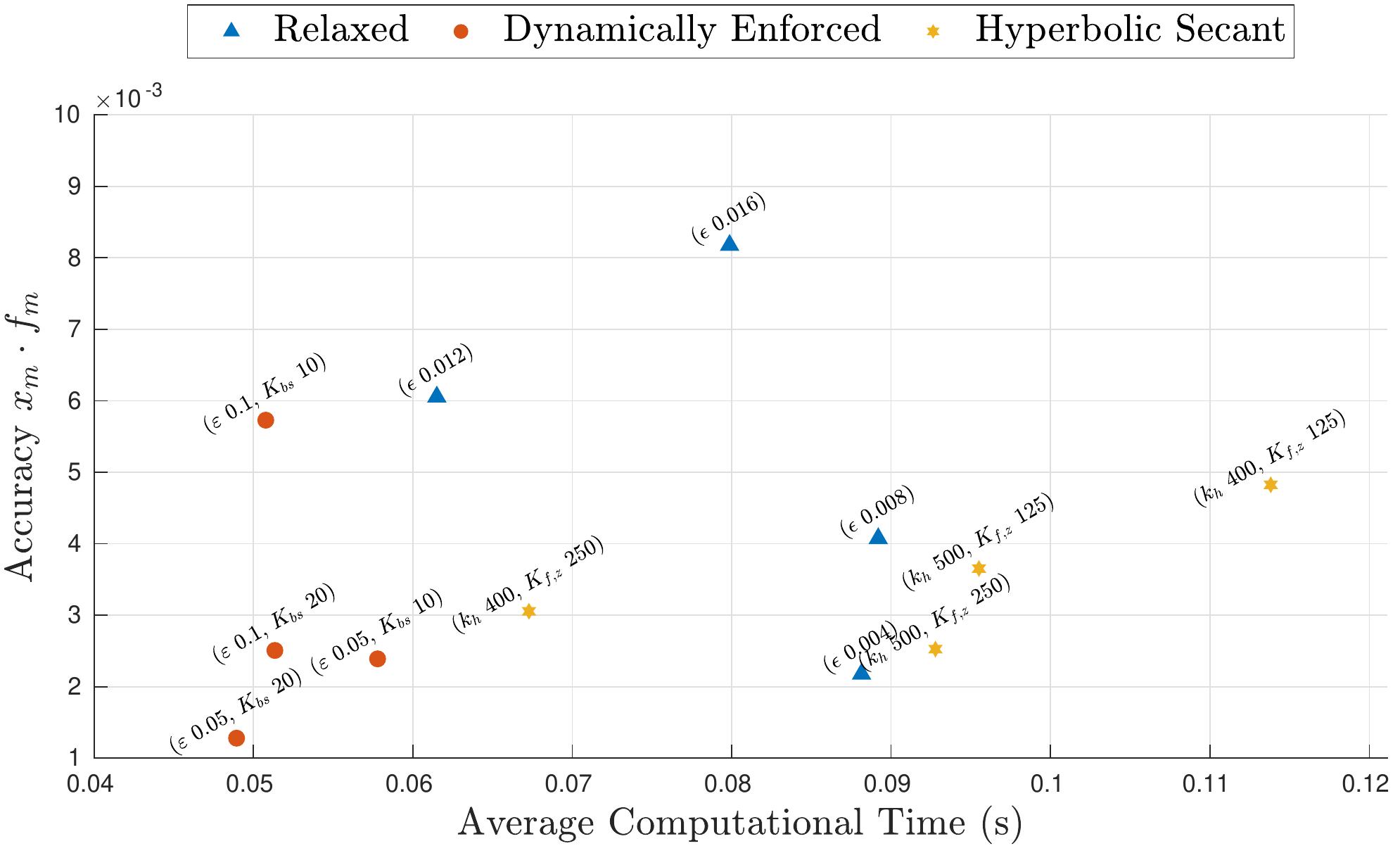} \label{fig:toy_parameters_variation_toy}}
    \caption{\modtextBis{Performance of the different complementarity methods in the toy problem after parameter changes. The closer to the origin, the better is the performance.}}
    \label{fig:toy_parameters_variation}
\end{figure}

\modtextBis{
In this section, we compare the complementarity methods by varying the following parameters associated with each complementarity method:
\begin{itemize}
    \item $\epsilon$ for the \emph{Relaxed Complementarity};
    \item $K_\text{bs}$, $\varepsilon$ for the \emph{Dynamically Enforced Complementarity};
    \item $K_{f,z}$ and $k_h$ for the \emph{Hyperbolic secant in control bounds}.
\end{itemize}
The results are depicted in Fig. \ref{fig:toy_parameters_variation}, where each point has an associated label displaying the parameters being used. The initial mass height is set to 0.1$\mathrm{cm}$ and the points represent the average result over 100 repetitions. The \emph{Dynamically enforced} method with the parameters $(\varepsilon~0.05,~K_{bs}~20)$ is the one providing the best accuracy. At the same time, the \emph{Hyperbolic secant} method with parameters $(k_h~400,~K_{f,z}~250)$ is the one producing the fastest solution. In general, we can observe that the \emph{Dynamically enforced} and the \emph{Hyperbolic secant} method are those closer to the origin, thus providing the most accurate solutions at the lowest time. On the other hand, we can notice that the \emph{Hyperbolic secant} method is more sensitive to parameter variations, since the computational time can largely increase. Moreover, when using \texttt{MUMPS} on \texttt{Github Actions} the \emph{Hyperbolic secant} method is the one suffering the highest performance loss. Indeed, it can be noticed that its corresponding points appear to be shifted toward right in Figure \ref{fig:toy_parameters_variation_toy}. For what concerns the \emph{Relaxed complementarity} points, a higher $\epsilon$ corresponds to a lower accuracy, trading off with the computational time.
}

\modtext{\modtextBis{
\subsection{Comparison using the planner of Sec. \ref{sec:dcc_nlmpc}} \label{sec:comparison_nlp}}

		
	

\begin{figure}[tpb]
	\centering
	\subfloat[\modtext{Relaxed complementarity}] {\includegraphics[width=\columnwidth]{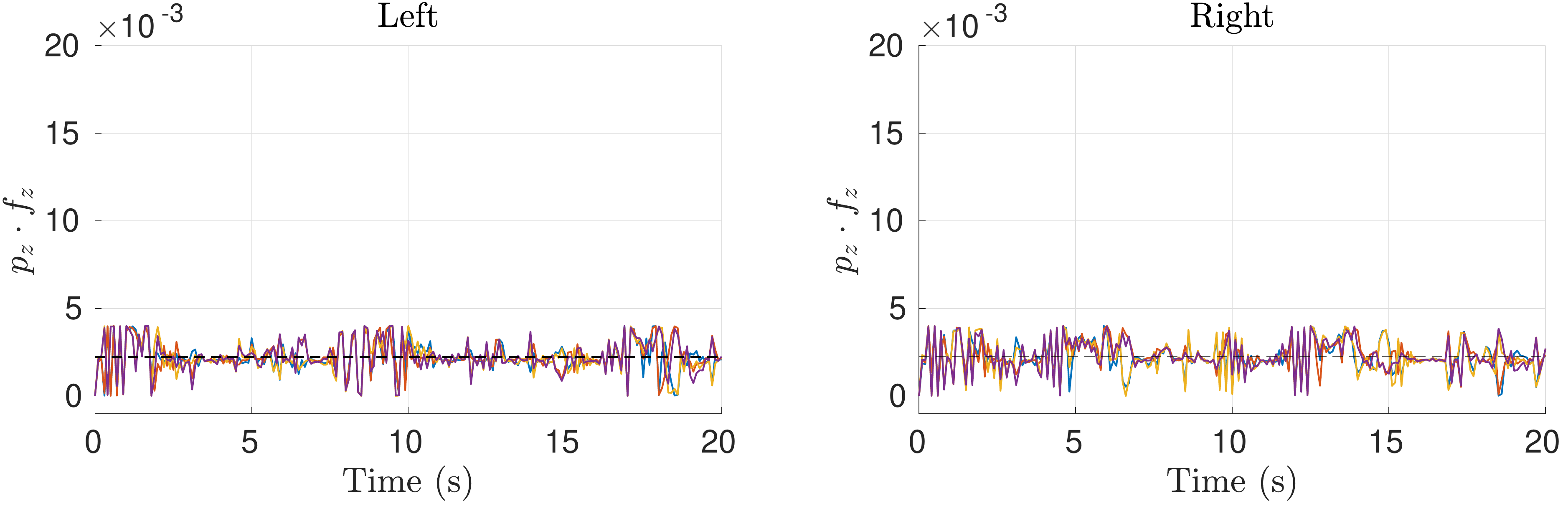}}

	\subfloat[\modtext{Dynamically enforced complementarity}] {\includegraphics[width=\columnwidth]{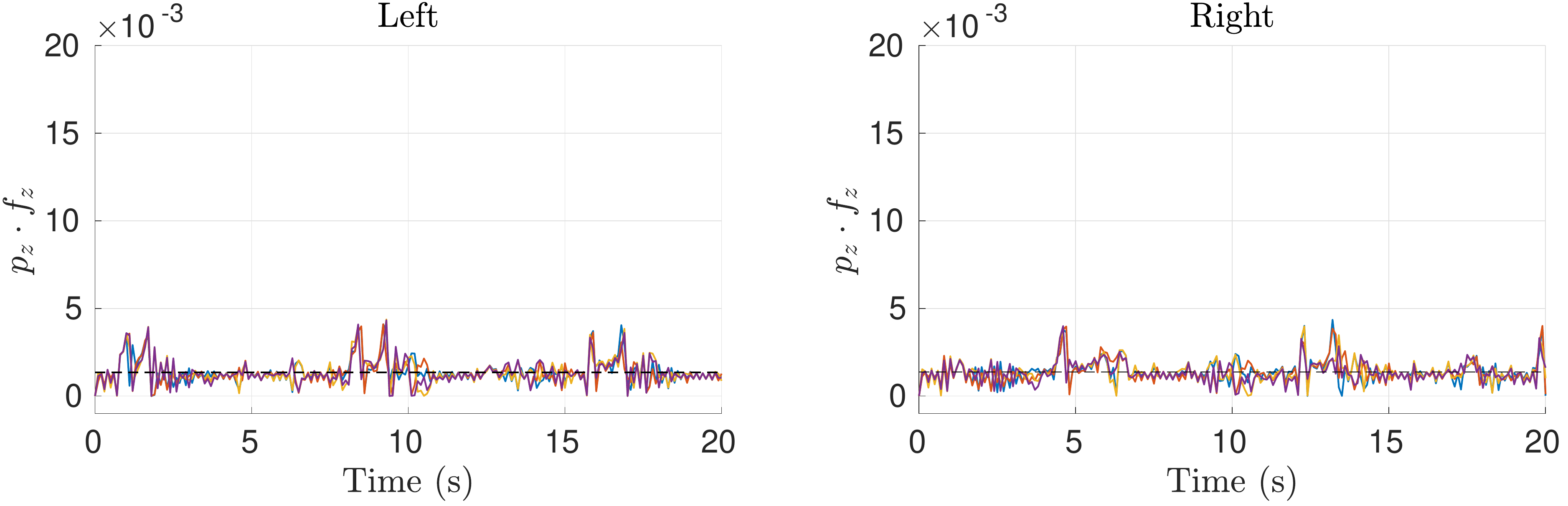}}
	
	\subfloat[\modtext{Hyperbolic secant in control bounds}] {\includegraphics[width=\columnwidth]{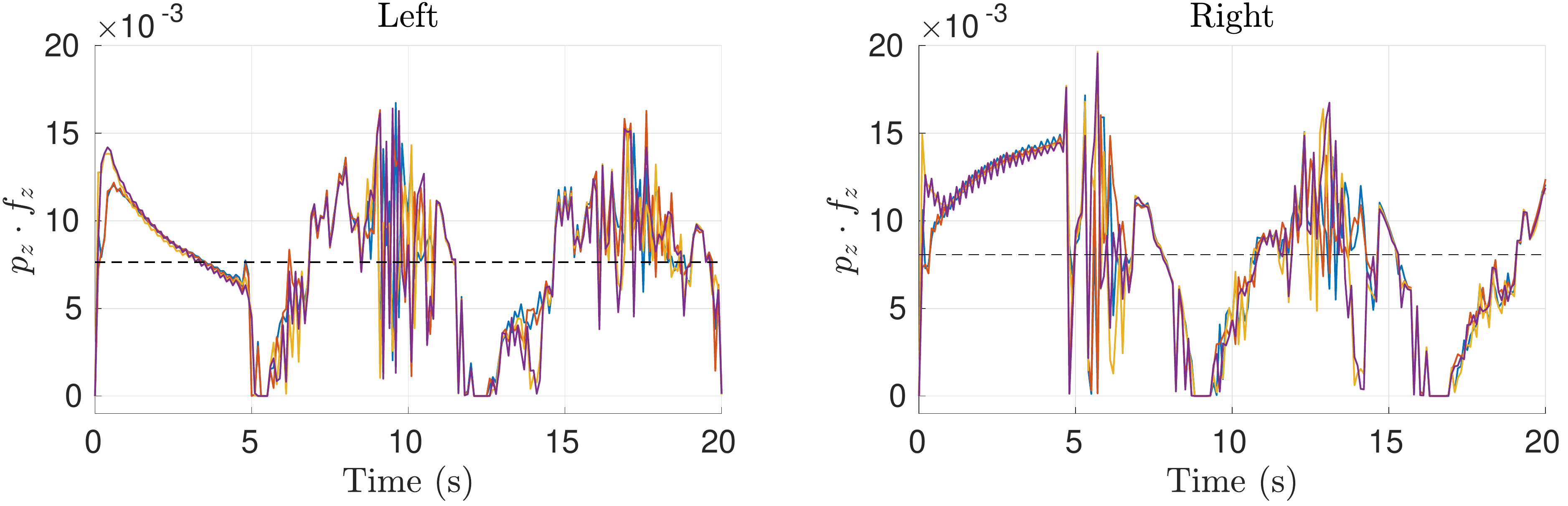}}
	\caption{Product between the vertical position and the normal force of each contact point, separated by foot, when using the different complementarity methods summarized in Sec. \ref{sec:complementarity_list} \modtextBis{in the setup of Sec. \ref{sec:validation}}. The black dashed lines indicate the mean values. By plotting the result of $p_z\cdot f_z$ for each point, we show the accuracy of each method in simulating a rigid contact.}
	\label{fig:complementarity_level}
\end{figure}

\begin{figure}[tpb]
	\centering
	\subfloat[\modtext{\modtextBis{Relaxed complementarity}}] {\includegraphics[width=.75\columnwidth]{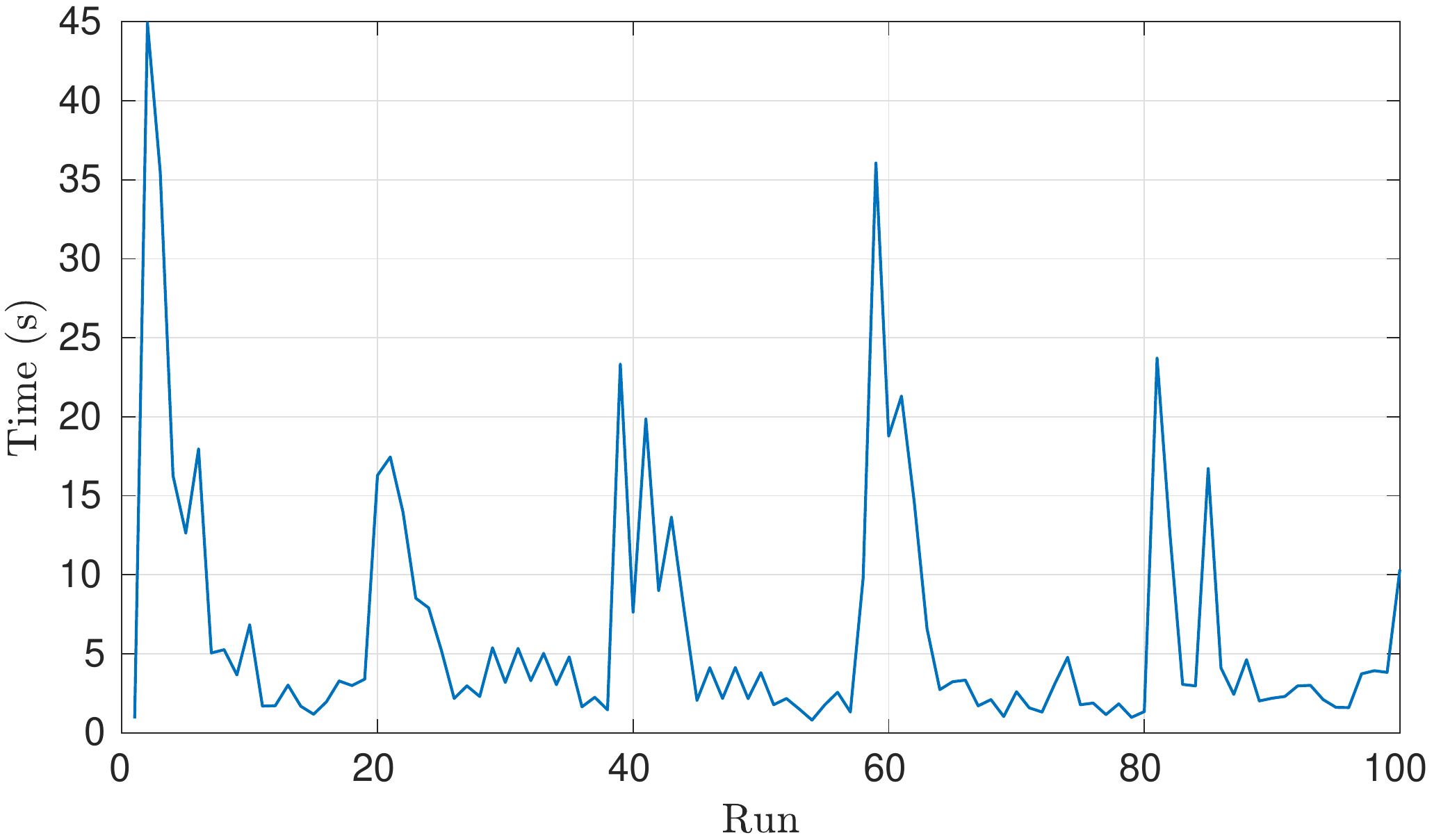}}
	
	\subfloat[\modtext{\modtextBis{Dynamically enforced complementarity}}] {\includegraphics[width=.75\columnwidth]{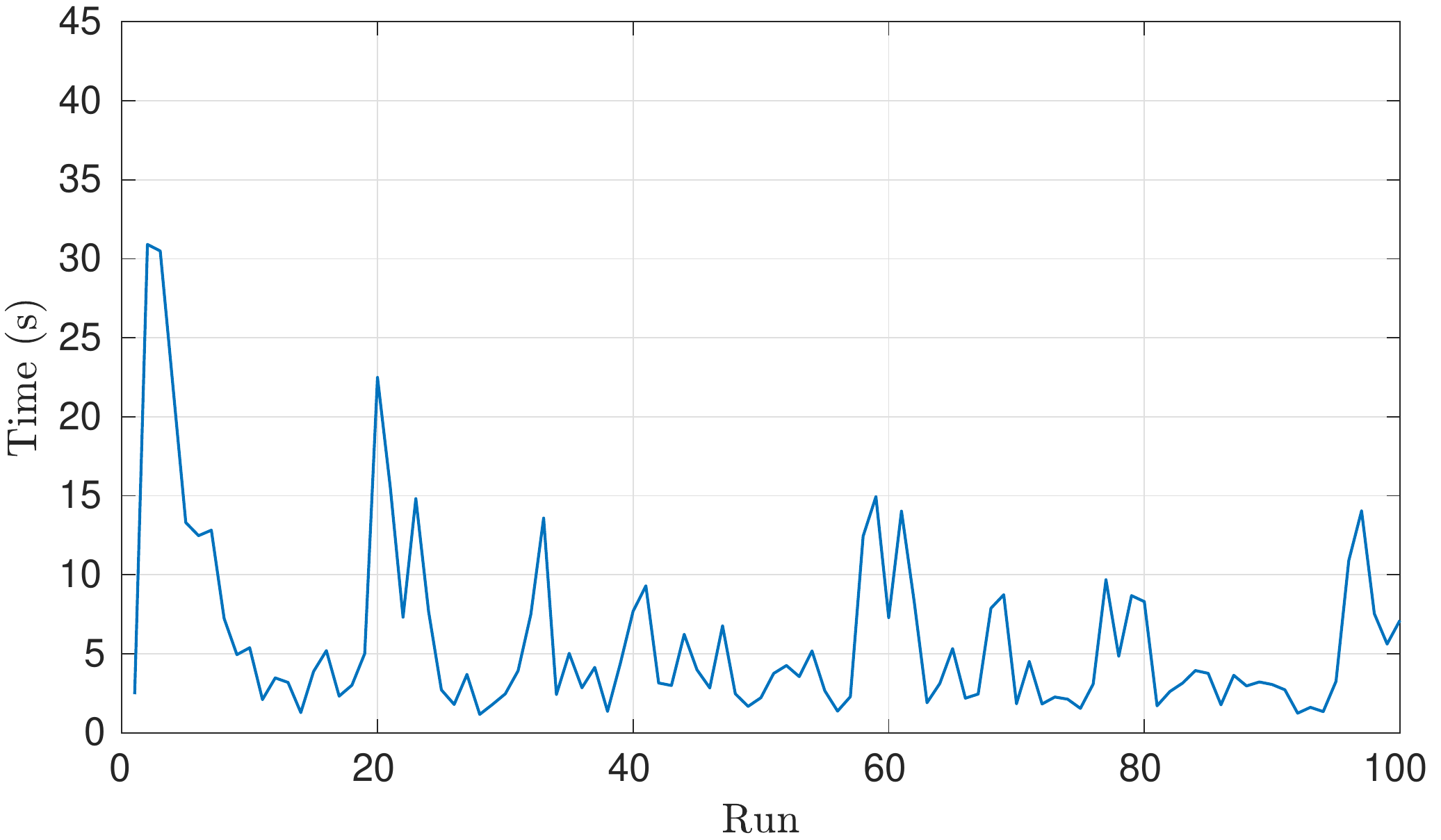}}
	
	\subfloat[\modtext{\modtextBis{Hyperbolic secant in control bounds}}] {\includegraphics[width=.75\columnwidth]{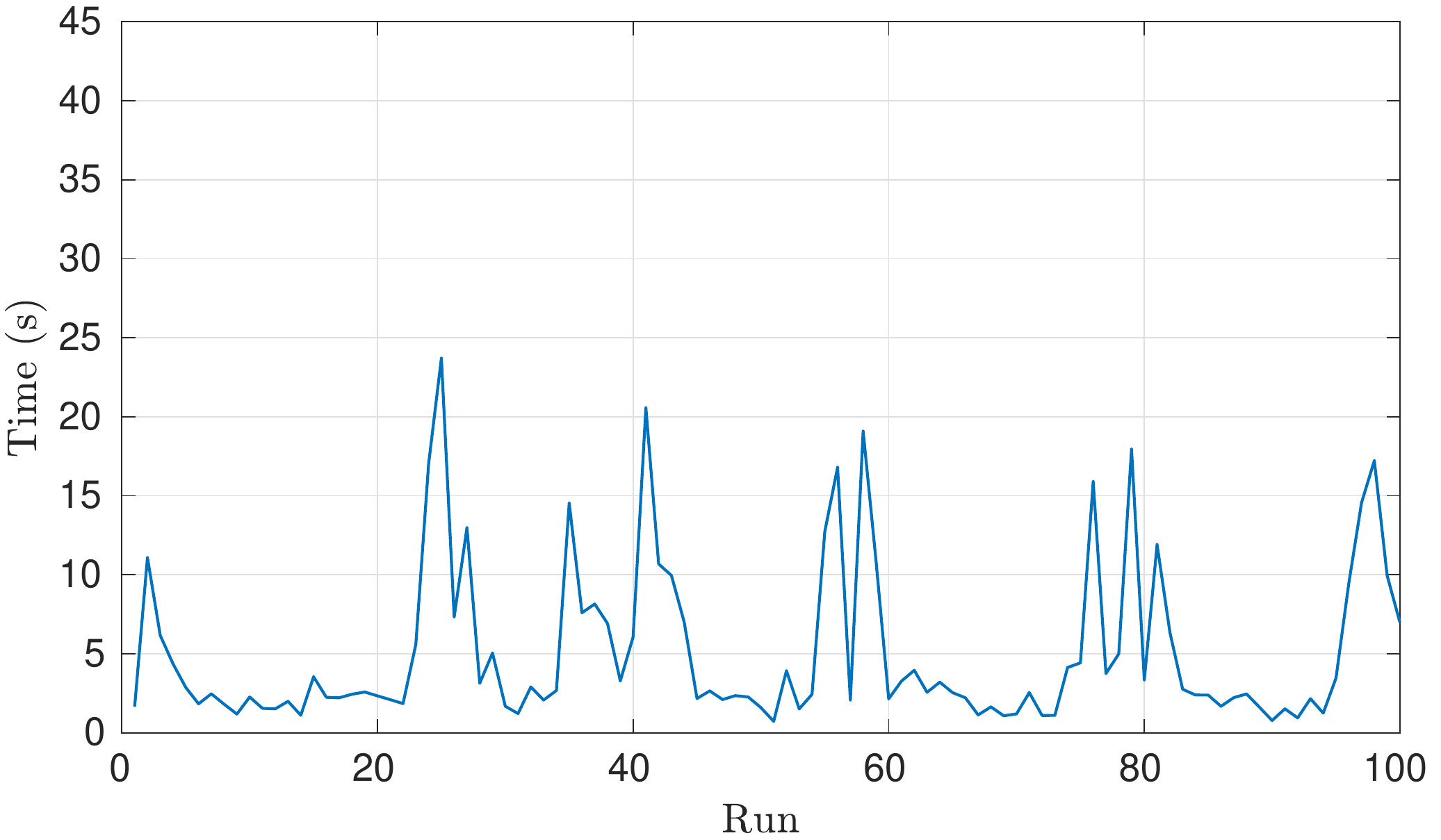}}
	\caption{The computational time required by each \modtextBis{non-linear trajectory planner run}. }
	\label{fig:computational_time}
\end{figure}

\modtextBis{In this section, we use the non-linear trajectory planning framework presented in Sec. \ref{sec:dcc_nlmpc} to compare the \emph{Dynamic Complementarity Conditions}}.
We perform a first comparison using the same setup presented in Sec. \ref{sec:validation}. The accuracy, namely the product ${}_ip_z\cdot{}_if_z$, is depicted in Fig. \ref{fig:complementarity_level} for each contact point of both feet, and numerically summarized in Table \ref{tab:accuracy}. The \emph{Dynamically enforced} complementarity method is the one with the best numeric performance, although very close to the \emph{Relaxed} complementarity method. In this context, the \emph{Hyperbolic secant} method is outperformed. The particularity of this method is that when the contact points are raised from the ground, the force drops to zero consistently on all the points, as it is possible to notice from Fig. \ref{fig:complementarity_level}. }
In fact, with this method, we force the normal force derivative to be strongly negative when the point is not on the ground. At the same time, this method does not prevent the point to move when a force is applied, possibly explaining the worst average accuracy compared to the other methods.

\modtext{Fig. \ref{fig:computational_time} presents the computational time required by the optimizer to complete the 100 \modtextBis{planner runs} that define the walking trajectory presented in Sec. \ref{sec:validation}. Table \ref{tab:dp_timings} presents the numerical results. In this case, the \emph{Hyperbolic secant} method is the best performing, requiring in average more than a second less compared to worst performing method, the \emph{Relaxed} complementarity. Nonetheless, a reason for this difference of performance can be explained by the longer initial double support phase observed when using the \emph{Hyperbolic secant} method. In fact, by looking at Fig. \ref{fig:computational_time}, it is possible to notice that the computational time in the first \modtextBis{runs} was much lower than the other methods. For all the methods it is possible to notice some pattern in the peaks. In particular, we have a high variation in the computational time when we move the reference for the centroid of the contact points. Hence the planner has to predict a full new step.}
Since we initialize the planner with the previously computed solution, in this instant the optimal point is far from the initialization. Hence, more time is required to find a solution. This issue can be addressed by providing the planner with a more effective initialization.

\begin{table*}[tbp]
	\centering
	\begin{tabular}{ l || c c | c c | c c }
		\multicolumn{1}{c||}{$p_z \cdot f_z$}& \multicolumn{2}{c|}{Relaxed} &\multicolumn{2}{c|}{Dynamically enforced} & \multicolumn{2}{c}{Hyperbolic secant} \Tstrut\Bstrut\\
		\hline
		         &  Left  & Right  & Left   & Right  & Left   & Right \Tstrut\\
		Average  & \modtext{0.0022} & \modtext{0.0023} & \modtext{\modtextBis{\textbf{0.0014}}} & \modtext{\modtextBis{\textbf{0.0014}}} & \modtext{0.0076} & \modtext{0.0081}\Tstrut\\
		Variance & \modtext{0.0007} & \modtext{0.0007} & \modtext{\modtextBis{\textbf{0.0006}}} & \modtext{\modtextBis{\textbf{0.0005}}} & \modtext{0.0036} & \modtext{0.0042}\\
		Maximum  & \modtext{0.0040} & \modtext{0.0040} & \modtext{\modtextBis{\textbf{0.0039}}} & \modtext{\modtextBis{\textbf{0.0038}}} & \modtext{0.0148} & \modtext{0.0172}\Bstrut\\ 
		\hline
	\end{tabular}
	\caption{Accuracy of the different complementarity methods \modtextBis{in the setup of Sec. \ref{sec:validation}. We consider the} left and right foot separately. \modtextBis{The best resulting method is in bold.}}
	\label{tab:accuracy}
\end{table*}

\begin{table}[tbp]
	\centering
	\begin{tabular}{ l || c | c | c }
		\multicolumn{1}{c||}{$[\mathrm{s}]$}& Relaxed & Dynamically  enforced & Hyperbolic secant \Tstrut\Bstrut\\
		\hline
		Average & \hphantom{0}\modtext{6.47} & \hphantom{0}\modtext{5.93} & \hphantom{0}\modtext{\modtextBis{\textbf{5.11}}} \Tstrut\\
		Variance & \hphantom{0}\modtext{7.99} & \hphantom{0}\modtext{5.65} & \hphantom{.}\modtext{\modtextBis{\textbf{5.17}}} \\
		Minimum & \hphantom{0}\modtext{0.80} & \hphantom{0}\modtext{1.17} & \hphantom{.}\modtext{\modtextBis{\textbf{0.73}}} \\
		Maximum & \modtext{44.93} & \hphantom{.}\modtext{30.91} & \hphantom{.}\modtext{\modtextBis{\textbf{23.71}}} \Bstrut\\ 
		\hline
	\end{tabular}
	\caption{Time performances using different complementarity methods \modtext{in the setup of Sec. \ref{sec:validation}}. All times are showed in seconds and obtained after \modtext{100} runs of the solver. \modtextBis{The best result for each row is in bold.}}
	\label{tab:dp_timings}
\end{table}

\modtext{
\subsubsection{Comparison with different walking velocities}
\begin{table}[tbp]
	\centering
	\begin{tabular}{ c || c | c | c }
		\multicolumn{4}{c}{Test running on a laptop using the \texttt{MA57} linear solver} \Tstrut\Bstrut\\
		\hline
		\multicolumn{1}{c||}{\modtext{Speed $[m/\mathrm{s}]$}}& \modtext{Relaxed} & \modtext{Dynamically  enforced} & \modtext{Hyperbolic secant} \Tstrut\Bstrut\\
		\hline
		\modtext{0.05} & 6.47 & 5.93 &  \hphantom{.}\textbf{5.11} \Tstrut\\
		\modtext{0.06} & $-$  & \textbf{6.89} &  7.16 \\
		\modtext{0.07} & 6.99 & \textbf{5.64} &  $-$ \Bstrut\\ 
		\hline
	\end{tabular}
	
	\vspace*{.1 cm}
	
	\begin{tabular}{ c || c | c | c }
		\multicolumn{4}{c}{Test running on \texttt{Github Actions} using the \texttt{MUMPS} linear solver} \Tstrut\Bstrut\\
		\hline
		\multicolumn{1}{c||}{\modtext{Speed $[m/\mathrm{s}]$}}& \modtext{Relaxed} & \modtext{Dynamically  enforced} & \modtext{Hyperbolic secant} \Tstrut\Bstrut\\
		\hline
		\modtext{0.05} & 13.19 & \textbf{9.88} &  \hphantom{i}11.70 \Tstrut\\
		\modtext{0.06} & \modtext{$-$} & \textbf{9.51} & 12.72 \\
		\modtext{0.07} & 11.57 & 13.07 &  \hphantom{i}\textbf{11.07} \Bstrut\\ 
		\hline
	\end{tabular}
	\caption{\modtext{Average computational time using different complementarity methods at different desired walking speeds. The symbol $-$ indicates that the planner was not able to plan a walking motion. The best result for each row is in bold.}}
	\label{tab:dp_velocity_comparison}
\end{table}
As a second comparison, we test the output of the planner at different desired walking speeds, measuring the corresponding computational time. Table \ref{tab:dp_velocity_comparison} presents the numerical results. Compared to Sec \ref{sec:validation}, the only parameter difference is the desired walking velocity. We notice that this (apparently) small change can have a strong impact, eventually making the planner to fail in finding walking patterns, keeping the robot constantly in double support. This has been the case for the \emph{Relaxed} complementarity method at $0.06 [m/s]$, and the \emph{Hyperbolic secant} method at $0.07 [m/s]$ (when using the \texttt{MA57} linear solver). In general, this behavior can be avoided with small variations of the initial guess, suggesting that the planner might get stuck in local minima. Nonetheless, for the sake of the comparison, we avoided additional changes. In this setting, the \emph{Dynamically enforced} method appears to be the most robust and consistent, even when running the test on different machines using different linear solvers.
}

\subsubsection{Comparison after parameter variations}
\begin{figure}[tpb]
    \centering
    \subfloat[Test running on a laptop using the \texttt{MA57} linear solver]{\includegraphics[width=0.91\columnwidth]{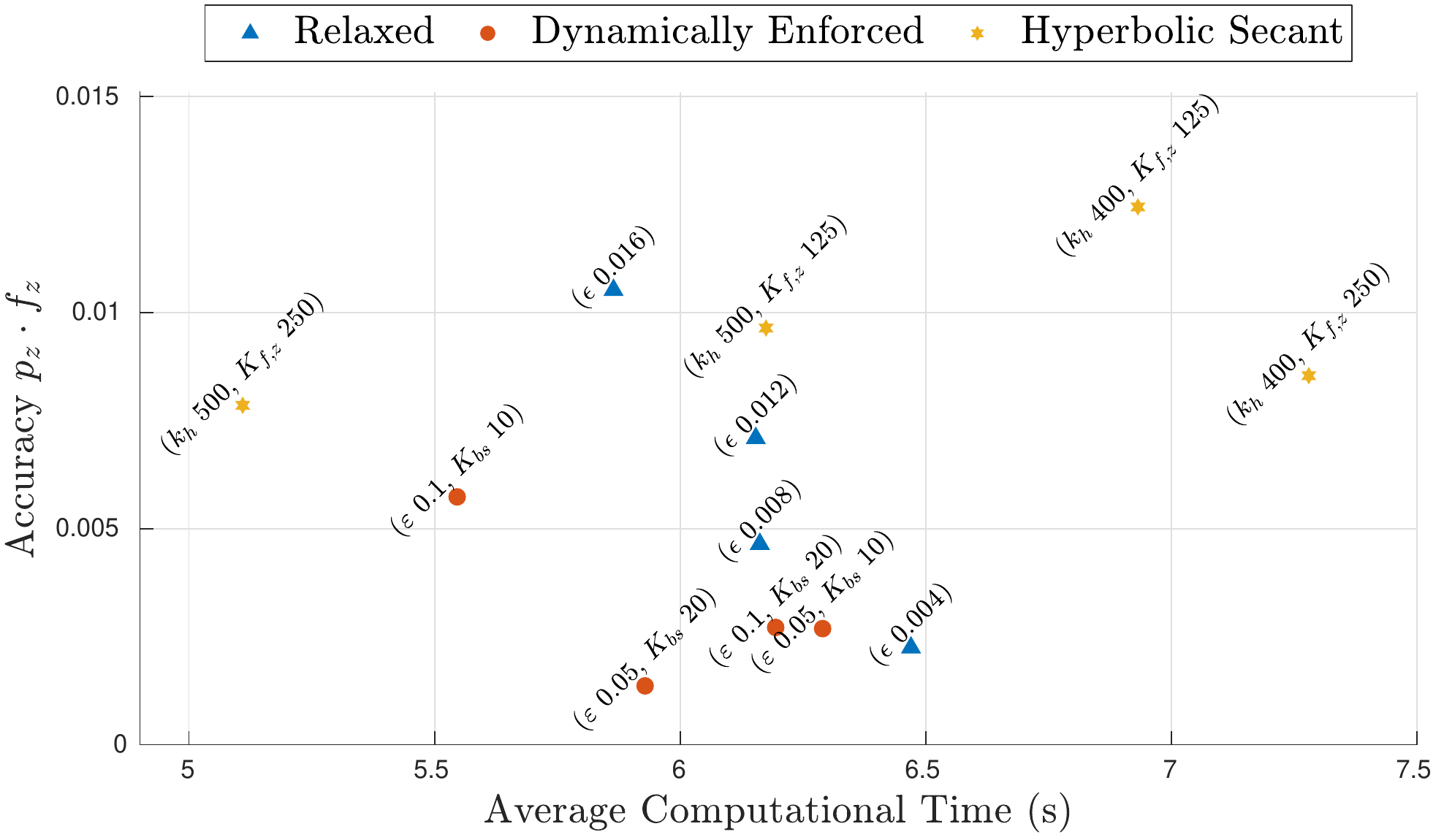}}
    
    \subfloat[Test running on \texttt{Github Actions} using the \texttt{MUMPS} linear solver]{\includegraphics[width=0.91\columnwidth]{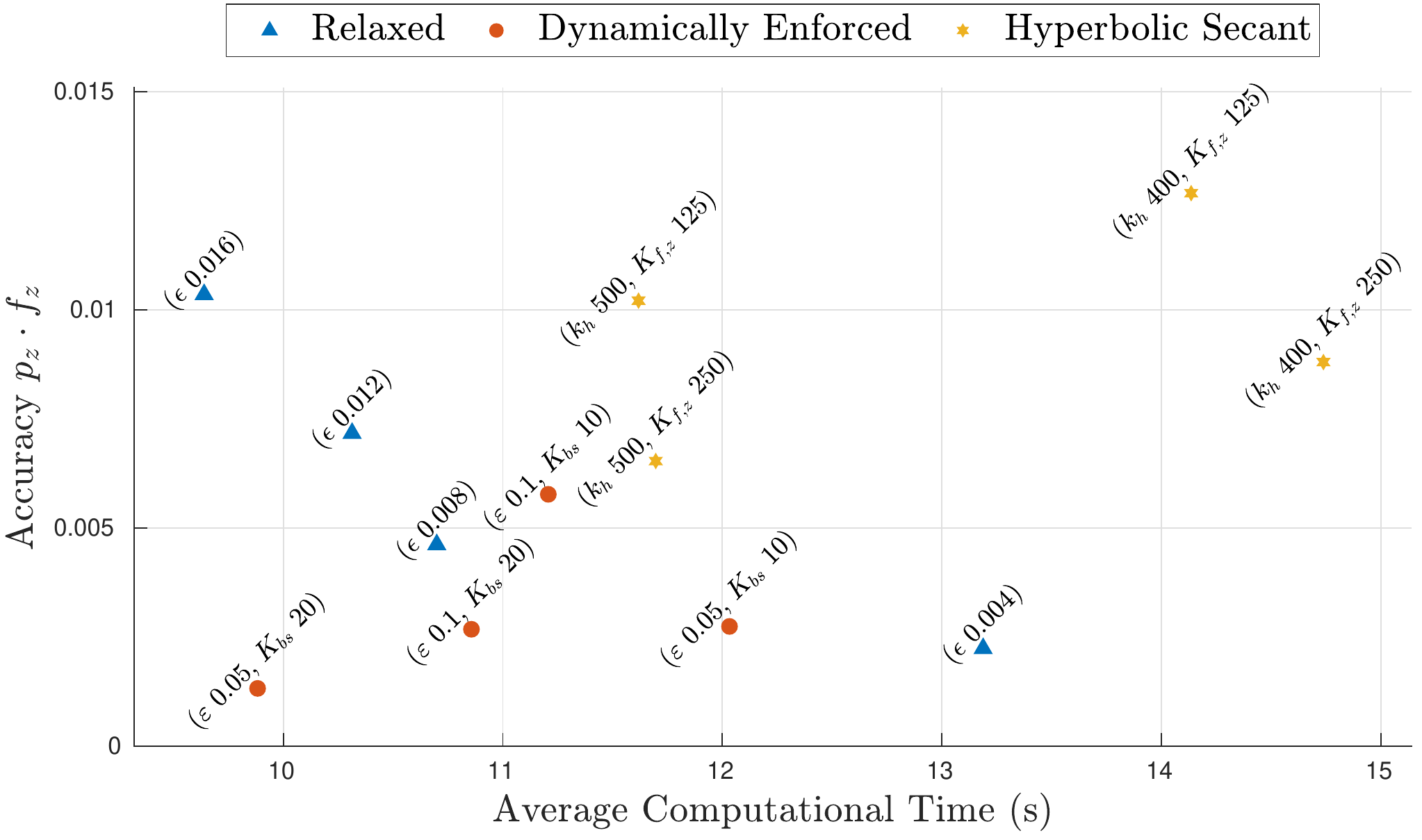}}

    \caption{\modtext{Performance of the different complementarity methods in case of parameter change \modtextBis{with respect to the setup of Sec. \ref{sec:validation}}. The closer to the origin, the better is the performance.}}
    \label{fig:parameters_variation}
\end{figure}
As a third comparison, we keep the desired forward velocity fixed while varying the same parameters as in Sec. \ref{sec:toy_parameters_variation}.
The results are depicted in Fig. \ref{fig:parameters_variation}, where each point is labeled with the parameters used. The plot presents the trade-off between the computational time and the accuracy. This is particularly evident by looking at the \emph{Relaxed} complementarity points. The higher the parameter $\epsilon$, the lower is the computational time, but at the same time the accuracy decreases. When running the test using a personal laptop running \texttt{Ipopt} with \texttt{MA57}, the \emph{Dynamically enforced} method seems to remain always below the line traced by these points, hence showing better overall performances in this setting. On the contrary, the \emph{Hyperbolic secant} method has proved to be very fast with the parameters used in Sec. \ref{sec:validation}, but the average computational time tends to increase considerably when changing the parameters.

Similarly to Figure \ref{fig:toy_parameters_variation_toy}, when running the test on \texttt{Github Actions}, the \emph{Hyperbolic secant} method points get shifted toward right, suggesting a higher perfomance loss compared to the other methods. Nonetheless, also in this case, the \emph{Dynamically enforced} method seems to provide the best trade-off between accuracy and computational time.

\section{Conclusions} \label{sec:conclusions}

This paper shows that walking trajectories can emerge by specifying a moving reference for the contact points' centroid and the desired CoM velocity only.
The planner considers relatively large time-steps. This enables the insertion of another control loop at higher frequency, whose goal is to stabilize the planned trajectories. 
The main bottleneck is represented by the computational time. A single planner \modtextBis{run} may take from slightly less than a second to minutes, also according to the machine used. This prevents an on-line implementation on the real robot. Nevertheless, the continuous formulation of the problem allows the application of techniques which solve the problem through the iterative application of fast LQR solvers \cite{neunert2018whole, farshidian2017real}.

\modtext{Among the different complementarity methods, the \emph{Dynamically enforced} complementarity method proved to be the best performing, \modtextBis{both in a toy problem, and when using the presented non-linear  trajectory  planning  framework}. In particular, it is the most accurate, and consistent in case of parameter variations. The \emph{Hyperbolic secant} method provided the lowest average computational time \modtextBis{in  generating walking trajectories}, but this result is sensitive to parameter variations.}

Finally, given the non-convex nature of the problem defined in Sec. \ref{sec:oc}, it is fundamental to provide a meaningful initial guess. Indeed, local minima may bring the planner to ``procrastinate'', as anticipated in Sec. \ref{sec:considerations}. 
An interesting future work consists in adopting Reinforcement Learning techniques, like \modtextDani{\cite{peng2018deepmimic, viceconte2022adherent}} to warm start the optimization problem.
In addition, the definition of references can affect the time necessary to find a solution. In future works, we will investigate both the adoption of faster solvers and the definition of a more sophisticated and efficient way of providing references.

\section{Code and Multimedia Material}
The code of this project is available at the link \url{https://github.com/ami-iit/dynamical-planner}. The definition of the \texttt{Github Actions} used for the comparisons of Sec. \ref{sec:complementarity_comparison} and the instructions on how to reproduce the results are available at \url{https://github.com/ami-iit/paper_dafarra_2022_tro_dcc-planner}. A video describing the paper is available at the link \url{https://youtu.be/Uc9o8TE32cw}.

\ifCLASSOPTIONcaptionsoff
  \newpage
\fi



\bibliographystyle{IEEEtran}
%
\bibliography{IEEEabrv,Bibliography/Bibliography}


%

\begin{IEEEbiography}[{\includegraphics[width=1in,height=1.25in,clip,keepaspectratio]{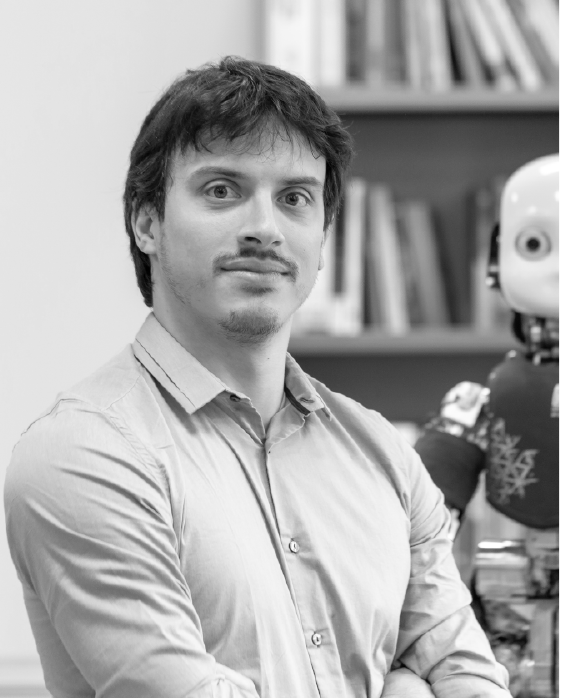}}]{Stefano Dafarra}
obtained his M.S. degree on Automation and Control Engineering from Politecnico di Milano in 2016. He received his Ph.D on Advanced and Humanoid robotics from the University of Genoa working at the Istituto Italiano di Tecnologia (IIT) in 2020. He is currently a Post-Doc at IIT. His research interests includes optimization, optimal control, and humanoid locomotion.
\end{IEEEbiography}

\begin{IEEEbiography}[{\includegraphics[width=1in,height=1.25in,clip,keepaspectratio]{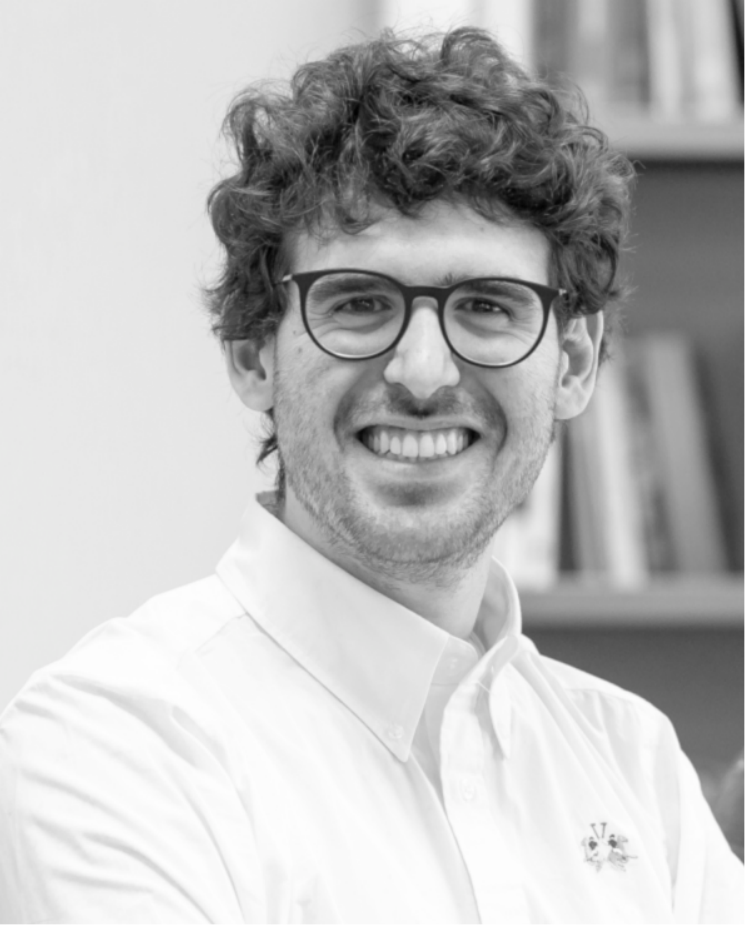}}]{Giulio Romualdi} received his B.S. degree in Biomedical Engineering from the University of Pisa, Italy, and his M.S. degree in Robotics and Automation Engineering from the University of Pisa, Italy, in 2014 and 2018, respectively. He is currently a Ph.D. student at the University of Genova, Italy and Istituto Italiano di Tecnologia, Italy. His research interests include bipedal locomotion, humanoid robots and nonlinear control theory.
\end{IEEEbiography}


\begin{IEEEbiography}[{\includegraphics[width=1in,height=1.3in,clip,keepaspectratio]{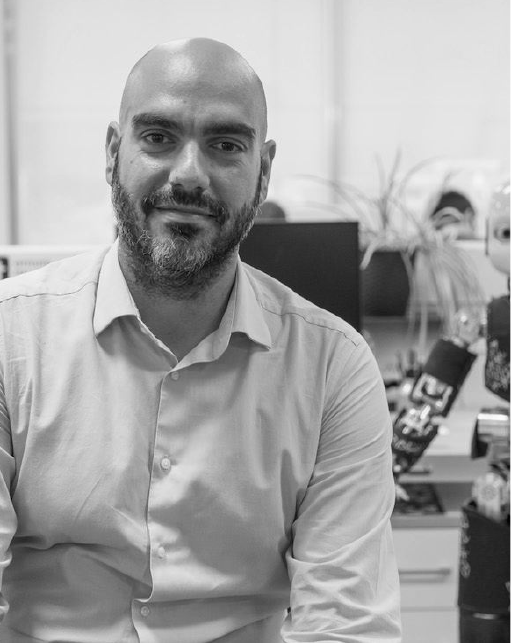}}]{Daniele Pucci}
\modtextDani{
received the bachelor and master degrees in Control Engineering with highest honors from ”Sapienza”, University of Rome, in 2007 and 2009, respectively. In 2009, he also received the ”Excellence Path Award” from Sapienza. In 2013, he earned the PhD title with a thesis prepared at INRIA Sophia Antipolis, France, under the supervision of Tarek Hamel and Claude Samson. The PhD program was jointly with ”Sapienza”, University of Rome, so in 2013 Daniele received also the PhD title in Control Engineering from Sapienza with the supervision of Salvatore Monaco. From 2013 to 2017, he has been a postdoc at the Istituto Italiano di Tecnologia (IIT) working within the EU project CoDyCo. From August 2017 to August 2021, he has been the head of the Dynamic Interaction Control lab.  The main lab research focus was on the humanoid robot locomotion and physical interaction problem, with specific attention on the control and planning of the associated nonlinear systems. Also, the lab has been pioneering Aerial Humanoid Robotics, whose main aim is to make flying humanoid robots. 
Daniele has also been the scientific PI of the H2020 European Project AnDy, he is task leader of the H2020 European Project SoftManBot, and he is the coordinator of the joint laboratory between IIT and Honda JP. In 2019, he was awarded as Innovator of the year Under 35 Europe from the MIT Technology Review magazine. Since 2020 and in the context of the split site PhD supervision program, Daniele is a visiting lecturer at University of Manchester. In July 2020, Daniele was selected as a member of the 
Global Partnership on Artificial Intelligence (GPAI). Since September 1st 2021, Daniele is the PI leading  the Artificial and Mechanical Intelligence research line at IIT. The research team combines AI and Mechanics to devise the next generation of Humanoid Robots.}
\end{IEEEbiography}




\end{document}